\documentclass[lettersize,journal]{IEEEtran}

\usepackage{amsmath,amsfonts}
\usepackage{algorithmic}
\usepackage{algorithm}
\usepackage{array}
\usepackage[caption=false,font=normalsize,labelfont=sf,textfont=sf]{subfig}
\usepackage{textcomp}
\usepackage{stfloats}
\usepackage{url}
\usepackage{verbatim}
\usepackage{graphicx}
\usepackage{tabularx}
\usepackage[dvipsnames]{xcolor}
\usepackage{hyperref}
\usepackage{makecell}
\usepackage[inkscapelatex=false]{svg}
\usepackage{pifont}
\newcommand{\cmark}{\color{ForestGreen}\ding{51}}%
\newcommand{\xmark}{\color{red}\ding{55}}
\usepackage[square,sort,comma,numbers]{natbib}
\usepackage{xcolor}
\usepackage{colortbl}

\usepackage{authblk}  
\makeatletter
\renewcommand\AB@affilsepx{\protect\Affilfont\quad}
\makeatother

\usepackage{tikz}
\usepackage[edges]{forest}
\definecolor{hidden-draw}{RGB}{205, 44, 36}
\definecolor{hidden-blue}{RGB}{194,232,247}
\definecolor{hidden-orange}{RGB}{243,202,120}
\definecolor{hidden-yellow}{RGB}{242,244,193}
\definecolor{tree-level-1}{RGB}{245,20,85}
\definecolor{tree-level-2}{RGB}{246,86,118}
\definecolor{tree-level-3}{RGB}{248,177,193}
\definecolor{tree-leaf}{RGB}{176,230,198}

\newcommand*{\revised}{\textcolor{black}}

\newcommand*{\moved}{\textcolor{black}}
\newcommand*{\solution}{\textcolor{black}}

\begin{document}
\bstctlcite{IEEEexample:BSTcontrol} 

\title{\huge Humanoid Locomotion and Manipulation: Current Progress and Challenges in Control, Planning, and Learning
\thanks{*co-corresponding authors}
}

\author[1]{Zhaoyuan Gu}
\author[2]{Junheng Li}
\author[3]{Wenlan Shen}
\author[4]{Wenhao Yu}
\author[5]{Zhaoming Xie}
\author[6]{Stephen McCrory}
\author[7]{Xianyi Cheng}
\author[1,8]{Abdulaziz Shamsah}
\author[6]{Robert Griffin}
\author[9]{C. Karen Liu}
\author[10,11]{Abderrahmane Kheddar}
\author[12]{Xue Bin Peng}
\author[13,14]{Yuke Zhu}
\author[15]{Guanya Shi}
\author[2]{Quan Nguyen}
\author[3]{Gordon Cheng}
\author[16]{Huijun Gao*}
\author[1]{Ye Zhao*}

\affil[1]{Georgia Institute of Technology}
\affil[2]{The University of Southern California}
\affil[3]{Technische Universität München}
\affil[4]{Google DeepMind}
\affil[5]{The AI Institute}
\affil[6]{The Institute for Human and Machine Cognition}
\affil[7]{Duke University}
\affil[8]{Kuwait University}
\affil[9]{Stanford University}
\affil[10]{CNRS-University of Montpellier LIRMM}
\affil[11]{CNRS-AIST Joint Robotics Laboratory}
\affil[12]{Simon Fraser University}
\affil[13]{The University of Texas at Austin}
\affil[14]{NVIDIA}
\affil[15]{Carnegie Mellon University}
\affil[16]{Harbin Institute of Technology}

\maketitle

\begin{abstract}
\revised{Humanoid robots hold great potential to perform various human-level skills, involving unified locomotion and manipulation in real-world settings. 
Driven by advances in machine learning and the strength of existing model-based approaches, these capabilities have progressed rapidly, but often separately. 
This survey offers a comprehensive overview of the state-of-the-art in humanoid locomotion and manipulation (HLM), with a focus on control, planning, and learning methods. 
We first review the model-based methods that have been the backbone of humanoid robotics for the past three decades. We discuss contact planning, motion planning, and whole-body control, highlighting the trade-offs between model fidelity and computational efficiency. Then the focus is shifted to examine emerging learning-based methods, with an emphasis on reinforcement and imitation learning that enhance the robustness and versatility of loco-manipulation skills. Furthermore, we assess the potential of integrating foundation models with humanoid embodiments to enable the development of generalist humanoid agents. This survey also highlights the emerging role of tactile sensing, particularly whole-body tactile feedback, as a crucial modality for handling contact-rich interactions. Finally, we compare the strengths and limitations of model-based and learning-based paradigms from multiple perspectives, such as robustness, computational efficiency, versatility, and generalizability, and suggest potential solutions to existing challenges.}
\end{abstract}

\begin{IEEEkeywords}
Humanoid robotics, Loco-manipulation, Model predictive control, Whole-body control, Imitation learning, Foundation models, and Whole-body tactile sensing.
\end{IEEEkeywords}

\section{Introduction}

\begin{figure}
\centering
\includegraphics[width=0.48\textwidth]{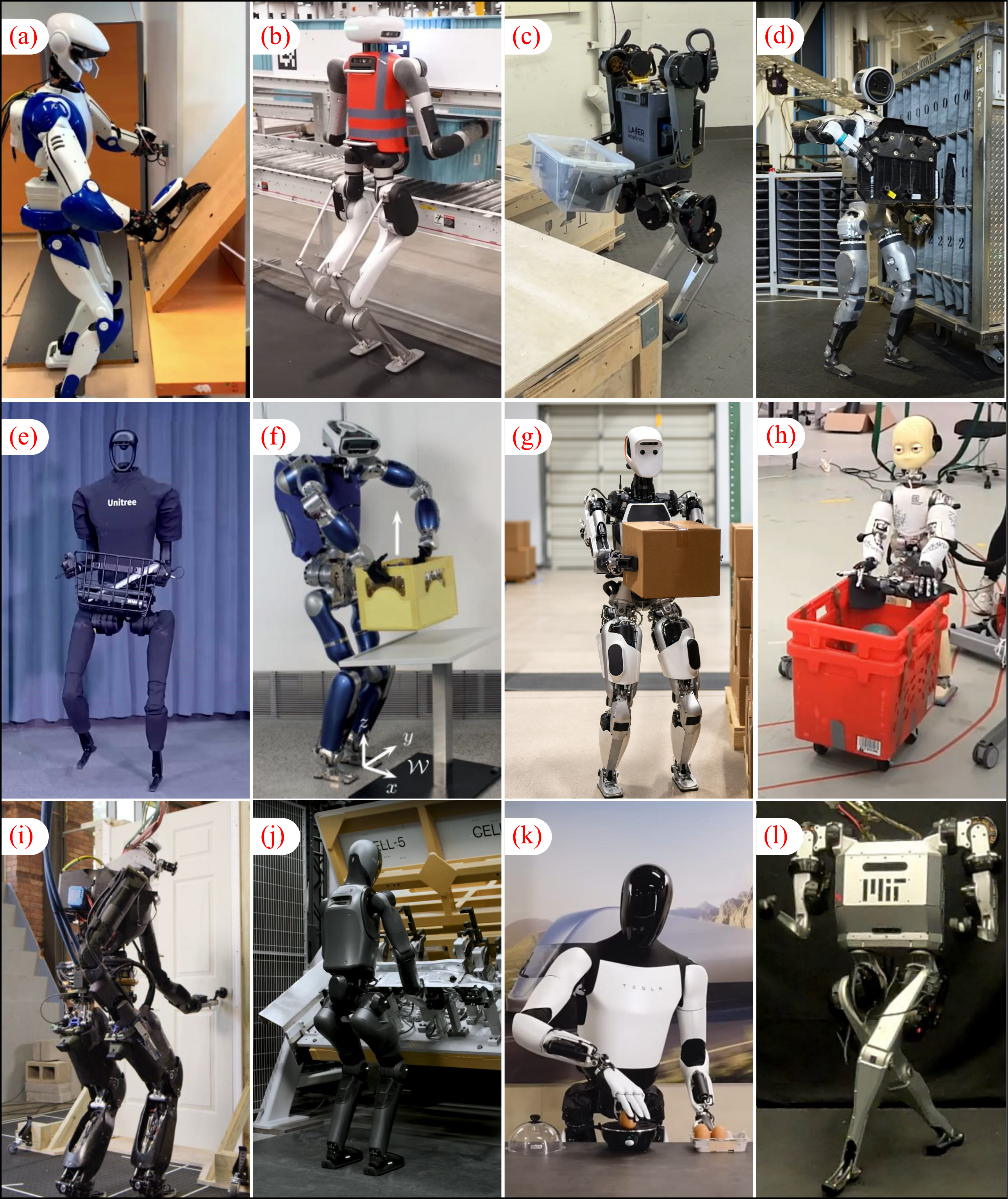}
\caption{
Humanoids executing locomotion and manipulation tasks: 
(a) HRP-4 wipes a wood board while adapting to terrain~\cite{humanoid_shuffle};
(b-g) Object pick and place by Digit, Hector~\cite{li2023dynamic}, Atlas, 
H1, \revised{TORO}~\cite{henze2016passivity}, and Apollo; (h) iCub pushes a cart~\cite{DS_iCub}; 
(i) Nadia opens a door~\cite{Nadia_door_auto}; 
(j-k) Object manipulation by \revised{Figure 02} and Optimus;
(l) MIT humanoid whole-body push recovery~\cite{khazoom2024tailoring}.
}
\label{fig:intro_figure}
\vspace{-0.2in}
\end{figure}


\moved{A humanoid robot refers to any anthropomorphic robot that resembles the form of a human~\cite{humanoid_reference}. Typically, a humanoid robot possesses a torso, two arms, and two legs, though the degree of anthropomorphism may vary. For instance, some humanoid robots feature simple hands or wheeled legs~\cite{BHR-WI_Compliant}.} 
\moved{The primary focus of this review is on humanoid robots that replicate human morphologies and functionalities, rather than those that closely mimic human-like visual appearance and external aesthetics. }
\revised{Specifically, this survey examines bipedal humanoid robots, with an emphasis on the whole-body motion of humanoid robots instead of on the dexterous manipulation of multi-finger hands.}

Humanoid robots are well suited for executing human-level tasks, as they are built to (ideally) replicate human motions and achieve various whole-body loco-manipulation tasks, \textit{e.g.}, applications ranging from manufacturing to services, as shown in Fig.~\ref{fig:intro_figure}. Their anthropomorphism makes them stand out from other robot forms in performing human-like tasks, such as payload carrying over stairs, \moved{especially in environments designed for humans.} Furthermore, humanoid robots can interact with humans for physical collaboration tasks, such as collaboratively moving a large table upstairs with human assistance. However, simultaneously achieving these intricate tasks while addressing highly complex robot dynamics is still challenging, let alone safe physical collaborations with humans and/or operations in unstructured environments. 
As a promising direction to solve this problem, humanoids can exploit the abundance of data collected from humans to quickly acquire motor and cognitive skills due to their human-like morphology.
Therefore, scaling human data and computational resources for humanoid embodiment is potentially a fast route to embodied intelligence.

Embodied intelligence at large is thriving at an unprecedented pace. Perception algorithms can detect, classify, and segment a wide variety of objects in real time. Model-based methods that leverage both predictive and reactive control have enabled agile and reliable locomotion and manipulation. Meanwhile, deep learning policies have demonstrated convincing control results on robot hardware through exploration and imitation. Large foundation models trained on massive, internet-scale datasets are beginning to show cognitive capabilities of open-world reasoning. Consequently, the building of autonomous humanoid robots for real-world applications has become possible, leading to the emergence of many humanoid robot companies and concrete deployment applications. Especially, companies with powerful GPU-based parallelization capabilities, such as NVIDIA, and companies building physical humanoid robots, such as Boston Dynamics, Tesla, and Figure,  have begun collaboration on the embodied intelligence of humanoid robots. 


\begin{figure*}[t]
\centerline{\includegraphics[width=0.65\textwidth]{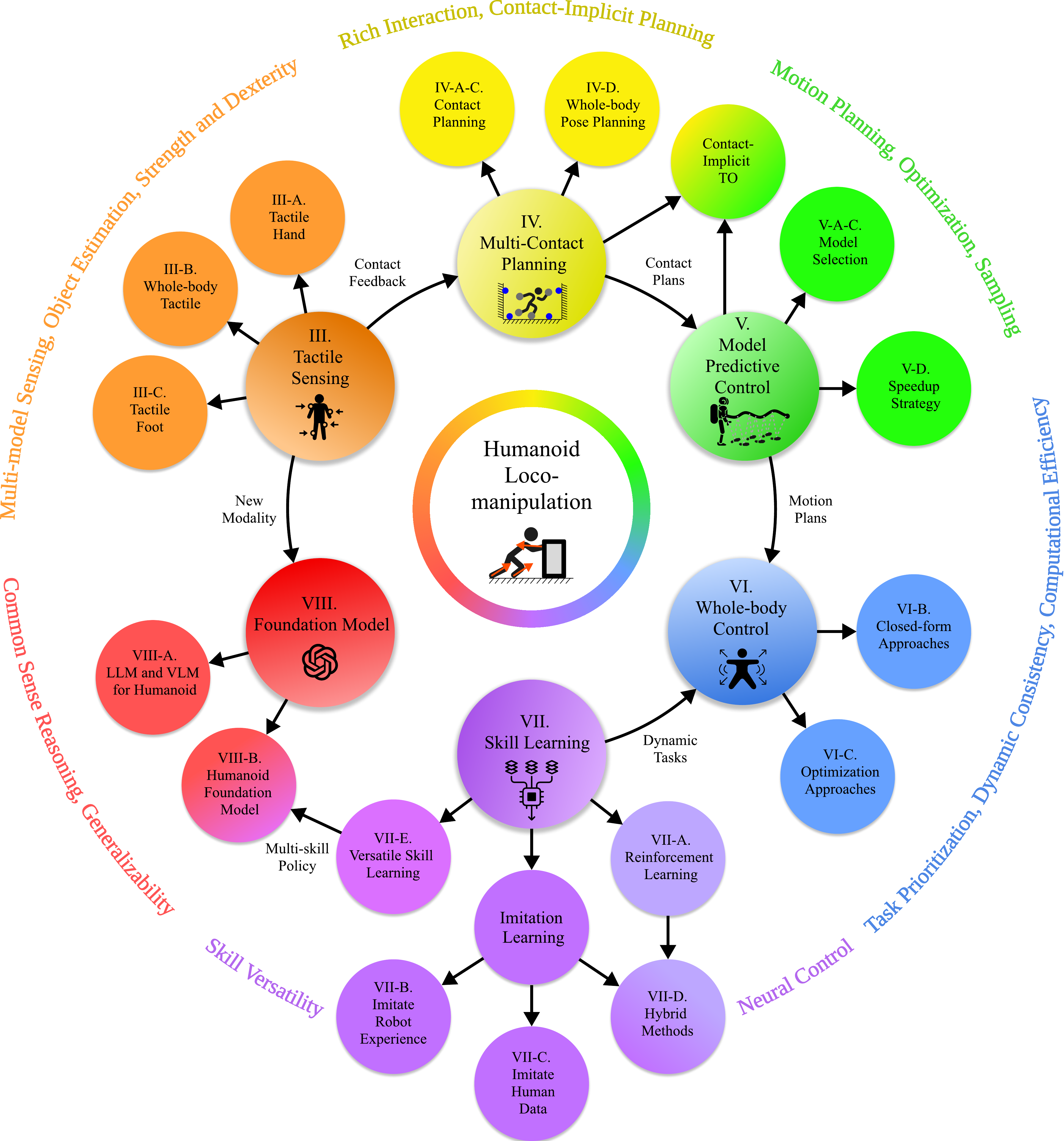}}
\caption{This survey begins by defining relevant concepts of humanoid robots and their locomotion and manipulation capabilities. Centered around achieving humanoid loco-manipulation tasks, the core of this survey delves into two main categories of methods: the traditional planning and control approaches, such as contact planning, motion planning, and control, as well as the emerging learning-based approaches, including skill learning and foundation models. In addition, this survey highlights whole-body tactile sensing as a crucial modality to achieve contact-rich loco-manipulation.}
\label{fig:framework}
\vspace{-0.15in}
\end{figure*}

Acknowledging the rapid advancements in humanoid robotics, this article reviews recent developments in Humanoid Locomotion and Manipulation (HLM). As laid out in Fig.~\ref{fig:framework}, humanoid robotics is a multidisciplinary field that spans domains in design, actuation, sensing, control, planning, and decision-making. In this survey, we mainly examine task planning, motion planning, policy learning, and control from the perspective of model-based and learning-based methods. We first synthesize traditional model-based methods for planning and control. Then, we shift our focus to more recent learning-based approaches, especially those leveraging reinforcement learning, imitation learning, and foundation models. 

Model-based methods serve as the cornerstone for enabling HLM capabilities. These methods depend critically on physical models, which can significantly influence the quality, speed, and guarantees of motion generation and control. Over the past decade, planning and control techniques have shown a trend of converging to the predictive-reactive control hierarchy, employing a whole-body model predictive controller (MPC) or simplified model (centroidal dynamics) MPC coupled with local task-space Whole-Body Controllers (WBC)~\cite{Wensing_survey_TRO}. These planning and control techniques are usually formulated as Optimal Control Problems (OCPs) that are solved by off-the-shelf or customized numerical solvers. Although these numerical optimization methods are well-established, research continues to focus on enhancing their computational efficiency, numerical stability, robustness, and scalability for high-dimensional systems. 

Learning-based approaches have witnessed a rapid surge in humanoid robotics and achieved impressive results that attract an increasing number of researchers to the field. Among the diverse learning approaches, Reinforcement Learning (RL) has proven its ability to achieve robust motor skills. However, despite its ability to discover novel behaviors via trial and error, pure RL without demonstration data is often prohibitively inefficient for HLM tasks, which are characterized by high degrees-of-freedom robots and sparse reward settings. Therefore, sim-to-real RL, where training is conducted in simulation before transferring to the physical robot, has become a prevalent method, though it faces the challenge of bridging sim-to-real gaps. \revised{From the existing literature, we observe that sim-to-real RL has gained significant traction for achieving humanoid whole-body motion. There is an ongoing debate between model-based and learning-based control, and therefore we highlight two key points: (1) sim-to-real RL that relies on the accurate dynamics models in a simulator can also be viewed as a ``model-based," method, even though the RL algorithm itself does not explicitly model the dynamics; and (2) model-based control and sim-to-real RL do not conflict; on the contrary, they often complement each other and can be combined to achieve a better performance than either method alone.}

Imitation learning (IL) from expert demonstrations has proven to be an efficient method of acquiring motor skills. IL techniques, such as behavior cloning~\cite{diffusion_policy}, have shown impressive abilities to mimic a wide range of skills. In pursuit of versatile and generalizable policies through IL, many researchers and companies have focused on scaling data. While robot experience data can be diverse and high-quality, its acquisition is both expensive and time-consuming. Thus, learning from human data, which is abundant and readily available from Internet videos and public datasets, emerges as a pivotal strategy for humanoid robotics. Learning from humans is a unique advantage exclusive to humanoid robots. However, even though humanoid robots may attain human-level motor skills, a deeper question of embodied intelligence persists: how to learn the intentions (source) behind human actions rather than merely replicating the observed motions (outcome). It is hypothesized that understanding human intention is achieved via Foundation Models (FMs) that are capable of semantic interpretation of the environment and the task. This hypothesis motivates us to include FMs as a part of our survey. 

The remarkable success of FMs has sparked a surge in \textit{general} robotics research, as highlighted by several comprehensive surveys~\cite{firoozi2023foundation, Robots_via_Foundation_Models}. This paper surveys the application of FMs specifically for humanoid robots. 
FMs offer a promising solution to the persistent challenge of generalizability in robotics by efficiently harnessing internet-scale datasets to acquire extensive knowledge. A pre-trained FM exhibits capabilities for open-world reasoning and multimodal semantic understanding. These capabilities are invaluable for robots engaged in complex environments requiring long-term, logically coherent task planning. Within the realm of humanoid robots, FMs have been successfully implemented as task planning modules in hierarchical planning and control frameworks.
Training end-to-end FMs for humanoid applications is an emerging area, poised for significant future advancements.

\subsection{Survey Roadmap}

\revised{For the selection criteria of papers cited in this survey, we prioritize topics most relevant to humanoid loco-manipulation capabilities, including tactile sensing \cite{dahiya2009tactile}, model-based planning~\cite{MPC_survey} and control~\cite{Wensing_survey_TRO}, as well as learning-based methods such as RL~\cite{DRL_Legged_Survey, ha2024learningbased}, IL~\cite{Gams2022, ParraMoreno2023ImitationMF}, and FMs~\cite{Robots_via_Foundation_Models, firoozi2023foundation}. Distinct from survey papers on these topics, our paper aims to present a broad overview with a focus on humanoid loco-manipulation.}

\revised{
Each of these topics has extensive studies. For each topic, we aim to spotlight three categories of papers: (i) foundational studies that established key principles in these areas, (ii) recent works representing the state of the art, and (iii) survey papers for further reading. Our survey spans 30 years of research. Specifically, model-based planning and control papers have been collected since 1990, and learning-based methods predominantly span the past 20 years.}

\revised{Compared with existing survey reviews on humanoid robotics~\cite{humanoid_reference, humanoid_survey_IJAS}, this survey collects up-to-date papers and introduces in detail the new paradigm of learning-based methods.}

As shown in Fig.~\ref{fig:framework}, this survey is organized in the following order. We first establish the background, defining humanoid robots and the key capability of locomotion and manipulation in Sec.~\ref{sec:background}. We detail whole-body tactile sensing in Sec.~\ref{sec:sensing}. 
We then present traditional approaches that achieve loco-manipulation, including contact planning (Sec.~\ref{sec:contact_planning}), motion planning (Sec.~\ref{sec:planning}), and control (Sec.~\ref{sec:control}). 

We then examine the state-of-the-art learning-based algorithms. In Sec.~\ref{sec:learning}, we explore approaches using reinforcement learning and imitation learning to acquire loco-manipulation skills. In Sec.~\ref{sec:foundation_model}, we discuss how foundation models have become the backbone of semantic understanding and decision-making for effective humanoid task planning.  
Finally, we highlight significant challenges in this field and present our perspectives on potential future research directions and emerging opportunities in Sec.~\ref{sec:discussion}.

\subsection{Survey Contributions}
\revised{
We summarize our core contributions as follows:
\begin{itemize}
\item The survey serves as an effective resource for graduate students and researchers new to the field, offering a comprehensive review of humanoid technical methods, while also providing perspectives for humanoid experts in academia and industry with the latest advancements. 
\item We advance the understanding of loco-manipulation in bipedal humanoid robots by clearly defining the problem, surveying state-of-the-art methods and capabilities, and outlining remaining challenges along with promising solutions. 
\item We provide a clear comparison between model-based and learning-based methods, and their corresponding loco-manipulation skills achieved on humanoid robots. This comparison offers valuable insights for researchers deciding their methods. 
\end{itemize}}

\section{Background}
\label{sec:background}

In this section, we focus on the humanoid robot's main capabilities: bipedal locomotion and whole-body manipulation. Finally, we detail the combined loco-manipulation skills with the state-of-the-art methods and current challenges.

\subsection{Bipedal Locomotion}

Bipedal locomotion is a significant characteristic of humanoid robots. Therefore, in the past three decades, bipedal locomotion has been a prolific field of humanoid research.
Interested readers can refer to the excellent reviews (most of which are recent)~\cite{wieber2016shr,carpentier2021crr,khan2023robotica}, and the monographs~\cite{Westervelt2018,kajita2014book}.
In summary, model-based bipedal locomotion has evolved significantly, progressing from passive walking~\cite{passive_walking_McGeer, Passive-Walkers} to quasi-static walking~\cite{Kajita2003}, and then to dynamic walking \cite{Westervelt2018}. Bipedal walking on flat surfaces has been well explored and mastered through periodic motions with model-based methods~\cite{Westervelt2018, kim_dynamic_2020}. 
These approaches have also expanded to more agile motions such as jumping \cite{li2024continuous, wensing_thesis} and back-flipping \cite{pardis2024probabilistic}. 

Bipedal locomotion under external perturbations and force loads has been extensively studied. Such capabilities lay the foundation for simultaneous locomotion and manipulation, the focus of this survey. Model-based methods are among the early works that demonstrate these capabilities. 
For example, ~\cite{englsberger2020mptc} introduced a passivity-based controller with task space dynamics that treats external forces as part of the robot’s generalized forces. \cite{li2023dynamic} incorporated a payload into a simplified rigid body model to enable dynamic walking while carrying a load, and \cite{agravante2019human} integrated external forces into a simplified linear inverted pendulum model. 

In addition to model-based methods, bipedal locomotion has also been successfully addressed by learning-based methods~\cite{Atkeson_Biped_Learn_2007, Nakanishi2004_LearnBiped, Calandra2015_BO_Biped}, particularly in the context of periodic motions on flat surfaces. Furthermore, learning-based approaches have also demonstrated capabilities in more complex settings, such as running~\cite{Cassie_dash}, jumping~\cite{li2023robust}, and handling non-periodic motions such as stair climbing~\cite{siekmann2023blind} and parkour~\cite{humanoid_parkour_learning}. Similar to the trend in model-based methods, learning-based methods have further extended their capabilities to handle external forces and payloads~\cite{dao2022sim, transformer_digit_RL}. 

\revised{Researchers have explored energy efficiency in bipedal locomotion by examining the Cost of Transport (COT), which designates energy expenditure per unit distance traveled, normalized by body weight. Early work in passive dynamic walking \cite{passive_walking_McGeer, Passive-Walkers} achieved higher efficiency than humans but offered limited versatility. Meanwhile, today's humanoid robots $(\text{COT}>0.7)$ are significantly less efficient than humans $(\text{COT}=0.2)$ \cite{Locomotion_efficiency_survey_2018}. Therefore, efficient walking continues to motivate hardware designs that incorporate passive-compliant components for energy storage \cite{DURUS, lock_ankle} and control algorithms that exploit the system’s natural dynamics \cite{xia2024duke, RAL_MOE_Humanoid}.} 

\subsection{Bipedal Navigation}
\revised{Proficiency in bipedal locomotion has naturally progressed to humanoid robots' ability to effectively navigate complex environments, including indoor and outdoor areas with uneven terrain and dynamic obstacles. A navigation stack often incorporates a hierarchical structure: a global path planner and a local step planner. The global path planner~\cite{huang_cassie_navigation,muenprasitivej2024bipedal,mccrory2023bipedal,li2023autonomous, cheng2024navila} is typically responsible for understanding the overall navigation task and generating a path that avoids obstacles and reaches the target location. On the other hand, local step planners, \emph{e.g.,}~\cite{Acosta_Cassie, duan2023learning, griffin2019astar} focus on determining the precise foot placements that adhere to the bipedal dynamics within the immediate surroundings of the robot while also tracking the global path.}


\revised{From the aforementioned navigation stack, bipedal navigation capabilities have progressed from static obstacle avoidance on flat terrain~\cite{peng2023safe} to more challenging scenarios, including locomotion through height-constrained space~\cite{li2023autonomous, Zhao_CDC2016}, avoiding dynamic obstacles in a constrained environment~\cite{shamsah2023TRO}, navigating dynamic social environments using reachable sets~\cite{shamsah2024socially}, or deep RL that incorporates the perceived pedestrian emotions into the navigation plan~\cite{zhu2025emobipednav}, and traversing rough terrains~\cite{huang_cassie_navigation,mccrory2023bipedal, gibson2022terrain, muenprasitivej2024bipedal, global_step_planner_TMECH, shamsah2024terrain_predictivecontrol}. 
A persistent challenge for these methods is that they are tailored to specific use case scenarios and lack the versatility to handle a wide range of different situations.}

\revised{While bipedal locomotion and navigation have been widely studied, real-world deployment remains a significant challenge due to inherent uncertainties. Uncertainty can arise from the environment and the robot model. Real-world environments have uneven, varying terrain, dynamic obstacles, and occlusion, making it difficult to ensure the safety and robustness of bipedal navigation. On the other hand, model uncertainty arises from discrepancies in the mathematical representation of the robot model and the physical system. Model uncertainty also exists in most current navigation frameworks that employ reduced-ordered models at the high level for collision avoidance and goal-reaching tasks and a full-order model at the low level for tracking high-level commands. A coupled framework that considers both the navigation task and whole-body control stability and accuracy is under-explored. Although previous works have addressed various aspects of environment uncertainties~\cite{jiang2023abstraction} and model uncertainties~\cite{ahmadi2021risk}, a comprehensive navigation stack capable of handling the full spectrum of real-world uncertainties is still essential.}

\subsection{Whole-body Manipulation} 
\label{sec:whole-body-manipulation}

Anthropomorphic manipulation has been the inspiration for bimanual manipulation~\cite{smith2012dual}, loco-manipulation, and dexterous manipulation~\cite{pollard1994wholehand}. The ultimate form of anthropomorphic manipulation is whole-body manipulation, referring to the ability to manipulate objects using any part of one's body. For example, humans use their elbows or hips to hold a door open for convenience; humans use their palms or fists instead of fingertips to provide large forces; humans curl their little fingers to hold a small object while still using other fingers for manipulation. In comparison, most robots often have predefined end-effectors, such as foot soles or fingertips, as the only parts allowed to physically interact with the world. 
Whole-body manipulation is a grand problem that shares challenges in bimanual manipulation, loco-manipulation, and dexterous manipulation. 
This general ability has yet to be developed, but its emergence will indicate a breakthrough for robotic manipulation.

The idea of whole-body manipulation was originally studied in the whole-arm manipulation community~\cite{salisbury1987whole}. 
Whole arm manipulators were designed and built to explore the benefit of manipulating objects with all surfaces of a robot manipulator~\cite{townsend1993mechanical}. 
This brings a unique challenge that manifests itself in all the system levels in perception, estimation, planning, and control. 
Since there are an infinite number of such contacts, the planning complexity suffers from the combinatorial explosion of contact modes~\cite{cheng2023enhancing} and exponential computational costs from the high degree of freedom of the system~\cite{smith2012dual}.

Numerous breakthroughs in mechanical design, control, and planning have been achieved in the endeavor to address the challenges of whole-body manipulation. 
On the mechanical design side, robots made with soft materials and full-body sensing, such as Punyo~\cite{Punyo}, provide whole-body manipulation capability in a built-in manner. 

For control, the coordinative and contact-rich nature requires forceful and compliant control. Traditionally, robot arms were hard-coded to switch across different control strategies according to task requirements~\cite{smith2012dual}. 
Different task requirements, such as reaching a point or wiping a table, require different control strategies, such as pure position control or hybrid force position control.  
However, it is still unclear how to define and enumerate the control strategies for whole-body manipulation. In addition, a general control framework that can take in the sensor data, perform state estimation, and reactively control each body contact has yet to exist~\cite{stasse2020whole}. Such general frameworks require innovations in advanced hardware and algorithm architecture, including whole-body sensing~\cite{cheng2019robotskin} and robot designs with compliance and force control capabilities for reactive manipulation~\cite{gupta2022extrinsic}. 

From the planning perspective, the challenge of whole-body manipulation can be potentially alleviated via human behavior imitation algorithms~\cite{hsiao2006imitation, yuan2019reinforcement,zhang2023plan}. Most of these works focus on simple manipulation strategies such as whole-body grasping and pushing. To enable the robot to mimic more complex human whole-body manipulation behaviors, it is important to address the cross-morphology gap between humans and humanoids.

To achieve humanoid whole-body manipulation, full-stack system integration at all system levels is crucial. In the future, we expect to see hardware advances in whole-body sensing, compliant materials, and force-transparent mechanism design. Significant improvements on the algorithm side will also be needed. While classical planning and control approaches suffer from huge complexity issues, pure learning methods lack the flexibility to react to contacts and adapt to different tasks. We foresee that the solution will be an integrated approach, which combines the strength of both. Ultimately, this could lead to more complex, human-like capabilities in humanoid robots, merging improved control, adaptive learning, and comprehensive sensing. Furthermore, addressing the core issues in loco-manipulation will also shed light on whole-body manipulation, as both areas involve handling complex, contact-rich interactions on different body parts. 

\subsection{Loco-manipulation}
\label{sec:loco-manipulation}

\begin{table}[]
\centering
\setlength{\tabcolsep}{3pt}
\caption{Taxonomy of Whole-body Locomotion and Manipulation}
\begin{tabular}{llll}
\hline
\hline
 & \begin{tabular}[c]{@{}l@{}}(a) Whole-body \\ Manipulation\end{tabular} & \begin{tabular}[c]{@{}l@{}}(b) Whole-body \\ Loco-manipulation\end{tabular} & \begin{tabular}[c]{@{}l@{}}(c) Loco-\\ manipulation\end{tabular} \\ \hline \hline
\begin{tabular}[c]{@{}l@{}}Object movement \\ (Manipulation)\end{tabular}       &     \multicolumn{1}{c}{\cmark}  &   \multicolumn{1}{c}{\cmark}  &      \multicolumn{1}{c}{\cmark} \\ \hline
\begin{tabular}[c]{@{}l@{}}Robot self mobility\\  (Locomotion)\end{tabular}     &     \multicolumn{1}{c}{\xmark}   &  \multicolumn{1}{c}{\cmark}    &   \multicolumn{1}{c}{\cmark}      \\ \hline
\begin{tabular}[c]{@{}l@{}}All surface interaction \\ (Whole-body)\end{tabular} &    \multicolumn{1}{c}{\cmark}     &   \multicolumn{1}{c}{\cmark}  &  \multicolumn{1}{c}{\xmark} \\ \hline
\hline

\end{tabular}
\vspace{-0.1in}
\end{table}

\begin{figure}[]
    \centering
    \includegraphics[width=0.99\columnwidth]{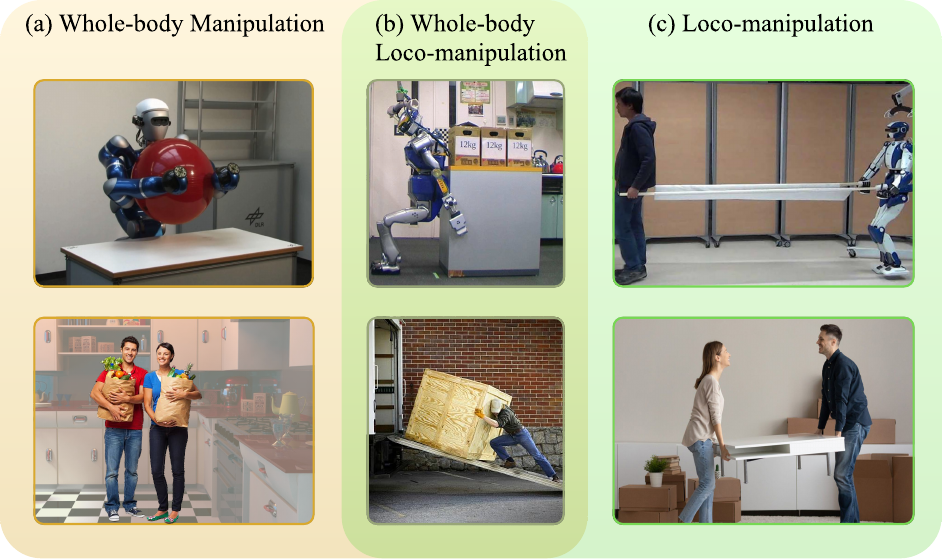}
    \caption{(a) Whole-body manipulation exemplified by human and humanoid Justin \cite{WAM_Justin} interacting with objects using all surfaces. (c) Loco-manipulation involves simultaneous locomotion and manipulation, as shown in the collaborative tasks performed by humans and a humanoid \cite{agravante2019human}. (b) Whole-body loco-manipulation is an intersection of (a) and (c), as exemplified by a human and a humanoid HRP-4 \cite{Murooka_heavy_push} pushing heavy objects using their legs and arms.}
    \label{fig:human-loco-manipulation}
    \vspace{-0.15in}
\end{figure}

One of the key features of humanoid robots is their ability to simultaneously perform locomotion and manipulation (abbreviated as loco-manipulation hereafter) tasks. 
As suggested by its name, loco-manipulation involves both the movement of objects through manipulation and the mobility of the robot itself through locomotion. In a more general case of \textit{whole-body} loco-manipulation, the \textit{whole-body} refers to the use of all body surfaces to interact with the environment. We summarize the relationship between loco-manipulation, and whole-body manipulation in Table~I. 
Both whole-body manipulation and loco-manipulation highlight the importance of utilizing physical contact. 
As shown in Fig.~\ref{fig:human-loco-manipulation}, loco-manipulation considers the movement of the robot itself while it manipulates an object, whereas whole-body manipulation emphasizes leveraging all accessible robot contact surfaces, such as using the chest as extra support to move large objects.

Loco-manipulation capability has been widely demonstrated on quadruped robots, specifically those achieving loco-manipulation capability by using their limbs as manipulators~\cite{legAsManip, he2024learningvisualquadrupedallocomanipulation, 2024pedipulate}. For quadrupeds with upper-body manipulators, whole-body control is widely adopted for pick and place tasks from the model-based~\cite{Ferrolho2023, HutterScience_locomanip} and learning-based community~\cite{portela2024learning, liu2024visualwholebodycontrollegged}. 

Loco-manipulation for humanoid robots is particularly challenging, compared with quadrupeds. Humanoid robots have a smaller support region on the ground and a higher center of mass, which is challenging for dynamic balance. Therefore, early humanoid frameworks focus on separate control for locomotion and manipulation. For example, in locomotion tasks, most studies constrain the upper body to remain upright, which simplifies the whole-body problem to a bipedal locomotion problem that considers only the low\revised{er} limbs. Conversely, in most table-top manipulation tasks, the lower body of the humanoid remains stationary \cite{DRACO_teleop, SR_humanoid_bimanual}. In such cases, any external force exerted on the upper body is treated as a disturbance to the legs, whose goal is to solely maintain balance. On the contrary, in~\cite{humanoid_shuffle}, there is no such categorization of contacts: all contacts contribute simultaneously to achieve the task and balance.

Humanoid loco-manipulation requires a holistic and strategic use of the entire body to explore the humanoid's full behavioral capability space. Additionally, whole-body loco-manipulation needs to schedule contact for all limbs to simultaneously achieve robust movement and safe object interaction. Acquiring this technique unlocks a broad range of useful tasks such as opening doors~\cite{seo2023deep,jorgensen2019finding}, pushing trolleys~\cite{Murooka_Pattern, Vaz2023}, rolling large bobbins~\cite{chappellet2023humanoid}, or climbing ladders~\cite{vaillant2016auro,ferrari2023multi}. 


\section{Tactile Sensing}
\label{sec:sensing}
Humanoid locomotion and manipulation involve extensive physical interactions with the environment and objects, requiring multimodal sensing for understanding the environment, tracking manipulated objects, and evaluating how contact impacts the balance of both the robot and the objects. 
Visual sensors have shown effectiveness in object tracking and simultaneous localization and mapping (SLAM)~\cite{chappellet2023humanoid}, while proprioceptive sensors are usually combined to estimate contact information in contact-rich tasks~\cite{seo2023deep}.  These sensory modalities have been widely adopted in existing systems and have been thoroughly reviewed in the literature~\cite{humanoid_survey_IJAS}. 

\revised{In contrast, tactile sensing, despite being uniquely suited for capturing detailed contact information critical for both locomotion and manipulation, has received relatively limited attention in existing humanoid literature. This survey specifically highlights tactile sensing to address this gap and to underscore its complementary role alongside other sensing modalities in humanoid systems.}

Mimicking the human sense of touch, tactile sensing provides more accurate and comprehensive contact information over large areas of robot skin compared to proprioceptive sensors~\cite{kheddar2011robio}, and allows the robot to perceive complex environments and assess object properties through physical interactions, especially in scenarios where vision is occluded~\cite{jain2013reaching}. Additionally, tactile sensing can be used to estimate contact-based object properties such as roughness, texture, and weight, complementing traditional visual information such as location, shape, and color~\cite{gao2022tactile}. A combination of tactile with other sensory modalities can significantly enhance humanoid perception capabilities in solving complex loco-manipulation tasks.

\begin{figure}
    \centering
    \includegraphics[width=\linewidth]{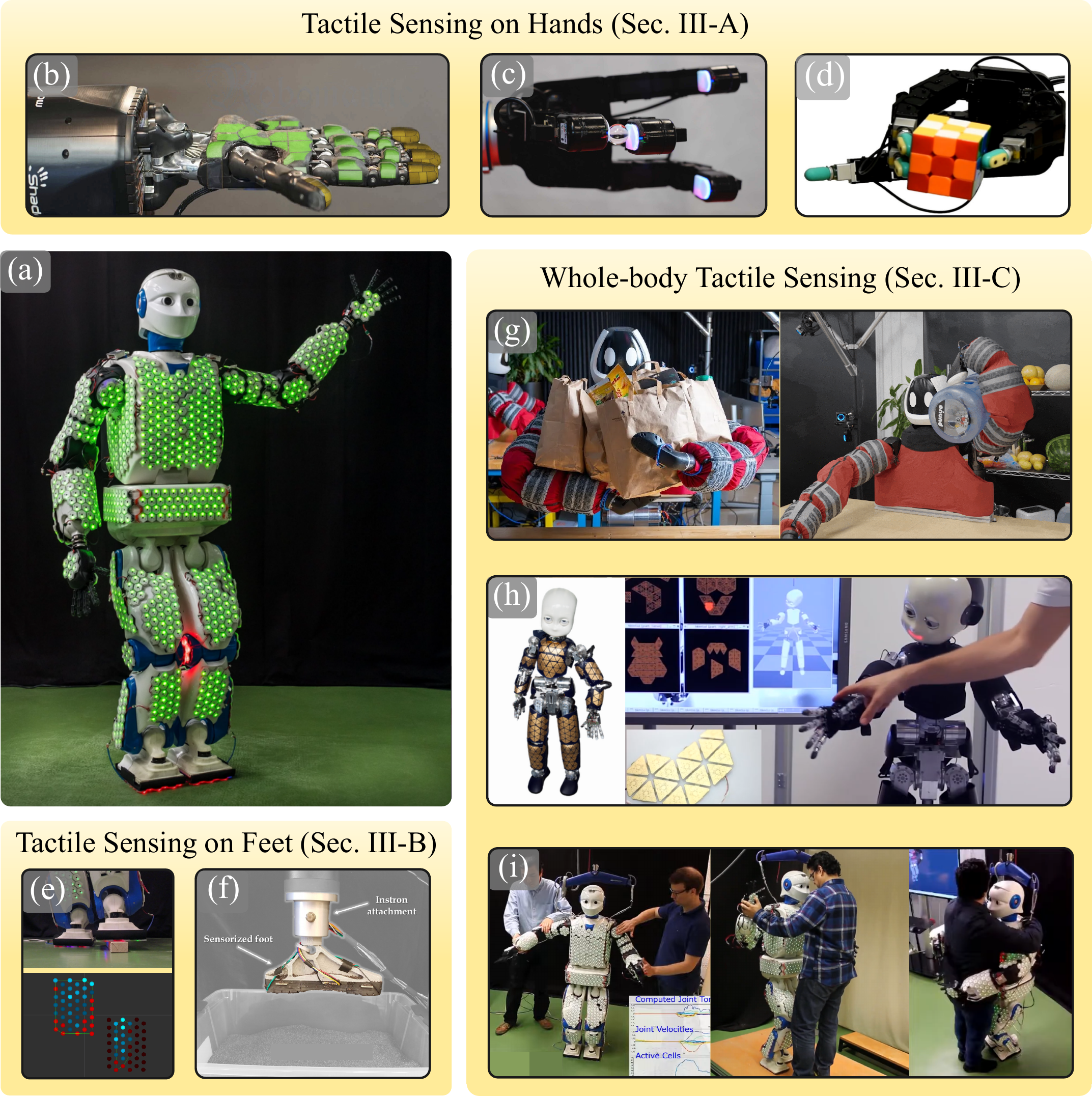}
    \caption{Tactile sensing on humanoid robots, exemplified by (a) REEM-C fully covered with artificial skin \cite{cheng2019robotskin} (image copyright: A. Eckert), which cover three body regions: hand, feet, and the whole body. (i) Hand tactile sensors demonstrated by (b) Shadow-Dexterous-Hand equipped with tactile sensors on palm and fingertips~\cite{melnik2021using}, (c) Allegro Hand equipped with DIGIT sensors~\cite{lambeta2020digit}, and (d) BioTac~\cite{veiga2020hierarchical} tactile sensors for dexterous manipulation; (ii) Tactile sensors on foot soles for (e) obstacle recognition \cite{guadarrama2018enhancing} and (f) terrain classification \cite{xiaofeng_foot}; (iii) Whole-body tactile sensors for (g) whole-body manipulation by Punyo-1 \cite{Punyo} and whole-body human-robot interaction by (h) iCub~\cite{maiolino2013flexible} and (i) REEM-C \cite{dean2019whole, tactile_dance}}
    \label{fig:tactile-sensing}
    \vspace{-0.1in}
\end{figure}

Numerous studies have developed tactile sensors based on various transduction principles that can sense normal and tangential forces, vibration, temperature, and pre-contact proximity information. Comparative studies of various sensor designs can be found in~\cite{dahiya2009tactile, kappassov2015tactile, al2020tactilesurvey}. This survey instead focuses on their application in humanoid loco-manipulation, categorized into three areas: (i)~tactile sensing on hands, (ii)~tactile sensing on foot soles, and (iii)~whole-body tactile sensing. The following sections review recent advancements in each domain, emphasizing their roles in balancing control, scheduling contacts, and enhancing interaction capabilities, as illustrated in Fig.~\ref{fig:tactile-sensing} and Table~\ref{tab:tactile}.

\begin{table*}[t]
    \centering
    \caption{Roles and challenges of tactile sensing in locomotion and manipulation applications}
    \begin{tabular}{lllll}
    \hline \hline
        Application & Roles of Tactile Sensing & Body Region Involved & Challenges & Solutions\\
    \hline \hline
        Manipulation & 
        \begin{tabular}[c]{@{}l@{}} \textbullet\ Grasping force and stability regulation\\~\cite{li2013control,veiga2020grip, song2013efficient, WBM_tactile_gordon} \\ \textbullet\ Object property and state estimation\\~\cite{sommer2014bimanual, hogan2020tactile, kaboli2018robust} \\ \textbullet\ Contact dynamics estimation\\~\cite{tian2019manipulation, lambeta2020digit} \\ \textbullet\ Input modality for RL/IL/FM training\\~\cite{van2015learning, melnik2021using, lin2024learning, fu2024touch, Yu-RSS-24} \end{tabular} & 
        \begin{tabular}[c]{@{}l@{}} \textbullet\ Hand (in IHM$^*$) \\ \textbullet\ Whole-body (in WHM$^*$) \end{tabular} &  \begin{tabular}[c]{@{}l@{}} \textbullet\ High-dimensional \\ contact space \\ \textbullet\ Tactile physics simulation \\ \textbullet\ Sim-to-real transfer\end{tabular}& 
        \begin{tabular}[c]{@{}l@{}} \textbullet\ Dimensionality reduction \\~\cite{chebotar2014learning, van2016stable,melnik2021using}\\ \textbullet\ Tactile simulators \\~\cite{wang2022tacto, lin2022tactile, akinola2024tacsl}
        \end{tabular}\\
    \hline
        Locomotion & 
        \begin{tabular}[c]{@{}l@{}} \textbullet\ GRF and support region estimation \\~\cite{guadarrama2018enhancing} \\ \textbullet\ Terrain property recognition \\~\cite{xiaofeng_foot,suwanratchatamanee2009simple} \end{tabular} &
        \textbullet\ Feet & \begin{tabular}[c]{@{}l@{}} \textbullet\ Sensor durability against \\ heavy weight and various \\terrains \end{tabular} & \begin{tabular}[c]{@{}l@{}} \textbullet\ Durable materials \\ and mechanical design \\ \cite{guadarrama2018enhancing, xiaofeng_foot, suwanratchatamanee2009simple}\end{tabular}\\
    \hline
        Loco-manipulation & \begin{tabular}[c]{@{}l@{}} 
        \textbullet\ All of above \\ \textbullet\ Unified locomotion and manipulation \\ planning \end{tabular} & \textbullet\ Whole-body & \begin{tabular}[c]{@{}l@{}} \textbullet\ All of Above \\\textbullet\ Real-time contact \\ reasoning  and adaptation \end{tabular} & \textbullet\ To be explored\\
    \hline \hline
    \end{tabular}
    \label{tab:tactile}
    \begin{flushleft}
    \footnotesize{\scriptsize{$^*$ IHM - \textit{In-hand Manipulation}; \quad \ WBM - \textit{Whole-body Manipulation}.}}    
    \end{flushleft}
\vspace{-0.8cm}
\end{table*}

\subsection{Tactile Sensing on Hands}
Tactile sensors on dexterous hands provide contact information, addressing the challenges in object manipulation such as grasped object controllability and object property estimation. In this subsection, we survey studies that integrate tactile perception into the control, planning, and learning of complex manipulation tasks. Due to the similar nature of contact-rich interactions, the tactile sensing techniques on hands also offer valuable insights for whole-body tactile sensing and manipulation, which is discussed in Sec.~\ref{sec:whole-body-tactile}.

To achieve the grasping objectives, sensed contact forces serve as real-time feedback in force or impedance control loops to regulate the desired object behavior~\cite{li2013control}. Moreover, slip detection and prediction based on tactile sensor data are used to adapt grasping forces, thereby enhancing grasp stability~\cite{song2013efficient,veiga2020grip,yuan2017gelsight}. 


More complex in-hand manipulation tasks demand interactive perception beyond static object models. Dynamic contact information, including real-time tracking of object states, monitoring contact stability~\cite{hogan2020tactile}, and predicting interaction outcomes~\cite{tian2019manipulation,lambeta2020digit}, \textit{i.e.}, how contact forces affect the balance of both objects and the robot, are crucial to achieving complex interactive behaviors. 
However, due to the inherent complexity of multi-contact dynamics and increased dimensionality of the contact state space, model-based methods still struggle to match human-level dexterity and versatility in multi-finger manipulation.

Alternatively, model-free Reinforcement Learning (RL) has shown the ability to address complex contact interactions. These approaches integrate tactile measurements directly into the state space to train end-to-end policies~\cite{van2015learning, melnik2021using}. 
Other learning methods besides task-specific RL have been sought for more generalizable policies.~\cite{lin2024learning} employs diffusion policy to achieve complex and long-horizon bi-manual manipulation tasks, while recent work has integrated tactile sensing into foundation models alongside vision and language~\cite{fu2024touch, Yu-RSS-24}. Though limited to simple control tasks, these models may eventually enable more natural and versatile physical interactions in humanoid robots.

Advancing robot hands with tactile sensing for humanoid tasks requires addressing the dual demands of high dexterity for delicate manipulation and high payload capacity for heavy object lifting. While human hands naturally achieve this balance, most robotic hands prioritize dexterity but support limited payloads. 
In the short term, swappable modular hands tailored to specific tasks are practical, but the long-term goal should be a unified hand combining both capabilities. A promising approach involves multimodal sensing modules, integrating sensors optimized for different force ranges and resolutions. 
Progress in sensor design, material science, sensor fusion, and high-fidelity simulation is critical to this effort.

\subsection{Tactile Sensing on Feet}
Other than manipulation, tactile sensing has started to gain {traction} for locomotion problems. For legged locomotion, estimation of Ground Reaction Forces (GRFs) and terrain properties is critical for maintaining whole-body stability on diverse, uneven surfaces. While vision and proprioception sensors can provide an indirect estimation of the terrain, these sensing modules lack the capability of accurately estimating GRFs and various terrain properties. 
Tactile sensing on foot soles has the potential to provide direct, unobstructed, and accurate contact measurements, but {remains} largely underexplored. 

To measure GRFs, existing works use Force/Torque sensors mounted on ankles~\cite{Honda_Humanoid, Dafarra_2024} or load-cell sensors for point-wise measurement~\cite{Kuindersma2016}. However, such methods inform only the zero moment point and lack accurate information on contact patch location, force distribution, and detailed terrain properties. To obtain such information, contact sensing arrays~\cite{suwanratchatamanee2009simple} and multimodal sensing suites \cite{xu2021flexible,xiaofeng_foot, TYLER2023Foot} have been integrated into legged robotic systems for diverse contact information.

To date, tactile sensors for legged systems have been primarily developed for monopods, quadrupeds, and hexapods.  \revised{Only a few studies have built tactile feet sensors for humanoid robots, with applications such as terrain classification~\cite{xiaofeng_foot} and ground slope recognition~\cite{suwanratchatamanee2009simple}. The sensed tactile information should aid the humanoid control and enhance the locomotion performance. A notable work in this direction~\cite{guadarrama2018enhancing} reconstructs the pressure shape of the foothold, enabling the recognition of uneven terrain and footstep replanning in real time.}

Building tactile sensors for humanoid feet is more challenging due to larger impulse and shear force during intermittent ground contact caused by fewer legs and heavier robot weight. Another challenge lies in developing robust and reliable sensors capable of withstanding various terrains, prompting researchers to seek durable materials and dependable mechanical designs. In addition, humanoids have stricter requirements for system integration. For example, the computing and power units of an adult-size humanoid robot are potentially more distal from the foot.

To enable robust humanoid locomotion in the wild, future directions for tactile sensing of feet need to address the following challenges: (i) how to accurately estimate more terrain properties such as stiffness, damping, plasticity, heterogeneity, and porosity; (ii) what are the appropriate metrics to measure the level of terrain complexity such as density, height, slickness, and roughness (\textit{e.g.}, size and wavelength of rocks in terrain), and the effect induced by weather and lighting conditions (\textit{e.g.}, rainy, snowy, sunny, night); and (iii) how to fuse terrain tactile sensing with other conventional sensing modules such as proprioception and visual perception to jointly inform postures, speeds, and gaits for intelligent, terrain-aware locomotion.

\subsection{Whole-body Tactile Sensing} 
\label{sec:whole-body-tactile}
Whole-body tactile sensing extends the aforementioned single-body sensing to all parts of the body, enabling humanoid robots to interact with unknown environments not only by the fingertips or foot soles but also by the arms, legs, and torso.

With explicit tactile feedback, humanoid robots such as iCub and REEM-C have achieved whole-body compliance~\cite{nori2015icub, dean2019whole}, controlling the contact force from whole-body regions. This level of contact awareness facilitates safe and intuitive physical human-robot interactions including dancing with human~\cite{tactile_dance}. Contact awareness is also useful for improving balance and collision avoidance in unstructured environments. 

Large-area tactile sensing significantly enhances a robot's ability to handle \textit{large} objects, including object identification through tactile exploration and whole-body manipulation. For example,~\cite{kaboli2018robust} enables a humanoid robot NAO, covered with artificial skin over its entire upper body, to classify large, heavy objects with different weights and textures.~\cite{jain2013reaching} demonstrates whole-arm tactile sensing by reaching objects in cluttered spaces while regulating contact forces across its arms. Close-proximity whole-body capacitive sensing is implemented in~\cite{AirBus}, enabling a cobotic humanoid with workers close-presence awareness. The same technology is used to draw semantics in human-humanoid physical interaction in~\cite{wong2022thmc}. Moving away from traditional methods~\cite{collision_free_walking_TMECH} that prioritize collision-avoiding trajectories,~\cite{armleder2024tactile} uses tactile feedback to detect and clear movable obstacles, thereby solving the problem of navigation among movable objects. Additionally, with tactile sensors covering their arms and chest, humanoid robots HRP-2 and Punyo-1 can use their entire upper body to grasp and lift large, heavy boxes~\cite{ohmura2007humanoid, WBM_tactile_gordon} or various household items~\cite{Punyo}.



\subsection{Challenges in Tactile Sensing}

\revised{Current works of tactile-based manipulation, whether in-hand or whole-body, are still largely limited to grasping or simple motions like rolling, pushing, or pick-and-place. Although tactile sensing holds great promise, its integration into more dexterous manipulation or loco-manipulation tasks that involve more dynamic interactions and contact shifting over the full body continues to pose substantial challenges. These include understanding the complex multi-contact dynamics, handling the high dimensionality of the sensor data, and addressing the sim-to-real gap.} 

\solution{Several approaches have been explored to address these difficulties. For instance, dimensionality reduction techniques such as spectral clustering, principal component analysis~\cite{chebotar2014learning}, and autoencoders~\cite{van2016stable} have been employed to manage the high-dimensional input space. Others have proposed simplifying sensor outputs by using boolean contact states (i.e., contact or no-contact) instead of continuous force values in RL training~\cite{melnik2021using}. Furthermore, advances in tactile simulation~\cite{wang2022tacto, lin2022tactile} are making simulated tactile data more accessible, thereby supporting zero-shot sim-to-real transfer~\cite{akinola2024tacsl}.}

One of the remaining fundamental challenges is to dynamically reason about contacts. This involves not only estimating contact states and static object models but, more crucially, understanding how these contacts and changes of contacts impact the system in real time, including the balance of both the robot and the object. Such information is vital for a planner to make informed decisions and, in a learning framework, can enhance sample efficiency. \revised{Addressing this challenge calls for deeper collaboration between the sensing and control communities to determine what tactile information should be abstracted and how to tightly integrate real-time tactile perception into control and planning for humanoid tasks.}



\revised{\textit{Key Takeaways}: Tactile sensing is yet an underexplored modality for advancing humanoid loco-manipulation, providing direct contact information necessary for tasks involving complex interactions with environments and objects. While it has improved task performance, achieving human-level dexterity and versatility requires advances in dynamic perception, and multi-modal sensing integration to enable systematic, real-time decision-making during interactions. Future directions include optimizing whole-body contact scheduling based on object properties and understanding how contact dynamics affect robot and object balance during loco-manipulation. Moreover, the design of whole-body tactile systems should account for varying sensor resolutions and load requirements, \textit{i.e.} hands need higher resolution for delicate tasks, while body skin can operate at lower resolution but withstand higher payloads.
For further reading, we recommend a survey paper on humanoid tactile sensing~\cite{dahiya2009tactile}, and a book chapter on tactile sensing technologies with an emphasis on deployment on humanoid robot~\cite{natale2017tactile}.}

\section{Multi-Contact Planning for Humanoids}
\label{sec:contact_planning}

\begin{figure}[!t]
    \centering
    \includegraphics[clip, trim=0.95cm 0.5cm 0.7cm 0.5cm, width=\columnwidth]{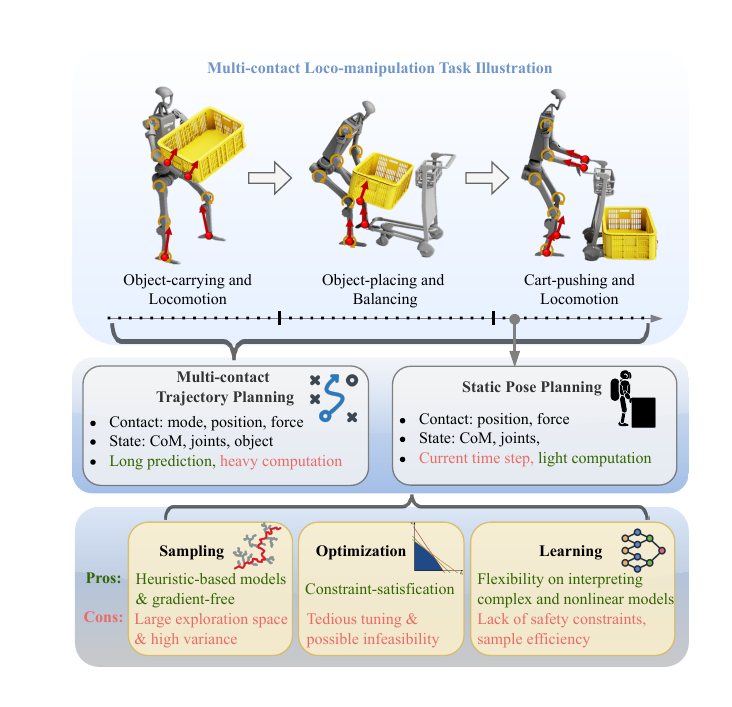}\vspace{-0.2cm}
    \caption{An illustration of a task sequence for loco-manipulation planning in humanoid robots, involving carrying and placing a box and pushing a cart. The planning techniques explored include (a) multi-contact trajectory planning and (b) whole-body pose planning, highlighting their contact and state planning strategies. Additionally, the pros and cons of categorized approaches in (i) sampling-based, (ii) optimization-based, and (iii) learning-based methods are summarized.}
    \label{fig:multi-contact}
    \vspace{-0.1in}
\end{figure}
Multi-contact planning remains one of the most challenging tasks in robotics. Specifically, in the context of humanoid whole-body loco-manipulation, a planner ought to solve trajectories that handle rich and intricate interactions with environments or objects. 
Particularly, beyond robot state trajectories, the planner is also expected to decide \textit{contact position} (or \textit{contact location}), \textit{contact mode}, and \textit{contact force} in a loco-manipulation task. Given the underactuated nature of humanoid robots and the addition of manipulation interaction dynamics, maintaining balance and manipulating objects rely solely on these contact interactions. This already makes multi-contact planning a challenging problem. Moreover, the diverse physical properties of environments and objects (\textit{e.g.}, rigid or soft, fixed or movable) complicate the problem even further. 

Over the past decade, the field has produced fruitful results in multi-contact humanoid planning, demonstrating promising potential across various locomotion and manipulation tasks~\cite{lengagne2013generation, ponton2016convex, henze2016passivity, bolotnikova2020ral}. However, these works require pre-planned contact mode sequences before planning robot whole-body motion trajectories~\cite{Bouyarmane2019}. This raises an open problem: how to solve the locomotion and manipulation contact planning problem simultaneously with the whole-body trajectory planning in a unified fashion, \textit{a.k.a}, Contact-Implicit Planning (CIP)~\cite{posa2014direct,mordatch2012discovery}. The primary challenge of this CIP lies in its high computational burden and combinatorial complexity of identifying potential contact mode sequences. 
Therefore, selecting suitable approaches depends on the specific problem requirements, including factors such as solving time, numerical robustness of the solution, resolution of the solution, and dependency on numerical models.

State-of-the-art procedures select planning algorithms for the underactuated system with three main categories: (i) searching, (ii) optimization, and (iii) learning, as illustrated in Fig.~\ref{fig:multi-contact}.

\subsection{Search-based Contact Planning} 
Search-based approaches employ state expansion that allows exploring configurations to make and break contacts; collisions and kinematic feasibility are often checked during each search step. Heuristics can be applied in search-based methods for efficient exploration. The search result is an optimal sequence of contact modes that ensure stability and task efficiency. Whole-body motions can be optimized during the search to verify the dynamic feasibility of candidate contact sequences~\cite{zhu2023efficient} or after the search in a \textit{contact-before-motion} style~\cite{contact-before-motion}. 
Search-based methods are commonly used for gait planning in legged robot locomotion~\cite{ferrari2019integrated, amatucci2022monte, taouil2024non, ravi2024efficient}. Expanding their capabilities in more intricate multi-contact loco-manipulation planning,~\cite{Murooka_Search} seeks to implement a graph search method for humanoid grasp contact planning and replanning, and~\cite{escande2013planning} introduces a contact-before-motion planner for multi-contact behaviors. 

However, search-based methods usually struggle to cover the entire exploration space in a limited time budget for online planning and may result in solutions with high variance. To tackle this,~\cite{janson2017monte} incorporates control variate and importance sampling as statistical variance-reduction techniques for faster solution convergence.~\cite{li2021mpc} avoids the time-consuming re-planning by incorporating only forward path expansion with informed possible paths to achieve reliable online kinodynamic motion planning.
\begin{figure}[!t]
    \vspace{0.2cm}
    \centering\includegraphics[width=0.65\columnwidth]{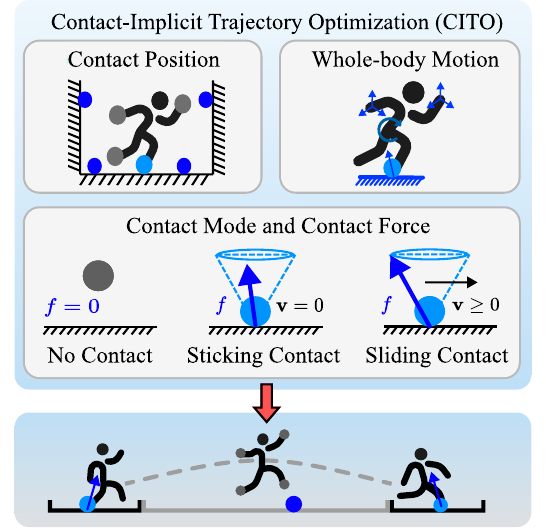}
    \caption{An illustration of Contact-Implicit Trajectory Optimization (CITO) that simultaneously plans contact mode, contact position, contact force, and whole-body motion. However, solving CITO problems online for humanoid loco-manipulation tasks still poses a challenge.}
    \label{fig:CITO}
    \vspace{-0.1in}
\end{figure}

Furthermore, the feasibility guarantee of the results via search-based methods can be made through Pose Optimization (PO), a subset of multi-contact planning in humanoid loco-manipulation. This holds true when the contact locations, timings, and manner of interaction are predetermined—such as in scenarios where a humanoid safely assists a person~\cite{bolotnikova2020ral} with feasible contact locations through point cloud processing. PO focuses on leveraging optimization-based techniques to plan whole-body poses and kinematic configurations at specific time steps, given a predefined contact mode. While PO is limited to handling discrete keyframes and does not account for continuous dynamics, this makes it highly suitable as a subsequent pose generator for gradient-free multi-contact planners, reducing the kinematic computation load during the search process. 
Furthermore, task-oriented objectives can be incorporated during PO, such as to maximize the interaction force~\cite{brossette2018tro,TORO_push} and to efficiently retarget the operator's motion into safe and feasible robot poses~\cite{rouxel2022multicontact}. 
Given a nominal pushing pose,~\citet{force_dist_optim} optimizes the distribution of reaction forces among all contacts to guarantee the friction constraints in heavy object pushing. Kinematics-and-mass-model-based posture generator is employed on HRP-4 humanoid to leverage leaning pose and body contacts to improve force in a heavy object pushing task~\cite{Murooka_heavy_push}.
A kino-dynamics-based PO approach is used in generating optimal humanoid pushing poses for dynamic non-prehensile loco-manipulation~\cite{Li_pose_optim}.
Search-based multi-contact planning and PO are often paired with online whole-body control that effectively tracks the optimal pose while adaptively interacting with the environment and objects. We detail the whole-body control strategies in Sec.~\ref{sec:control}.

\subsection{Optimization-based Contact Planning} 
Unlike search-based contact planning, which primarily checks kinematic feasibility for expansion and often requires additional lower-level planning to generate dynamically feasible motion, optimization-based contact planning~\cite{posa2014direct} offers the possibility of \textit{simultaneous} planning of whole-body motion and contact interactions, as illustrated in Fig.~\ref{fig:CITO}. This approach integrates dynamics directly into the contact planning process, eliminating the need for a hierarchical structure. A Contact-Implicit Trajectory Optimization (CITO) is formed by incorporating contact dynamics into the 
trajectory optimization formulation, 
allowing the solver to determine the contact modes, contact forces, contact positions, and whole-body motions all at once~\cite{chatzinikolaidis2020contact,manchester2020variational, posa2016optimization}. 

Due to the inherently large problem size, CITOs often rely on speed-up strategies, such as warm-starting with reasonable initial guesses for fast convergence~\cite{mastalli2016hierarchical} and separating into contact planning and whole-body motion planning subproblems in a hierarchical fashion~\cite{chen2023online}.  
With the increasing demand for computation efficiency, CITOs have witnessed a rise in computation speed via sequential quadratic programming (SQP) (\textit{e.g.},~\cite{posa2014direct}), differential dynamic programming (DDP)(\textit{e.g.},~\cite{tassa2012synthesis}), and iterative linear quadratic regulator (\textit{e.g.},~\cite{carius2018trajectory}). 
These improvements have even enabled the use of CITO in a Model Predictive Control (\textit{a.k.a.} CI-MPC) framework for real-time planning on quadruped robots~\cite{le2024fast,kong2023hybrid} and robotic arms~\cite{aydinoglu2023consensus, kurtz2023inverse}. However, for the humanoid robots, applying CITO to loco-manipulation has yet to be achieved.  


\revised{Migrating such CITO as real-time CI-MPC to humanoid loco-manipulation presents its own set of challenges, including high-dimensional space of optimization variables, complex/undifferentiable contact dynamics models, proper modeling of interaction dynamics, resolution of initial guess, and tedious tuning. \citet{esteban2025reduced} has made an initial effort to provide reasonable reference trajectories with the Hybrid Linear Inverted Pendulum (HLIP) model to allow natural contact behaviors at both hand and foot locations. }



\subsection{Learning-based Contact Planning} 
Learning-based approaches have demonstrated promising potential in planning multi-contact tasks, such as using reinforcement learning to generate velocity commands and contact sequences for quadruped locomotion~\cite{pmlr-v164-margolis22a, tsounis2020deepgait}. These learning-based planners are mostly modular, making it possible to form a hierarchical architecture with model-based planners and controllers at the low level. 
Compared with traditional optimization-based or heuristics-based approaches, learning-based elements enhance the computation efficiency in multi-contact planning. For example,~\cite{lin2019efficient} \revised{learns to predict the evolution of} centroidal dynamics and contact sequence \revised{for dynamic humanoid loco-manipulation} in under $0.1$~s, achieving a computation speed boost $300$ times faster compared to traditional optimization-based methods.

In addition, learning-based approaches can assist contact prediction, which allows additional information for contact (re)planning in real time.
Precise contacts are often hard to obtain from motion capture data, making it challenging to learn directly from data. To synthesize plausible motion, a naive supervised learning approach often leads to objects moving without any contact or significant penetration between the predicted human body and the objects. \cite{interdiff} introduces contact correction and predicts motions relative to the contacts predicted \revised{in human-object interaction tasks}. \cite{object_guided_motion_tog2023} separates the contact prediction and whole-body motion prediction by first predicting the contact positions with a large, dynamic object, which are then used as constraints to synthesize whole-body human motion. These models have the potential to serve as a loco-manipulation planner for humanoid robots. \cite{bahl2023affordances} learns to find from the video scenes the affordance, \textit{i.e.}, potential contact points, for \revised{household object manipulation tasks}. These contacts can be used as heuristics for subsequent motion planning.


\subsection{Challenges in Multi-contact Planning}

\revised{Despite the advancements in multi-contact planning, several key challenges remain. First, computational efficiency remains a fundamental bottleneck. Contact planning inherently involves a combinatorial explosion in possible contact mode sequences, which poses major scalability issues. Optimization-based methods, such as CITO, often suffer from local minima and sensitivity to initialization. Search-based methods, though capable of global exploration, are computationally expensive and unsuitable for online replanning. Learning-based approaches, though computationally efficient, struggle with adaptivity and precision in dynamic contact-rich control. \solution{A promising direction lies in combining these paradigms to balance global reasoning, dynamic feasibility, and real-time performance.} Second, current approaches rely on simplified point contact models, which fail to capture the complexity of real-world loco-manipulation. In practice, especially with full-body contact and manipulation, robots frequently engage in patch contacts that are complex to model with model-based control and planning frameworks. \solution{Extending contact modeling beyond point approximations is essential to better align with real physical interactions.} Third, most existing methods work around fixed contact mode sequences, which diverge from the unpredictable and rapidly changing nature of real-world contact interactions. Unexpected contact events, which are often hard to observe due to visual occlusion and lack of whole-body tactile sensing, pose serious challenges for accurate contact estimation, control, and online planning. This raises a fundamental question: is precise contact planning always necessary? \solution{To enable more adaptable loco-manipulation behaviors in uncertain environments, future research could explore planning, control, and learning strategies that are robust or even agnostic to small, transient changes in contact.}}

\revised{\textit{Key Takeaways:}  While significant progress has been made in humanoid multi-contact planning, future work should focus on developing more integrated approaches that combine the strengths of search-based, optimization-based, and learning-based methods. Specifically, addressing the computational complexity of CIP and improving real-time performance will be key. \solution{Future directions could explore hybrid approaches that incorporate efficient contact sequence generation/contact dynamics, apply contact-implicit constraints in real-time, and achieve learning-based contact prediction to enhance robustness and adaptability in complex loco-manipulation tasks.}
The readers are recommended to further read the survey on humanoid multi-contact planning~\cite{Bouyarmane2019}. }

\section{Model Predictive Control for Loco-manipulation}
\label{sec:planning}

Optimization-based Model Predictive Control (MPC) has advanced significantly in robotics. The advantages of its flexibility to define versatile motion objectives, rigorous mathematical formulations, and widely available solvers establish MPC as one of the most popular approaches to trajectory planning for locomotion and manipulation.

A uniform optimization formulation of the loco-manipulation planning problem seeks an optimal state trajectory and control input over a finite horizon in the future. MPC is often formulated as an Optimal Control Problem (OCP):
\begin{align}
\allowdisplaybreaks
\min_{\boldsymbol{x}(\cdot), \boldsymbol{u}(\cdot), \boldsymbol{\lambda}(\cdot)} \;\; &  \mathcal{L}(\boldsymbol{x}(\cdot), \boldsymbol{u}(\cdot), \boldsymbol{\lambda}(\cdot)) \label{eq:cost} \\
\textrm{s.t.} \quad & \boldsymbol{\dot{x}}_{k}  = f(\boldsymbol{ {x}}_{k} , \boldsymbol{{u}}_{k} , \boldsymbol{\lambda}_{k} ) \label{eq:nonlinear_dynamics}\\%
& h_{\rm task}(\boldsymbol{{x}}_{k}, \boldsymbol{\lambda}_{k}, \boldsymbol{u}_{k}) = 0 \label{eq:contact} \\
& g_{\rm task}(\boldsymbol{{x}}_{k}, \boldsymbol{\lambda}_{k}, \boldsymbol{u}_{k}) \ge 0  \label{eq:weighted_task}
\end{align}
where $\boldsymbol{x}(\cdot), \boldsymbol{u}(\cdot), \boldsymbol{\lambda}(\cdot)$ are the trajectories of the states, control inputs, and constraint forces, respectively. 
$\mathcal{L}(\cdot)$ is the cost function. The dynamics is represented in (\ref{eq:nonlinear_dynamics}). $h_{\rm task}$ and $g_{\rm task}$ are other tasks represented as equality and inequality constraints. $h_{\rm task}$ are holonomic constraint tasks to be enforced strictly (\emph{e.g.}, a contact-explicit formulation~\cite{chignoli2021humanoid}), and $g_{\rm task}$ are unilateral constraints to encode set-valued tasks (\emph{e.g.}, joint limits, non-sliding contact with fiction cones, etc.). 

Depending on the choice of dynamics models (\ref{eq:nonlinear_dynamics}), costs, and constraints, the OCP formulation is commonly transformed as a linear convex MPC (\textit{e.g.}, \cite{chignoli2021humanoid,li2023dynamic}) or a Nonlinear MPC (NMPC) (\textit{e.g.}, \cite{ romualdi2022online, elobaid2023online}). 
Table~\ref{tab:MPC} summarizes recent MPC-based works on humanoid robots in loco-manipulation tasks. 

\begin{figure}[!t]
    \centering
    \includegraphics[clip, trim=0.5cm 0.8cm 0.5cm 0.5cm, width=\columnwidth]{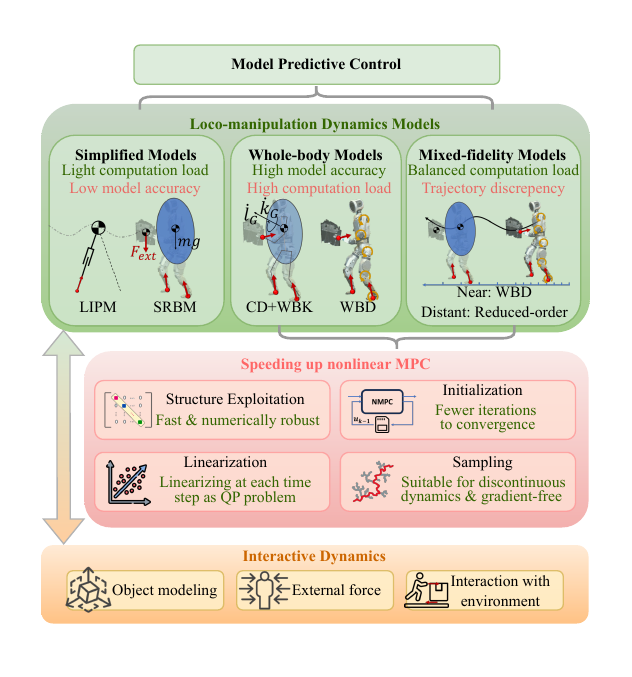}
    \vspace{-0.1in}
    \caption{An illustration of model predictive control in humanoid robotics, showcasing three primary categories of dynamics modeling in loco-manipulation: (i) simplified dynamics, (ii) nonlinear dynamics, and (iii) Mixed-fidelity dynamics. The figure also highlights the consideration of interactive dynamics modeling with environments and/or objects for loco-manipulation tasks. Additionally, four common approaches to speeding up/simplifying NMPC are summarized.}
    \label{fig:MPC_structure}
    \vspace{-0.1in}
\end{figure}

\subsection{Simplified Models}
\label{subsec:simplified_model}
In pursuit of high-frequency online planning with lightweight computation for motion control, simplified dynamics models, or reduced-order models (ROMs) are often employed in MPC. For example, the Single Rigid Body Model (SRBM) can be linearized by providing explicit foot position sequence reference and be formulated in a linear convex MPC~\cite{li2021force}. Using SRBM,~\cite{chignoli2021humanoid} realizes dynamic aerobatic behaviors on the MIT humanoid. Extending to humanoid loco-manipulation,~\cite{li2023dynamic} achieves object-carrying and rough terrain locomotion by simplifying the interaction dynamics as external gravitational forces applied to robot CoM. 

\begin{table*}[!t]
\centering
\setlength{\tabcolsep}{5.5pt}
    \renewcommand{\arraystretch}{1.1}
\caption{Recent MPC Approaches on Humanoid Loco-manipulation}
\begin{scriptsize}        
\begin{tabular}{p{0.04\textwidth}p{0.04\textwidth}p{0.05\textwidth}p{0.25\textwidth}p{0.30\textwidth}p{0.08\textwidth}p{0.08\textwidth}}
\hline \hline
 \makecell[l]{Paper} & \makecell[l]{\revised{Year}} &  \makecell[l]{Robot \\Model$^*$} & \makecell[l]{Interaction Modeling Method} & {Locomotion (\textit{L}) and Manipulation (\textit{M}) Planning} & \makecell[l]{MPC \\ Frequency} & \makecell[l]{Solving \\Method} \\
\hline  \hline 
 \cite{henze2014posture}& \revised{2014} & SRBM & Optimizing external wrench(es) at contact(s) & Unified & $20$ Hz & QP \\ \hline
 \cite{li2023dynamic}& \revised{2023} & SRBM & Predefined external force & Unified & $300$ Hz & QP \\ \hline
 \cite{bang2024rpc}& \revised{2024} & SRBM & Negligible object dynamics & Separated: \textit{L}: MPC; \textit{M}: Keyframe interpolation & $-$ & QP \\ \hline
 \cite{penco2019multimode}& \revised{2019} & LIPM & Negligible object dynamics & Separated: \textit{L}: MPC; \textit{M}: Teleoperation and retargeting & $-$ & QP \\ 
\hline 
 \cite{Kuindersma2016}& \revised{2016} & CD & Optimizing external wrench(es) at contact(s) & Unified &  offline &  SQP \\ \hline
 \cite{agravante2019human}& \revised{2019} & CD & Predefined external wrench  & Unified &  $-$ & QP \\ \hline
 \cite{Dai_centroidal_kinematics}& \revised{2014} & CD & Optimizing external wrench(es) at contact(s) & Unified &  $5$ Hz/offline &  SQP \\ \hline
 \cite{polverini2020multi}& \revised{2020} & CD & Optimizing external wrench(es) at contact(s) & Unified &  $10$ Hz &  Interior-point \\ \hline
 \cite{elobaid2023online}& \revised{2023} & CD & Estimated as external wrench through sensors & Unified &  $5$ Hz &  Interior-point \\ \hline
 \cite{Memmo_2021}& \revised{2021} & WBD & Optimizing external wrench(es) at contact(s) & Unified & $100$ Hz & DDP \\ \hline \hline
\label{tab:MPC}
\end{tabular}
\end{scriptsize}    
\vspace{-0.6cm}
\begin{flushleft}
\footnotesize{\scriptsize{$^*$ SRBM - \textit{Single rigid-body model}; \quad \ LIPM - \textit{Linear inverted pendulum model}; \quad CD - \textit{Centroidal dynamics}; \quad  WBD - \textit{Whole-body dynamics}.}}    
\end{flushleft}
\vspace{-0.6cm}
\end{table*}


On the other hand, the Linear Inverted Pendulum Model (LIPM) has been a popular choice as a linearized dynamics model for humanoid locomotion~\cite{Kajita2003} and in multi-contact~\cite{audren2014iros}. 
Extending LIPM to loco-manipulation tasks is achieved through teleoperation~\cite{penco2019multimode}. However, such \revised{a} model inherently lacks the ability to address contact interactions and loco-manipulation dynamics, necessitating a lower-level whole-body control for balancing and manipulation tasks.

\subsection{Whole-body Models} 
\label{subsec:WB_model}
While simplified dynamics models offer computation efficiency, the simplification assumptions cause a detriment to the model's
accuracy and limit the capability of whole-body motion planning. Conversely, whole-body models are more accurate representations of the robot dynamics and are better suited for planning versatile motions and interactions with objects and the environment. NMPC comes into play when constraints or cost functions become nonlinear, for example, dynamics constraints formed by kinodynamics and Whole-Body Dynamics (WBD). 

In the context of humanoid motion planning, kinodynamic constraints are often referred to as the combination of Centroidal Dynamics (CD) and whole-body kinematics (WBK) constraints~\cite{Dai_centroidal_kinematics}, where CD is derived from the total momenta of the system, and captures the effect of full-body inertia of a multi-linkage dynamics system~\cite{orin2013centroidal}. 
For example, achieving consensus between CD and full-body kinematics in one Trajectory Optimization (TO) generates versatile humanoid motions~\cite{Dai_centroidal_kinematics}.


\revised{Leveraging joint-space WBD (eqn. (\ref{eq:equation-of-motion})) in MPC has gained popularity, particularly for free-floating articulated robots like humanoids. While WBD allows flexible contact modeling, its high nonlinearity and nonconvexity impose significant computational burdens on WBD-based Nonlinear Programs (NLP) or NMPC, making real-time planning challenging. 
This issue is especially critical for high-DOF humanoids in loco-manipulation tasks, such as payload transport, which require additional modeling of object dynamics and safety constraints. Thus, this section surveys methods to accelerate NMPC while preserving solution accuracy.}

\subsection{Mixed-fidelity Models}
\label{subsec:MF_model}
Instead of using full joint-space dynamics across the entire horizon of an MPC, mixing multiple models of varying abstraction levels demonstrates improved performance and efficiency. 

Cascaded-fidelity models (\textit{a.k.a.} hierarchical dynamics) use different models to govern segments of the horizon~\cite{li2021model, Wang_Multi-Fidelity, li2024adapting}. 
These methods typically employ high-fidelity (\textit{e.g.}, full-order) models in near horizons and low-fidelity (\textit{e.g.}, simple) models for distant horizons, thus maintaining the solution accuracy in near horizons while solving the myopic issue by allowing a longer horizon using simple models. This approach could be suitable in loco-manipulation tasks as it would either simplify interaction dynamics as simple external forces or impose the object dynamics as part of CD in the far horizons to allow improved real-time computation compared to full dynamics models.

Another method is to overlap different dynamics models across their horizons. In such cases, achieving a consensus between these overlapped models is necessary. To solve problems with such mixed-fidelity models,~\cite{Budhiraja_Consensus} decomposes a single TO that incorporates both dynamics into two subproblems and then alternates between the two to achieve consensus. Similarly,~\cite{herzog2016structured} alternates between the CD and WBK subproblems. Overall, model simplification over MPC horizons will remain an effective approach~\cite{Wensing_survey_TRO}. On the other hand, mixed-fidelity models demonstrate superior capability but require careful consideration of combined models.

\begin{table}[!t]
\centering
\setlength{\tabcolsep}{2.0pt}
\caption{\revised{NMPC Speed-up Strategies: \\Accuracy, Efficiency, and Limitations}}
\begin{scriptsize}        
\begin{tabular}{p{0.135\textwidth}p{0.055\textwidth}p{0.06\textwidth}p{0.20\textwidth}}
\hline \hline
 \makecell[l]{\quad Method} & \makecell[l]{Accuracy} &  \makecell[l]{Efficiency} & \makecell[l]{Limitation and Challenge} \\
\hline  \hline 
\multicolumn{4}{ l }{ \makecell[l]{\textit{i. Dynamics Modeling}} } \\ \arrayrulecolor{black!30}\hline 
 \makecell[l]{\quad Simplified model \\ \quad (Sec. \ref{subsec:simplified_model})} &  \makecell[c]{low} &  \makecell[c]{high} & \makecell[l]{Possible infeasible motion \\ trajectory due to simplification} \\ 
 \hline 
  \makecell[l]{\quad Whole-body model \\ \quad (Sec. \ref{subsec:WB_model})} &  \makecell[c]{very high} &  \makecell[c]{very low} & \makecell[l]{Requires very accurate dynamics \\ modeling of robot and object}  \\
 \hline 
 \makecell[l]{\quad Mixed-fidelity model \\ \quad (Sec. \ref{subsec:MF_model})} &  \makecell[c]{medium} &  \makecell[c]{medium} & \makecell[l]{Motion trajectory discrepancy \\ due to model mismatch} \\
 \arrayrulecolor{black}\hline  
 \multicolumn{4}{ l }{ \textit{ii. NLP Structure Exploitation}}  \\ \arrayrulecolor{black!30}\hline
\multicolumn{4}{ l }{\textbf{\quad Backward recursion:}} \\ 
 \quad DDP \cite{original_DDP} &  \makecell[c]{medium} &  \makecell[c]{high} &  \\ 
 \quad iLQR \cite{Tassa_iLQR} &  \makecell[c]{low} &  \makecell[c]{very high} & \vspace{-0.4cm}  Poor constraint-handling \\\hline  
 \multicolumn{4}{ l }{\textbf{\quad Derivative calculation:}} \\
 \quad Auto-Diff \cite{rall1981automatic} &  \makecell[c]{very high} & \makecell[c]{low}  &   \\
 \quad Hypergraph \cite{rosmann2018exploiting}&  \makecell[c]{high} &  \makecell[c]{medium} &  \vspace{-0.4cm}May lead to high memory usage \\\hline
 \textbf{\quad Exploring sparsity:} \\
 \quad FATROP \cite{vanroye2023fatrop} &  \makecell[c]{very high} &  \makecell[c]{high} &  Requires specific problem structure \\ 
 \quad AdaptiveNLP \cite{callens2024adaptivenlp} & \makecell[c]{high}  & \makecell[c]{high}  & Less effective in small-scale NLPs  \\
 \arrayrulecolor{black}\hline 
 \multicolumn{4}{ l }{ \makecell[l]{\textit{iii. Linearization}} } \\ \arrayrulecolor{black!30}\hline 
 \quad SQP \cite{wilsonSQP} &  \makecell[c]{high} &  \makecell[c]{medium} &  Scalability to high-order \\\hline
 \makecell[l]{\quad Successive \\ \quad Linearization \cite{zhakatayev_successive_2017}} & \makecell[c]{medium}  & \makecell[c]{high}  & Requires good initial guess \\ \hline
 \textbf{\quad QP Speed-up:}\\ 
  \quad Condensed \cite{jerez2011condensed} &  \makecell[c]{very \makecell[c]{high}} &  \makecell[c]{medium} & No state trajectory output \\  
 \quad ReLU-QP \cite{bishop2023relu} & \makecell[c]{high}  & \makecell[c]{very high} & Only time-invariant states matrices  \\ 
 \arrayrulecolor{black} \hline 
 \multicolumn{4}{l }{ \makecell[l]{\textit{iv. Warm Start}} } \\ \arrayrulecolor{black!30}\hline 
 \quad Gait Library \cite{gait_library_IJRR} & \makecell[c]{high}  & \makecell[c]{very high}  & Constrained by onboard storage  \\\hline 
 \makecell[l]{\quad Memory of \\ \quad Motion \cite{Memmo_2021}} & \makecell[c]{high}  & \makecell[c]{low}  & Requires large dataset coverage  \\
 \arrayrulecolor{black}\hline \hline 
\label{tab:MPC_speedup}
\end{tabular}
\end{scriptsize}
\vspace{-0.6cm}
\begin{flushleft}
\footnotesize{\scriptsize{\:\:Performance scale (low to high): very low$\rightarrow$ low$\rightarrow$ medium$\rightarrow$ high $\rightarrow$ very high.}}    
\end{flushleft}
\vspace{-0.4cm}
\end{table}


\subsection{NMPC Speed-up}
\revised{In addition to tailoring dynamics modeling strategies for specific loco-manipulation setups, solving NMPC can be accelerated using several key methods. Below, we summarize these methods and provide Table \ref{tab:MPC_speedup} to compare their solution accuracy, real-time efficiency, and application limitations.}

\textit{NMPC Speed-up via Structure Exploitation:}
NMPC problems often involve complex dynamics and constraints that can be computationally intensive to solve. Exploiting the structure within these problems can significantly enhance their solvability and efficiency, such as extracting variables that directly interact with each other, identifying repetitive and symmetric structures, and arranging block-diagonal structures. 
One of the most common approaches to solving NMPC is direct methods, which transform the NMPC into a \revised{NLP} with the complexity of $O(N^3)$, where $N$ is the problem size \cite{Diehl2006}. 
Some direct methods, such as direct multiple shooting and direct collocation, result in sparse NLPs, whose computation complexity can be reduced to $O(N)$~\cite{Betts}.
Another approach to solve NMPC is single-shooting methods, such as DDP~\cite{original_DDP} and its variant, Iterative Linear Quadratic Regulator (iLQR)~\cite{Tassa_iLQR}, which only retains the first-order derivative approximation of dynamics and exhibits a linear increase in computation over the horizon~\cite{ WB_MPC_Dantec}. 
With proper exploitation of the sparsity structure through the hypergraph approach,~\cite{rosmann2018exploiting} shows improvement of the nonlinear solver in computation efficiency. 
\revised{Furthermore, AdaptiveNLP leverages the previous NLP structure to significantly reduce the overhead and update time for constructing the current NLP~\cite{callens2024adaptivenlp}, which is particularly suitable for humanoid robot NMPCs with static kinematics and actuation constraints.}

\textit{NMPC Speed-up via Linearization:} Another way to tackle the computational burdens of NMPC is through successive linearization, which involves linearization at every timestep around the nominal system state and control input. The linearized dynamics become piece-wise affine, which can be formulated in a large, sparse Quadratic Program (QP) and can be solved online~\cite{zhakatayev_successive_2017, ding_real-time_2019}. 
\revised{SQP solves NLP problems by iteratively linearizing them into QPs and computing search directions at each step. While breaking down an NLP into QPs is viable, it requires efficient and scalable QP solvers.
\revised{Aiming for high efficiency,} ReLU-QP~\cite{bishop2023relu}, a GPU-accelerated QP solver, has improved linear convex MPC real-time control frequency in high dimensional space balancing tasks from the original 206 Hz to up to 2600 Hz. }
\revised{While providing reasonably efficient real-time deployment, linearization techniques highly rely on linearization assumptions, which compromise model fidelity (accuracy), leading to motion errors compared to full nonlinear models. However, trading accuracy for speed is often preferred in humanoid robot control, as controllers may struggle to precisely track full-order trajectories by NMPC, making highly accurate solutions not practically beneficial. }

\textit{NMPC Speed-up via Warm Start:} The real-time requirement motivates many researchers to seek a more effective initialization. One simple yet effective approach is to warm start with the solution from the previous iteration. 
Another common approach is to offload the computation burden from online to offline, \textit{e.g.}, the \textit{gait library}~\cite{gait_library_IJRR}. It can be regarded as a specific type of warm-start technique and requires only a cheap online interpolation among the gaits to obtain an approximately optimal full-body trajectory. 
Similarly,~\cite{Memmo_2021} uses \textit{memory of motions} to warm-start an MPC and overcome the sensitivity of initial conditions. 
However, the key challenge lies in the management of massive trajectories with limited storage. In Sec.~\ref{sec:skill_composition}, we discuss a solution from the learning community: learning compact models to distill from large-scale offline trajectories.

\textit{NMPC Speed-up via Sampling:} Real-time sampling-based planning, such as Model Predictive Path Integral (MPPI) control~\cite{williams2016aggressive}, is a simple and effective scheme. 
However, extending MPPI to high-dimensional loco-manipulation tasks presents significant computational challenges, especially within contact-implicit settings. Two primary techniques have enabled the recent success of MPPI: reducing the search space and leveraging parallelization in modern simulators. To limit the search space, researchers use suboptimal planners to guide the search, apply constraints, and employ spline control points to reduce the number of planning knot points~\cite{Carius_MPPI2022}. Furthermore, advancements in sampling speed also facilitate real-time planning. For instance, MuJuCo MPC (MJPC)~\cite{howell2022} leverages the established parallelization capabilities of MuJuCo~\cite{todorov2012mujoco} on multi-core CPUs. Additionally, modern simulators such as IssacLab~\cite{Orbit} and MuJuCo can roll out thousands of samples on GPUs, which allows additional randomization for robust control~\cite{turrisi2024benefits}. 

\begin{figure}[!t]
    \centering\includegraphics[width=0.99\columnwidth]{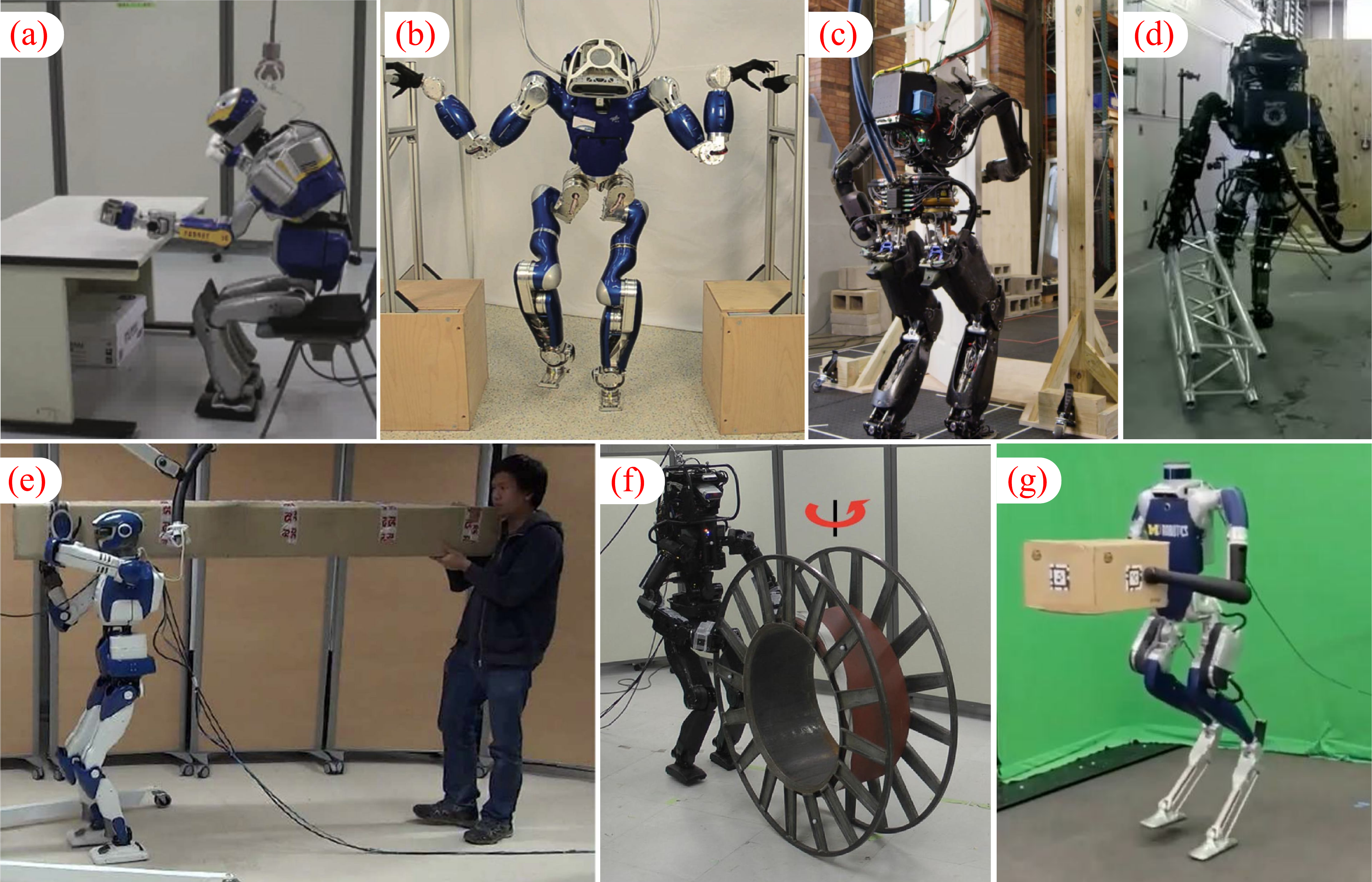}
    \caption{Loco-manipulation skills from model-based methods. (a) Getting up from a chair~\cite{lengagne2013generation}. (b) Multi-contact balancing~\cite{henze2016passivity}. (c) Door traversal~\cite{Nadia_door_auto}. (d) Transporting a bulky beam~\cite{Kuindersma2016}. (e) Collaborative carrying~\cite{agravante2019human}. (f) Rolling a bobbin~\cite{Murooka_Pattern}. (g) Box loco-manipulation~\cite{Fabrics_WBC}.}
    \label{fig:robot_loco_manip}
    \vspace{-0.1in}
\end{figure}

\subsection{Environment and Object Interaction Models for Loco-manipulation} 

In this subsection, we survey motion planning algorithms for loco-manipulation tasks that involve interactions with the environment and/or objects with large weights and sizes, specifically in the context of loco-manipulation MPC. Assuming the sequence of contact modes is defined through contact planning methods outlined in Sec.~\ref{sec:contact_planning}, the loco-manipulation MPC algorithms find a feasible trajectory that leads to a viable state over a horizon, while satisfying the dynamics constraint and contact stability constraints. 
The existing loco-manipulation MPC algorithms differentiate the interaction with a fixed environment and a manipulated object. 

\subsubsection{Interaction with a static environment} 
An environment, including static surfaces such as the ground and walls, provides contact forces that contribute to the robot's stability and enable interactive tasks such as walking and pushing. An example of a static environment is in Fig.~\ref{fig:robot_loco_manip} (a) and (b). Since the environment is static, the robot does not need to consider the environment's state or stability during planning. Instead, the robot is often required to deal with acyclic contact patterns and non-coplanar contacts given specific environment geometries. This challenging problem is referred to as multi-contact planning (MCP)~\cite{Bouyarmane2019, CarpentierTRO, ferrari2023multi}. MCP is a widely studied area that involves both contact planning and motion planning. Since contact planning has been discussed in Sec.~\ref{sec:contact_planning}, in this subsection, we focus on the motion planning aspect of MCP, specifically in terms of real-time multi-contact MPC. Given a sequence of contact modes, MCP aims to find feasible motion and contact wrenches of all contacts. 

Multi-contact MPC for humanoid robots can be solved by optimization-based methods~\cite{Dai_centroidal_kinematics, polverini2020multi}. Among these methods, \revised{CD} is the most common model due to its accurate representation of contact forces and the system's centroidal momentum. 
Despite the model’s accuracy, CD contains a nonlinear term derived from the cross-product between the state (CoM position) and control (contact wrench), posing a challenge to trajectory optimization. 
Using multi-contact MPC as a motion planning technique also has limited dynamic locomotion capabilities because it treats arms and legs uniformly as general contacts, rendering it less efficient at handling frequent contact switches compared to pure locomotion models such as the LIPM. 
Although MPC has the ability to plan contact with any surface of the robot, the regularization of the planned contact forces usually requires accurate joint torque sensing or whole-body tactile sensing (Sec.~\ref{sec:whole-body-tactile}), which still has significant space to explore and presents great potential for rich and safe environment interaction.

\subsubsection{Modeling interaction with a manipulated object} 
In the context of loco-manipulation MPC for humanoid robotics, modeling strategies for manipulated objects represent a crucial aspect and an area of ongoing research, alongside contact planning. An object can be a free-floating body (\textit{e.g.} a box), a fixed-base articulated mechanism (\textit{e.g.} a door or drawer), or actuated joints (\textit{e.g.} another robot)~\cite{Otani2017Retarget}, as shown in Fig.~\ref{fig:robot_loco_manip} (c-g). Unlike interactions with a static environment, the contact force exerted by an object depends not only on the robot's joint torque but also on the object's mass and inertia. 
As a result, interacting with objects in loco-manipulation tasks introduces significant complexity. Planning such tasks typically requires accurate knowledge of the object's state and physical properties, especially when handling heavy, irregular, or dynamically moving objects.
\revised{As a result, loco-manipulation tasks become significantly more complex, often requiring precise knowledge of an object's state and physical properties—especially when dealing with heavy, irregular, or dynamic objects.}


\revised{To address unknown object states and properties, adaptive control and online estimation techniques have been proposed to enhance robustness against dynamic effects and external loads. For example, \cite{Harada_HPR_Table_TMECH} compensates for residual dynamics without requiring predefined object parameters, while \cite{foster2024physically} estimates reflected inertia to handle changing loads. \citet{object_mass_estimation_iros2010} estimates object mass to optimize manipulation strategies for bulky items. \citet{chappellet2023humanoid} leverages wide-angle camera tracking to aid in large object handling. However, integrating these methods with MPC-based approaches introduces challenges, such as predicting object dynamics over preview horizons and managing increased computational load. \citet{Bang2024VariableIM} takes a step toward simplifying centroidal momentum evolution through supervised learning, preserving a convex CD-based MPC formulation for humanoid locomotion.}

Given the diversity of tasks, creating a unified model for the robot-object system is essential. 
We introduce two common approaches to incorporate object dynamics into the MPC-based planning process~\cite{audren2014iros}. 

The first approach models the manipulated object as external wrenches and plans the control input to compensate for them~\cite{Murooka_heavy_push, li2023dynamic, polverini2020multi}. This approach offers a flexible solution, as it integrates well with the MPC regardless of the linearity of the model, which treats all contacts as external wrenches. However, the contact wrench needs to be predefined, for example, to compensate for the gravity of the object or to exert a user-specified pushing force. Obtaining accurate contact wrenches for dynamic tasks like swinging a baseball bat is already inherently challenging, especially when considering their evolution over the entire prediction horizon in MPC. Static/quasi-static assumptions are usually made to neglect the dynamics of the object, resulting in less dynamic loco-manipulation motions. Another aspect to note is that contact wrenches can be applied at the contact location~\cite{Murooka_Pattern} or at the robot's CoM~\cite{agravante2019human, li2023dynamic}. In the former setting, the object affects the contact wrenches for both self-balance and object manipulation. In the latter setting, the object affects only the contact wrenches that are responsible for balance, and contact wrenches for object manipulation require additional regulation. For example, the loco-manipulation MPC approach in~\cite{li2023dynamic} adjusts the foot contact wrenches to the weight of an object applied to the robot CoM and additionally regulates the object position via hand contact wrenches with a separate controller. Unlike MCP, such loco-manipulation MPC prioritizes mobility over manipulation and typically employs specialized locomotion models, such as a linear inverted pendulum model (LIPM). These models introduce additional assumptions for bipedal locomotion, such as assigning foot contact for locomotion and hand contact for manipulation, maintaining body height, and conserving angular momentum, making them computationally efficient in an online MPC setting \revised{albeit being} less general.

The second planning approach incorporates the object's dynamics directly into the robot's dynamic equation of motion, creating a unified robot-object dynamic system~\cite{audren2014iros, Li_pose_optim}. This approach eliminates the static/quasi-static assumption from the first approach and leverages the time-varying robot-object dynamics in MPC to achieve more dynamic and adaptive loco-manipulation behaviors. In such planning problems, the interaction wrenches are usually treated as control variables, and contact stability constraint on the interaction wrenches is enforced to securely attach the object to the robot. The planner generates the combined motions for both the robot and the object, leading to their desired states. Compared to modeling objects as external wrenches, this method requires a perfect knowledge of the object's state, which is more challenging from the sensing perspective.

\subsubsection{Interaction with a dynamic environment or deformable objects} 
Dynamically changing environments, such as those with a moving surface~\cite{gao2024timevaryingfootplacementcontrolunderactuated} 
or with physical human interactions~\cite{agravante2019human}, introduce additional challenges to loco-manipulation planning and control. Similar to a dynamic object manipulation problem, the interaction model between the robot and the dynamic environment is also time-varying. Although one can infuse the dynamics of the object with the robot model to form unified dynamics, it is impractical to model the dynamics of the environment numerically in most cases. Therefore, in an MPC setting, the planner may require sensor feedback to predict the movement of the environment and replan the loco-manipulation motion adaptively~\cite{gao2024timevaryingfootplacementcontrolunderactuated}.
For example, interacting with an environment that involves humans requires the anticipation of human intentions for collaborative manipulation such as lifting payload~\cite{agravante2019human}; see a more challenging recent achievement in direct human-humanoid physical interaction~\cite{lefevre2025ral}. For such tasks, the interaction force is an important way of communicating intentions, which can be measured as force feedback signals to trigger robot movements. However, the evolution of such sensed forces can not be well-predicted beyond the current timestep for MPC to leverage,  suggesting further static/quasi-static assumptions are required.
Otherwise, the robot can only treat the dynamic environment as a disturbance and counteract it through reactive control (\textit{e.g.}, whole-body control). Given the challenge of dealing with the changing environment, loco-manipulation in a dynamic environment is largely unexplored.


In addition to rigid objects with regular geometries, deformable objects are ubiquitous in our real world, such as those in caregiving or housekeeping scenarios. Modeling the dynamics of these objects requires a deep understanding of their physical properties and behaviors, such as flexibility, elasticity, and deformation under force. Consequently, simplifications tailored to specific problems and applications are often necessary~\cite{zhu2018dual, saha2007manipulation}. For example, to plan the manipulation of a deformable belt,~\cite{qin2023dual} simplifies the motion of the belt by representing only its tail movement in a 2D plane. 
However, to fully exploit the object's deforming property for effective loco-manipulation, integrating accurate deformable objects~\cite{zhang2024adaptigraph} into robot models is essential. Although this area is relatively underexplored for humanoid loco-manipulation, such integration opens up significant opportunities beyond basic pick-and-place operations, enabling robots to tackle more intricate and delicate tasks. 

\revised{\textit{Key Takeaways for Model Predictive Control:} With the advanced capabilities of gradient-based numerical optimization in motion planning, MPC is gaining popularity in humanoid loco-manipulation, showcasing numerous variations in recent years of literature. The essence of this method lies in making reasonable choices regarding the dynamics model, constraint, task definitions, and computation requirements. These choices often require expert design and tuning to trade-offs among task versatility, solution feasibility, and optimality. By identifying the computation intensities and proper dynamics representation of the loco-manipulation tasks, one can offset the computation load by introducing simplified models and relaxed constraints in MPC. In addition, the MPC efficiency can greatly benefit from proper solver choices, an evolving area presenting opportunities for research on both solver-level and problem-formulation-level innovations. For further reading, we recommend the survey on MPC for legged and humanoid robots~\cite{MPC_survey}. Additionally, loco-manipulation tasks present further challenges due to the complexity of dynamic interactions with both the environment and objects, which leaves open questions on how to choose and formulate the interactive dynamics effectively based on the specific task requirements in an MPC setting.}

\section{Whole-body control}
\label{sec:control}


Whole-Body Control (WBC) represents a body of controllers that generate joint torques, constraint forces, and generalized accelerations to achieve a given set of desired dynamic tasks~\cite{humanoid_reference}. 
Three common cases necessitate a computationally efficient whole-body controller that can track desired trajectories and send torque commands to a physical robot. 
(i)~The desired trajectory is computed based on a reduced-order model. Such a trajectory encodes only an important subset of the robot's full-body motion (\textit{e.g.}, desired CoM and end-effector trajectories in operational space~\cite{li2023multi}) and does not contain information for all joints. (ii)~The trajectories are planned with a full-order model but are too computationally heavy~\cite{gait_library_IJRR} to be used in real-time, particularly for humanoids in loco-manipulation scenarios. (iii)~Environmental uncertainties and planning inaccuracies induce disturbances that require robust WBCs to compensate~\cite{CBF}. Therefore, the WBC has been widely used in the humanoid community. 


\revised{A WBC is different from an MPC in two aspects. (i) The MPC often solves a receding horizon problem, whereas the WBC solves an instantaneous control problem (\textit{i.e.}, only for the current timestep). (ii) The model adopted by MPC may vary based on the user choice, as discussed in Sec.~\ref{sec:planning}, but WBC usually employs full-order Euler-Lagrangian dynamics} that express equation~(\ref{eq:nonlinear_dynamics}) as
\begin{equation}
M \boldsymbol{\ddot{q}} - J^T \boldsymbol{\lambda} - S^T \boldsymbol{\tau} = -C\boldsymbol{\dot{q}} - G, 
\label{eq:equation-of-motion}
\end{equation}
%
where the decision variables $\boldsymbol{X} = [\boldsymbol{\ddot{q}}, \boldsymbol{\lambda},\boldsymbol{\tau}]^T$ are generalized accelerations, external forces, and joint torques, respectively. $M, C, G$ are the spatial inertia matrix, bias terms (\textit{i.e.}, centrifugal and Coriolis forces), and the gravity term, respectively. $J$ is the Jacobian, and $S$ is the selection matrix. Given the selection of decision variables $\boldsymbol{X}$ above,~(\ref{eq:equation-of-motion}) becomes linear, which enables the WBC to be computed in real time.

In a humanoid robot with a floating base, the rank of $S$ is smaller than the dimension of the generalized position $\boldsymbol{q}$, meaning the system is underactuated. This requires physical contact with the environment to achieve balance, mobility, and manipulation. The contact constraint is described using contact Jacobian $J_c$:
\begin{equation}
J_c \boldsymbol{\dot{q}} = 0 \Rightarrow J_c \boldsymbol{\ddot{q}} + \dot{J}_c \boldsymbol{\dot{q}} = 0.
\label{eq:contact_constraint}
\end{equation}
These underactuated and contact-constrained dynamics (\ref{eq:equation-of-motion}), (\ref{eq:contact_constraint}) represent the main components for solving the WBC for humanoid robots.

\subsection{WBC Dynamic Tasks} 
A dynamic task vector $e_i$ can be expressed as a linear equation with respect to decision variables:
\begin{equation}\label{eq:task-template}
    e_i = A_i(\boldsymbol{q}, \boldsymbol{\dot{q}}, t) \begin{pmatrix}\boldsymbol{\ddot{q}} \\ \boldsymbol{\lambda} \\ \boldsymbol{\tau} \end{pmatrix} - \boldsymbol{b}_i(\boldsymbol{q}, \boldsymbol{\dot{q}}, t),
\end{equation}
where $t$ is the time. Dynamic tasks can be equality constraints ($e_i=0$), inequality constraints ($e_i\leq 0)$, or cost terms ($|e_i|^2$). The main idea of WBC is that the linear equation (\ref{eq:task-template}) is sufficient to describe a \revised{universal} set of locomotion and manipulation tasks.

Although the appropriate set of WBC tasks depends on factors ranging from robot morphology to available hardware sensing, we highlight some of the common tasks for loco-manipulation. A task for tracking reference joint-space accelerations $\boldsymbol{\ddot{q}}^d$ can be formed by setting $A_i$ to a selection matrix and setting $\boldsymbol{b}_i = \boldsymbol{\ddot{q}}^d$. Similarly, a task for tracking a desired operational-space acceleration is derived through the end-effector's Jacobian~\cite{khatib1987unified}. 
A  task for tracking a desired centroidal momentum rate $\dot{h}^d$ can be formed by differentiating the centroidal momentum $h$~\cite{hopkins2015compliant}. 
Other potential WBC tasks include capture point~\cite{koolen2016design}, reference reaction forces~\cite{chignoli2021humanoid} and collision avoidance~\cite{RMP}.
The source of these dynamic tasks varies and may be predefined, computed online (\textit{e.g.}, from an MPC), or commanded through teleoperation. 

MPC is commonly used to provide WBC with dynamic tasks in operational space. For example, an SRBM-based MPC~\cite{li2023multi, yu2024learning} outputs the centroidal trajectories and end-effector trajectories as dynamic tasks in operational space. These operational-space tasks can also be converted to joint accelerations and thus become joint-space tasks. For instance, whole-body inverse kinematics~\cite{Mistry_humanoidIK} is a common approach for this conversion. Additionally, Riemannian motion policy~\cite{RMP} and kino-dynamics fabric~\cite{Fabrics_WBC} can construct diverse joint acceleration from a hierarchy of primitive motions.

Teleoperation provides an interactive way to generate dynamic tasks such as the robot's posture, walking direction, and grasp targets~\cite{TRO_teleop_survey}. WBC setpoints are often mapped to a visual interface, enabling an operator to modify controller setpoints on the fly. This mapping may be direct~\cite{marion2018director} or retargeted in order to account for the robot's morphology~\cite{penco_2018_retarget} or to ensure the feasibility of the commanded motion~\cite{rouxel2022multicontact, Oh_retarget_TMECH}. Virtual-Reality (VR) interfaces enable spatially mapping handheld controllers to WBC poses. This approach has been deployed in various loco-manipulation scenarios, including doorway traversal, object grasping, and pushing tasks~\cite{jorgensen_2022_cockpit_vr, wallace2020multimodal}. Haptic feedback can inform an operator of the WBC state through various modalities, such as force feedback indicating CoM stability margin~\cite{abi2018humanoid} and vibrating gloves indicating contact during manipulation~\cite{Dafarra_2024}. Mapping to dynamic WBC setpoints, such as the capture point, has also been demonstrated and can account for variation in the natural walking frequency of the operator and robot~\cite{ramos2018humanoid, colin2023bipedal}.
\begin{figure}[!t]
    \centering
    \includegraphics[width=\columnwidth]{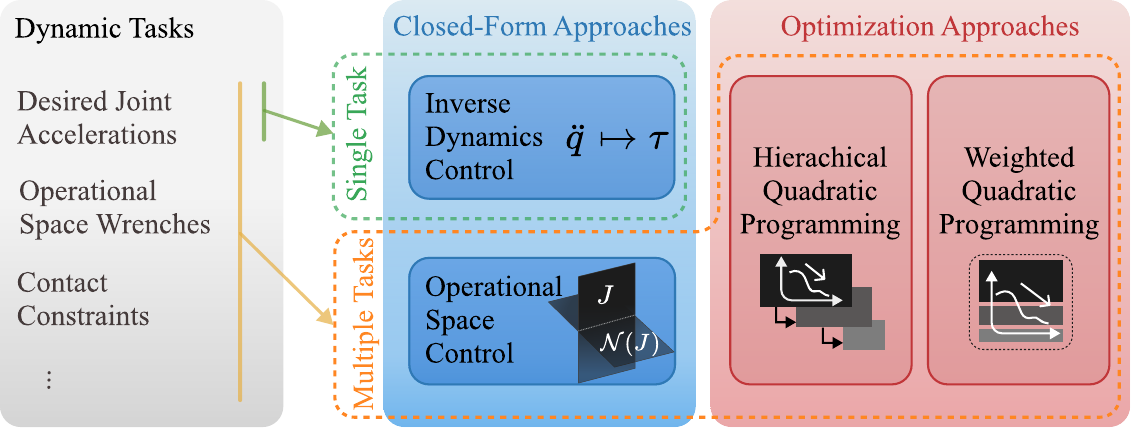}
    \caption{Whole-body control is categorized into closed-form approaches and optimization-based approaches. Both approaches can incorporate multiple dynamic tasks and resolve task conflicts. }
    \label{fig:WBC_structure}
    \vspace{-0.1in}
\end{figure}
As shown in Fig.~\ref{fig:WBC_structure}, WBC approaches can be categorized based on closed-form or optimization-based approaches to achieve a desired list of dynamic tasks. 

\subsection{WBC in Closed Form} 
An inverse dynamics controller is among the early works that address the WBC problem in closed form. In particular, it solves a single dynamic task: achieving the desired generalized acceleration $\boldsymbol{\ddot{q}} = \boldsymbol{\ddot{q}}^d$. The closed-form torque $\boldsymbol{\tau}$ can be solved from~(\ref{eq:equation-of-motion}) if we can measure all constraint forces $\boldsymbol{\lambda}$. However, $\boldsymbol{\lambda}$ \revised{is} usually unattainable due to the lack of sensing and estimation capability. To derive torque analytically, several methods~\cite{Sentis2007, Mistry2010, IDC_unify} project the system dynamics~(\ref{eq:equation-of-motion}) into a constraint-free manifold, establishing an equation between only $\boldsymbol{\ddot{q}}$ and $\boldsymbol{\tau}$. \cite{Sentis2007} projects~(\ref{eq:equation-of-motion}) into the actuated joint space, resulting in an equation of constrained dynamics independent of the constraint forces $\boldsymbol{\lambda}$.~\cite{Mistry2010} uses orthogonal decomposition of the constraint Jacobian $J$ to project~(\ref{eq:equation-of-motion}) into the nullspace of the constraints. Although different projection techniques are used,~\cite{IDC_unify} shows that these projections produce equivalent torque results. However, from the perspective of solving speed and numerical stability, orthogonal decomposition~\cite{Mistry2010} is favored as it is faster and does not require the inversion of the inertia matrix. 


Other than the task of tracking generalized acceleration $\boldsymbol{\ddot{q}} = \boldsymbol{\ddot{q}}^d$, a set of operational-space tasks and constraints can be achieved given the redundancy in the degrees of freedom of humanoid robots. As a multi-task example, a humanoid robot is often tasked to generate interaction forces with a low priority while maintaining the whole-body balance with a high priority. Operational-Space Control (OSC), \textit{a.k.a} task-space control, achieves multiple dynamic tasks by prioritizing tasks in a hierarchy~\cite{Sentis2007, mistry_OSC_2008}. Tasks with low priorities are solved within the null space of the high-priority tasks, enforcing that hierarchies are strictly maintained among tasks. Such a task hierarchy is also named the stack of tasks~\cite{SoT_unilateral}. 

Overall, closed-form approaches are computationally efficient and are straightforward to implement. However, they have difficulty in incorporating inequality tasks, such as joint limit and obstacle avoidance. Although this issue can be addressed within the closed-form approach, such as using a smooth operator~\cite{SoT_unilateral}, much of the community uses optimization-based methods that address this issue efficiently. 

\subsection{WBC through Optimization}
\label{sec:WBC_optimization}
In contrast with closed-form approaches, there have been a variety of studies formulating the WBC as an optimization problem. 
These optimization-based methods enhance the flexibility of WBC, enabling the modular addition and removal of dynamic tasks~\cite{keith2011iros,rocchi_opensot_2015}, including inequality tasks. 

A salient feature of optimization-based WBC is the resolution of conflicting dynamic tasks through two prioritization schemes: strict task hierarchy~\cite{ Escande_HQP_humanoid} or soft task weighting~\cite{bouyarmane_using_2011, koolen2016design, Kuindersma2016, hopkins2015compliant}. 
Due to the linear property of both the dynamics equation~(\ref{eq:equation-of-motion}) and dynamic tasks~(\ref{eq:task-template}), optimization-based WBC is often formulated as a Quadratic Program (QP), which can be solved efficiently to a global optimum and enjoys a wide range of solver selections. 

A strict task hierarchy can be ensured through a cascaded hierarchical QP. This method sequentially solves a series of QP subproblems with tasks from high to low priorities; lower-priority QPs produce solutions within the combined null space of all preceding QPs~\cite{Escande_HQP_humanoid}. The sequential solve of QP terminates either when it successfully solves all subproblems or when it encounters an infeasible subproblem and thus skips all the remaining low-priority tasks~\cite{de_lasa_feature-based_2010}.
Hierarchical QP is essentially equivalent to the closed-form stack-of-tasks approach, with the benefit of incorporating inequality constraints more naturally. 
However, solving multiple QP subproblems imposes significant computation and memory burdens. Additionally, the hierarchical QP inherits a common issue of OSC: the task Jacobians becomes rank-deficient when approaching singularities, which induce large unstable movements~\cite{pfeiffer_singularity_2018}.

In contrast to hierarchical QP, which solves tasks in a strict order, weighted QP resolves task conflicts by treating dynamic tasks as \textit{soft constraints} within the cost functions, with weights indicating their relative priority. Therefore, weighted QP can be regarded as a special case of hierarchical QP with only one hierarchy or vice-versa as documented in~\cite{bouyarmane2018tac}. Such a setup enjoys the benefit of solving only a single optimization, which is faster than a hierarchical QP and can be further accelerated by exploiting sparsity and warm-start. 
However, tuning weight parameters can be burdensome for a large number of tasks, and can lead to instability~\cite{djeha2023tro}. Even with well-tuned parameters, the loss of strict task priorities means low-priority tasks can interfere with high-priority ones. Nevertheless, weighted QP is still widely applied in many robotics studies due to its easy setup and computation efficiency compared with hierarchical QP. For example, many of the weighted QP methods were designed during the DARPA Robotics Challenge~\cite{koolen2016design, siyuan_JFR, Kuindersma2016, hopkins2015compliant}.

\subsection{WBC for Loco-manipulation}

WBC for loco-manipulation aims to achieve the desired motion while maintaining instantaneous \textit{balance} and \textit{contact stability}. 
Given the desired motion and contact sequence, loco-manipulation control can be categorized into two folds. 
(i)~When all interactions with the environment and objects are static or quasi-static, they can be modeled as external wrenches. In this case, the WBC solves a robot balancing problem with the external wrenches as dynamic tasks. 
(ii)~When the manipulated object has a substantial mass or is dynamically moving, such as carrying a heavy box, the object becomes an integral part of the robot-object system. Therefore, the WBC must account for the balance of both the robot and the dynamic object.

\subsubsection{Interaction as an External Wrench}
In this first category, a subset of contacts is responsible for interacting with the environment or objects to apply a desired wrench. This desired wrench can be specified by a user or derived from the estimated object weight. Considering the desired wrench from the interaction, the remaining contacts maintain the system balance using three distinct strategies. 

The first strategy involves simultaneously optimizing contact wrench, joint acceleration, and joint torque using the robot's full-body dynamics, as detailed in Sec.~\ref{sec:WBC_optimization}. In this approach, the desired wrench for interaction is a dynamic task within the WBC. The WBC must also satisfy the dynamics constraint, contact stability constraint, and balance stability constraint. The contact stability constraint enforces that the resultant contact wrench lies inside the contact wrench cone (CWC)~\cite{adios_ZMP}, maintaining firm and stable contact. The balance stability constraint designs a desired rate of centroidal momentum, often based on feedback in the CoM position and body orientation~\cite{orin2013centroidal}. 
In the presence of state deviations, the balance stability results in a redistribution of the contact wrenches or a movement of the centroidal state to counteract and restore stability~\cite{Murooka_Pattern}.

The second strategy, known as pre-optimization~\cite{henze2016passivity, force_dist_IDC_Ludovic}, involves two stages in sequence. First, it determines the optimal distribution of contact wrenches based on the desired rate of centroidal momentum derived from the balance stability of CD. The second stage computes the joint torques needed to realize the contact wrenches using inverse dynamics of full-body dynamics. Note that, deriving the desired rate of centroidal momentum in the first stage is particularly challenging due to the non-holonomy~\cite{Nonholonomy_Wieber} of angular momentum, \textit{i.e.}, the kinetic momentum of rotation is not directly related to the orientation of body links. As a result, the body orientation requires additional regulation (\textit{e.g.}, joint-level postural feedback ~\cite{henze2016passivity}) beyond simple feedback on angular momentum.

To address the non-holonomy issue, the third strategy uses post-optimization~\cite{Post_Optimization_balance}. The main idea is to treat the floating-base robot as a fixed-based system when calculating joint torques. The underactuated portion of the obtained torque is then mapped to contact wrenches through an optimal distribution problem. This method avoids the challenge of specifying the momentum of rotation in the pre-optimization strategy.

\subsubsection{Interaction as a Unified Robot-Object Model}
A unified robot-object system can leverage the additional object to regulate the robot's dynamics. This yields more dynamically feasible behavior when carrying heavy or dynamically-moving objects. The unified model incorporates each manipulated object as an additional ``robot" -- either a passive object or a real robot --  and connects the robot and object via action-reaction force pairs~\cite{QP_Multirobot_Kheddar}. 
The balance stability must consider the combined CoM and inertia of the robot-object system~\cite{Otani2017Retarget}. Additionally, the contact stability between the robot and the object is maintained to ensure that the object remains securely attached. 
While direct control of the interaction forces is feasible, adaptive force control that regulates the relative position between the object and the robot offers greater robustness. This approach mitigates the impact of inevitable inaccuracies in modeling inertia parameters and stiffness properties~\cite{audren2014iros}.

\revised{\textit{Key Takeaways for Whole-Body Control:} The core of whole-body control lies in addressing an inverse dynamics problem to produce joint-level torque control. However, this problem is challenging due to the underactuation and contact-constrained nature of humanoid robots. Closed-form approaches such as inverse dynamics control and operational-space control are computationally efficient. Therefore, they have been traditionally prevalent. On the other hand, optimization-based strategies, particularly quadratic programs, are increasingly favored as they adapt more effectively to a wide range of task specifications and offer more reliable solutions. Undoubtedly, both lines of WBC research have significantly advanced the progress of humanoid robot control over the past two decades. In the near term, optimization-based WBC will continue to be a popular choice for low-level control to achieve high-level loco-manipulation tasks. We also see neural WBC~\cite{ji2024exbody2, he2024omnih2o, humanplus} gaining popularity, as we will discuss in the following section. For further reading, we recommend the survey on optimization-based WBC for legged robots~\cite{Wensing_survey_TRO} and the chapter on closed-form WBC techniques for humanoid robots~\cite{humanoid_reference}.}

\subsection{Challenges in Numerical Optimization}

Robotic planning and control techniques that are formalized as numerical optimization problems heavily rely on advances in applied discrete mathematics and optimization theory. \solution{These advancements address challenges such as nonconvexity, numerical robustness, and real-time resolution performance.} However, a plateau may have been reached in the transfer of these techniques to the field of robotics at large -- humanoids specifically -- despite exploiting the unique properties of robotic models. These properties can enhance the efficiency of optimization problem formulation and its resolution by tailoring them to specific applications. However, as evidenced by the formulation OCP, including MPC (Sec.~\ref{sec:planning}) and WBC (Sec.~\ref{sec:control}), inherent physical uncertainties can disrupt closed-loop performance. Extending these formulations to loco-manipulation primarily involves (i)~augmenting models to incorporate loco-manipulated counterparts and (ii)~refining contact models formulations to account for various interactions (\textit{e.g.,} impact~\cite{wang2023ijrr}, rolling, deforming). However, these extensions risk overcomplicating the problem, potentially hindering effective formalization, even if the resulting formulations are sparse.


Contact-explicit optimization formulations~\cite{khazoom2024tailoring} are generally preferred due to their faster convergence and simplified formulation. However, they still suffer from the \textit{curse of dimensionality}\revised{, which limits the complexity of the considered problem, the length of the preview window, and the discretization granularity of the decision variables. While increasing the resolution of the discretization and the length of the preview horizon increases the quality and stability of the results, the corresponding decrease in solution speed can, in some cases, cause an overall reduction in system stability. \solution{However, as computational power increases, this lowers the significance of these challenges.}}

\revised{Importantly, however,} the contact-explicit formulations have the significant limitation of requiring the user to determine the contact mode sequence for the problem, which generally limits the ability to generate complex motions. Alternative\revised{ly,} \solution{contact-implicit formulations introduce complementarity conditions to eliminate the strict dependence on the contact mode sequence~\cite{manchester2020variational, patel2019contact, carius2018trajectory}}. However, contact complementarity conditions are nonsmooth, introducing severe computational challenges. Generally, this is tackled via regularizing the complementarity problem, \emph{e.g.},~\cite{aydinoglu2023consensus}, which approximates the constraint with a continuous affine function. Even with this linearized approximation, contact-implicit approaches struggle to scale to the high dimensionality of humanoid robots due to excessive computation and numerical difficulty. \solution{However, improved solver formulations, either as first-order solvers such as Alternating Direction Method of Multipliers (ADMM) \cite{aydinoglu2022real} or specialized methods \cite{le2024fast} for handling the complementarity conditions, have the potential to increase the tractability of real-time contact-implicit optimization.}

\revised{Possibly more importantly}, all these approaches, however, still have only \textit{local} optimality guarantees. \revised{The level of nonconvexity or nonlinearity of the problem influences the solution quality, which often results in poor local minima or sub-optimal solutions. When the structure of the problem requires deviations from local candidate solutions, a feasible solution, even one that is only locally optimal, may never be found. This feasibility issue remains an open challenge for nonconvex nonlinear optimization.}

Additionally, optimization approaches are almost always deterministic in nature, failing to capture the stochasticity in the state estimates and future contact events. \revised{This lack of consideration of uncertainty along the optimization horizon is critical, as it leads to limited robustness in translation to the real world}. \solution{Addressing this uncertainty and lack of global optimality has led to the combination of search techniques with traditional trajectory optimization. For example, Model Predictive Path Integral (MPPI)~\cite{williams2016aggressive} samples a variety of random control signals and their resulting state trajectories to determine the best action to take.} Alternative contacts and objectives can also be sampled with contact-implicit approaches to help avoid local minima and find the globally optimal solution to accomplish a task~\cite{Venkatesh_Penn_thesis}. Although both of these approaches heavily leverage computational parallelization for expediency, parallelizing the underlying optimization algorithms explicitly designed for trajectory optimization is gaining increasing prevalence, both on the CPU~\cite{plancher2020performance} and GPU~\cite{adabag2024mpcgpu, bishop2023relu}, as discussed in Sec.~\ref{sec:planning}. \solution{Despite this, the inclusion of the uncertainty into the optimization problem, whether through sampling \cite{kong2024saltation}, smoothing \cite{suh2022bundled}, or the introduction of stochasticity \cite{hammoud2022irisc}, is a promising approach for mitigating some of these challenges.}

Despite the gains in these algorithms for considering the full system dynamics, the robustness of the mathematical solution when performing numerical optimization has been a concern due to the infeasibility of complex optimization problems, regardless of the computation speed improvement. As discussed in Sec.~\ref{sec:control} WBC, arbitrating techniques address infeasibility by (i)~relaxing the hard constraint to the soft constraint by combining them in the cost with a weighted sum and (ii)~prioritizing the constraints by achieving the important ones first. However, how to design a smart solver to automatically resolve this issue and provide numerical robustness \solution{is still an open question}. In addition, weight-tuning in high-dimensional problems with complex objectives is nontrivial, highly task-dependent, and can lead to instability~\cite{djeha2023tro}. \solution{Researchers have made initial steps to apply auto-tuning techniques to streamline the tuning process in Optimal Control Problem (OCP) designs for humanoid robots~\cite{chen2024autotuning,sartore2024automatic}.} Until these are solved, assigning definitive but non-violating constraints and designing objective functions while maintaining global versatility still depend on expert knowledge.

\section{Learning Loco-Manipulation Skills}
\label{sec:learning}

Robot skill refers to the ability to use its own perception, planning, and control capabilities to complete specified tasks autonomously~\cite{Jiang2024skill_learning}. Among a variety of robot skills, loco-manipulation is highly valuable for augmenting and complementing human capabilities.
Traditionally, loco-manipulation skills are developed from human designer knowledge, distilled into pre-programmed planners or controllers. 
In contrast, learning-based methods leverage computation and data. Although learning skills require collecting extensive data from either autonomous exploration or expert guidance, this approach is powerful as it tends to yield novel behaviors that are difficult to encode from human knowledge.

This section reviews learning-based approaches that explore two main directions: 
(i) enhancing a specific skill in terms of agility, robustness, and safety, and (ii) broadening the overall skill set of robots, revolving around two pivotal goals: versatility and generalization. Versatility refers to the capability of a single framework or policy to master \textit{multiple} skills, whereas generalization involves adapting existing skills to new, out-of-distribution tasks and environments. \revised{Versatility can be viewed as a prerequisite for generalization. For a model to generalize across a wide range of tasks, it must possess a broad and versatile set of skills that can be applied in novel contexts. However, possessing versatile skills alone does not guarantee generalization—the model must also be capable of correctly prompting and adapting those skills to meet novel objectives. For instance, in the context of skill learning, a versatile policy can track multiple trajectories within the training dataset, while a generalizable policy should be able to track motions beyond those seen in the dataset, thereby demonstrating out-of-distribution capabilities.}

Among learning-based methods, Reinforcement Learning (RL) without demonstration and learning from demonstration, also known as Imitation Learning (IL), have shown remarkable proficiency for robotic skill learning. RL has been successful in coordinating complex full-body motions for humanoid robots, including dancing~\cite{zhang2024wococo,ji2024exbody2}, agile soccer maneuvers~\cite{SR_deepmind_soccer}, and robust locomotion~\cite{transformer_digit_RL, li2023robust}. However, RL policies are often fine-tuned for specific tasks within specific environments. 
This limitation largely stems from the reward function being narrowly tailored to a specific task, and the policy only capable of learning from the same or similar environments. 
In contrast, IL addresses this problem by leveraging large datasets of demonstrations~\cite{huang2024diffuseloco, humanoid_parkour_learning}. Recent advancements in IL have demonstrated promising results for scaling to a large number of skills~\cite{diffusion_policy}, showing potential for solving complex multi-skill tasks.

\begin{figure}[!t]
    \centering
    \includegraphics[width=\columnwidth]{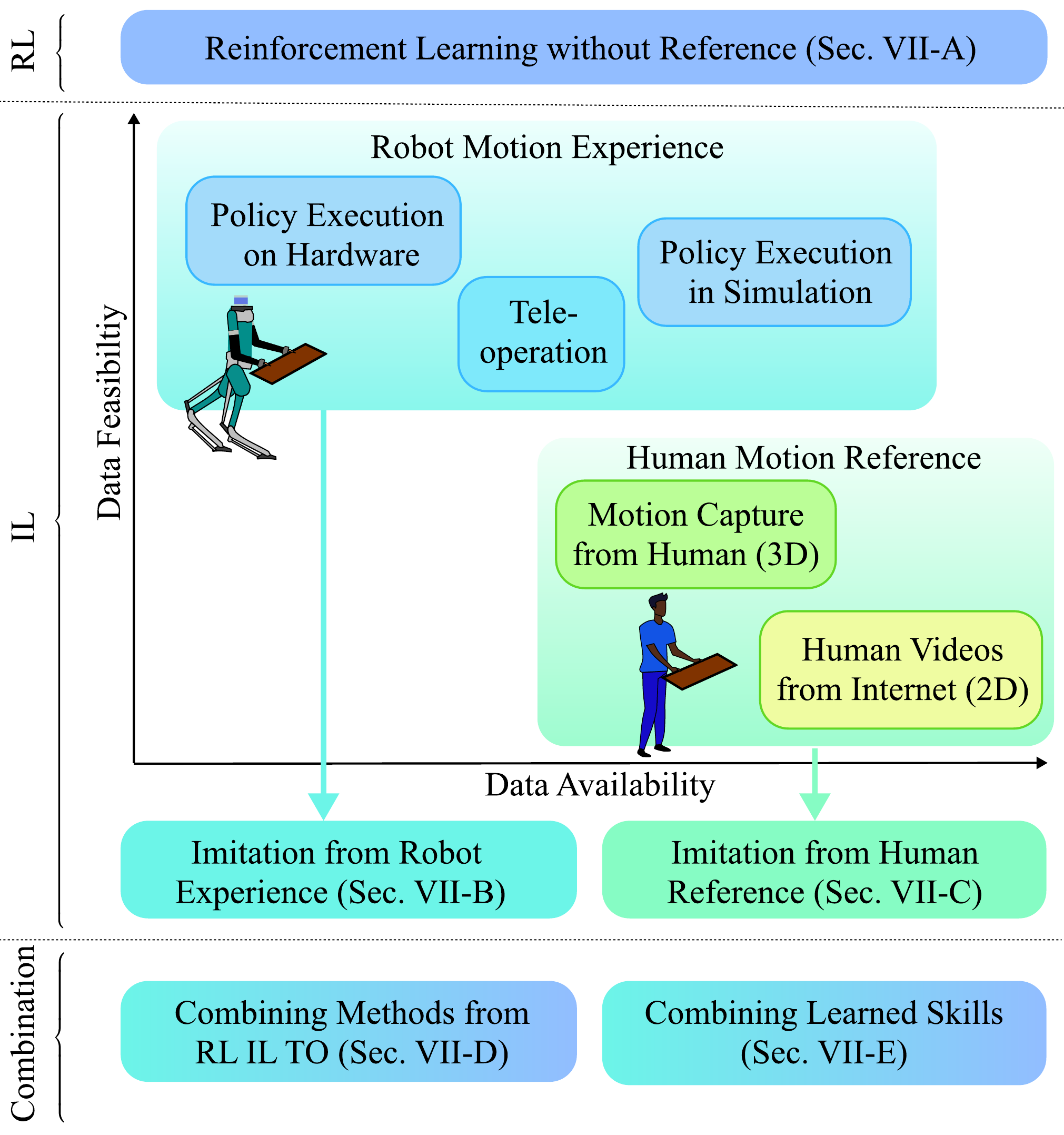}
    \caption{The organization of approaches for skill learning. RL does not require reference data in a standard setup. IL makes use of four different data sources for efficient learning. The larger the morphology gap or sim-to-real gap is, the more challenging it becomes to learn effectively from these data.}
    \label{fig:data_source}
    \vspace{-0.15in}
\end{figure}

For the basics of RL and IL, we refer readers to the survey paper~\cite{ha2024learningbased}. In this section, we discuss these methods for learning humanoid loco-manipulation skills. As shown in Fig.~\ref{fig:data_source}, we introduce RL in Sec.~\ref{sec:RL} and IL in Sec.~\ref{sec:expert} and Sec.~\ref{sec:human}. Thereafter, we discuss the benefits of combining model-based and learning-based methods in Sec.~\ref{sec:combine_method}. Finally, we discuss methods for learning versatile skills with a single policy in Sec.~\ref{sec:skill_composition}.

\subsection{Skill Learning: Reinforcement Learning from Scratch}
\label{sec:RL}
RL, enhanced by modern deep learning toolchains and algorithms, has garnered significant attention in the field of robotics over the past decade. RL promises an effective way to learn motor skills by rewarding desirable behaviors and penalizing undesired behaviors, with minimal or no supervision during training. The end-to-end RL policies translate raw sensory input to actuation and are executable in real time.

RL provides distinct benefits but also comes with its own challenges. Compared with model-based methods, numerous RL methods are model-free (see Sec.~\ref{sec:skill_composition} for model-based RL), eliminating the need for accurate dynamics. Furthermore, RL does not require demonstration data, making its training setup straightforward. 
However, it often requires meticulous design of reward functions to shape the policy's behavior. In addition, deploying a learned policy on robot hardware often encounters a sim-to-real gap, a well-known issue induced by the inaccurate physical model used by the simulator. Policy learning from scratch often requires extensive and time-consuming interactions with the environment without a guarantee for task completion. 
For example, popular RL algorithms such as Proximal Policy Optimization (PPO)~\cite{PPO} and Soft Actor-Critic (SAC)~\cite{SAC} fail in most humanoid loco-manipulation tasks~\cite{sferrazza2024humanoidbench}, partly due to the complexity of these tasks, the sample inefficiency, and the sparse reward design. 

\subsubsection{\revised{Comparison Between RL Algorithms}}
\revised{PPO \cite{PPO}, and similarly other on-policy reinforcement learning algorithms such as A3C \cite{A3C}, is sample inefficient, but the policy update is fast and scalable. PPO is widely used in sim-to-real settings where simulation data is relatively cheap to obtain, \textit{e.g.}, in a parallelized simulation environment such as Isaac Gym \cite{Orbit}. SAC \cite{SAC}, and similarly, Twin Delayed Deep Deterministic (TD3)~\cite{TD3} and Deep Deterministic Policy Gradient (DDPG)~\cite{DDPG}, are sample-efficient off-policy algorithms, but slower to train due to higher computational costs for the policy updates. In addition, Q-learning-based methods such as SAC can introduce model bias during the training process since they estimate gradients using a learned Q-function, whereas policy gradient methods such as PPO have the advantage of directly estimating the gradients of the learning objective. Off-policy algorithms are widely used when data collection is expensive, \textit{e.g.}, when the data can only be obtained from a real robot. The recent literature on RL for humanoid robots relies mainly on simulation data, and hence PPO is widely employed. For other RL settings, such as real-world RL, we envision that more off-policy RL algorithms and novel RL algorithms will play an important role in learning humanoid skills.
}

\subsubsection{Improving Learning Efficiency}
Several approaches have been proposed to improve learning efficiency. Curriculum learning expedites training by allowing the policy to achieve simple tasks in the early stage of training and then progressively increasing the task difficulty and complexity~\cite{pmlr-v100-xie20a}. 
Another approach is to promote exploration. Researchers use curiosity mechanisms, which encourage visiting unexplored states, to intrinsically motivate learning without an explicit reward design~\cite{Curiosity-schwarke23a}. This has been shown to overcome the sparse reward setting and achieve complex loco-manipulation behaviors such as door opening.~\cite{zhang2024wococo} also incorporates curiosity-based rewards to learn versatile loco-manipulation skills without any motion priors. 
Lastly, substituting reward terms with constraints in a constrained RL framework can significantly simplify reward tuning while achieving state-of-the-art locomotion performance~\cite{reward_constraint_RL}.

\subsubsection{Addressing the Sim-to-real Gap}
Sim-to-real is another formidable challenge in RL. Nevertheless, RL has been successfully applied to various areas of robotics, notably in quadrupeds~\cite{Hutter_SR2020}, where the sim-to-real gap has been consistently overcome. 
The success of quadrupeds hinges on new investments in infrastructure for affordable hardware and highly parallel physics engines, spearheaded by key players in the robotics field. It is also important to note that quadrupeds benefit from an inherently stable dynamic system similar to manipulators while operating in less complex environments compared to typical loco-manipulation tasks. 

In contrast, humanoid loco-manipulation faces steeper sim-to-real challenges. Humanoid robots possess higher DoFs and unstable dynamics, where the center of mass constantly moves out of the support polygon. Therefore, learning whole-body balance control is sensitive to parameters in a physics simulation, underlining the sim-to-real gap due to differences in dynamics between the virtual and real worlds. Additionally, humanoids are expected to perform human-level manipulation tasks where the complex environment and the differences in observation space, such as visual appearances, aggravate the sim-to-real-gap. 

To address the sim-to-real challenge, a diverse set of mainstream approaches have been explored for humanoid robots. 
Domain Randomization (DR) is among the most widely adopted approaches, which varies the properties of a robot model, such as mass, friction, and actuator dynamics, to train a generalized policy robust in the real world. Many humanoid works~\cite{SR_deepmind_soccer, transformer_digit_RL} achieve sim-to-real transfer through DR. While DR is straightforward to set up, policy training is sensitive to the parameter randomization range, inducing laborious tuning: a larger range is challenging for the policy to fit all physical parameters (\textit{i.e.}, fail to learn), and a smaller range does not cover the full spectrum of parameters that can occur in the physical world (\textit{i.e.}, fail to transfer). 

System Identification (SI) is another popular approach to enhancing model fidelity by approximating the system's input-output behavior from real-world data. 
Real-to-sim techniques use optimization~\cite{Tan_systemID} or search~\cite{li2024learningagilebipedalmotions} to obtain simulation parameters
that can best explain the real robot trajectories collected from policy execution. \revised{Learning to align simulation behavior with policy rollouts in the real world has also shown success in sim-to-real transfer~\cite{asap}.} However, it is challenging to collect real-world data covering the full space of states and actions, particularly for versatile, safety-critical tasks. 

While SI uses real-world data to obtain an accurate model, Domain Adaptation (DA) uses real-world data directly to fine-tune a simulator-trained policy. In DA, the parameter distribution in simulation is defined as the source domain, and parameter distribution in the real world is defined as the target domain. 
The fine-tuning effectively transfers the policy from the source domain to the target domain. For example, Sim-to-Lab-to-Real~\cite{hsuren2022slr} develops a two-stage transfer: pre-training in simulation and fine-tuning in the real world. Although only limited hardware data is needed for fine-tuning, safety is still a major concern. Safety filters are often deployed to prevent dangerous movements when collecting real-world data~\cite{SafeRLRoboticsSurvey}.

Despite these efforts to address the sim-to-real gap, a systematic solution remains elusive, as the aforementioned approaches are often case-specific.
Against this backdrop, developing physics engines that facilitate real-to-sim construction and sim-to-real transfer is crucial. 
\revised{However, physics engines are constrained by real-time requirements and continue to face challenges in accurately solving contact dynamics. For instance, MuJoCo~\cite{todorov2012mujoco} employs a soft contact model that can lead to unphysical penetration artifacts, while PhysX~\cite{Orbit} approximates friction cones using low-fidelity pyramids that widen the sim-to-real gap. Although existing solutions inevitably involve trade-offs between accuracy and efficiency, advancements in contact modeling will significantly narrow the sim-to-real gap.}  

\revised{\textit{Key Takeaways}: 
RL provides an effective way to learn novel behaviors for humanoid loco-manipulation. 
However, in practice, the success of RL often relies on informative representations for both observation and action, extensive reward engineering, curriculum learning design, and a vast amount of trial-and-error experiences to estimate gradients. Consequently, using RL to train robots is almost never practical in the real world, at least for the current stage of development. Therefore, RL policies are predominantly trained in simulations. This makes the sim-to-real gap the Achilles' heel of RL, significantly dampening its initial promise. Compared with quadruped robots, the sim-to-real gap is particularly challenging for humanoid robots with high DoFs executing complex loco-manipulation tasks. This is why IL, leveraging limited but in-domain real-world data, has gained popularity over pure RL without demonstration data. For further reading in RL, we recommend the survey on learning-based legged locomotion~\cite{ha2024learningbased} and the practical lessons for training robotic RL agent~\cite{RL_Robot_Levine}.}

\begin{table}[!t]
\centering
\setlength{\tabcolsep}{4pt}
\caption{Skill Learning Methods Based On Data Source}
\begin{tabular}{lll}
\hline
Methods and Data & Pros and Cons & Algorithms \\ \hline\hline
\begin{tabular}[c]{@{}l@{}}RL Without \\Reference\end{tabular} & \begin{tabular}[c]{@{}l@{}}\cmark novel behavior \\ \xmark \,reward tuning\end{tabular} & \begin{tabular}[c]{@{}l@{}}PPO~\cite{PPO}, SAC~\cite{SAC}\end{tabular} \\ \hline
IL Robot Execution   &  \begin{tabular}[c]{@{}l@{}}\cmark annotated data \\ \cmark dynamically feasible \\ \xmark \,scarce \\ \xmark \,limited diversity \end{tabular}   &  \begin{tabular}[c]{@{}l@{}} Diffusion~\cite{huang2024diffuseloco}, IRL~\cite{wu2023infer}\end{tabular}  \\ \hline
IL Teleoperation     &  \begin{tabular}[c]{@{}l@{}}\cmark multimodal behavior \\ \xmark \,rare full-body motion\end{tabular}   &   \begin{tabular}[c]{@{}l@{}}BC-RNN~\cite{seo2023deep},\\ ACT \cite{humanplus, cheng2024tv}\end{tabular} \\ \hline
IL Motion Capture    &  \begin{tabular}[c]{@{}l@{}}\cmark accurate kinematics \\ \xmark \,small dataset \\ \xmark \,limited outdoor data \\ \xmark \,proprioception-only \end{tabular}  &  \begin{tabular}[c]{@{}l@{}}RL motion imitation \\ \cite{physhoi}, GAIL~\cite{gail_stand}, \\AMP~\cite{tang2024humanmimic, zhang2024wholebody}  \end{tabular} \\ \hline
IL Human Video       & \begin{tabular}[c]{@{}l@{}}\cmark diverse abundant data \\ \xmark \,non-physical motion \\ \xmark \,proprioception-only\end{tabular}  & \begin{tabular}[c]{@{}l@{}}RL motion imitation~\cite{peng2018sfv}, \\GAIL \cite{gail_locomotion}, OKAMI~\cite{okami2024}\end{tabular} \\ \hline
\end{tabular}
\label{tab:skill}
\vspace{-0.4cm}
\end{table}

\subsection{Skill Learning: Imitation from Robot Experience} 
\label{sec:expert}
Imitation Learning (IL) is an umbrella term that represents a class of algorithms, including supervised, unsupervised, and reinforcement learning, that train policies from expert demonstrations. IL is particularly effective for complex tasks that are difficult to specify explicitly. Three essential steps exist in IL~\cite{ParraMoreno2023ImitationMF}. The first step is to capture the expert demonstration. The next step involves retargeting, where these demonstrations are mapped to the robot motions. If the captured motion comes from the same robot, such as from teleoperation, the retargeting step is unnecessary. The final step is policy training using the retargeted data. 

We discuss four possible sources of demonstrations for humanoid robots: (i)~policy execution, (ii)~teleoperation, (iii)~motion capture, and (iv)~human videos, as illustrated in Fig.~\ref{fig:data_source} and Table~\ref{tab:skill}. We group these data sources into two categories: the robot experience data, which represents those directly obtained from the robots through policy execution or teleoperation, and the human data, which includes human motion captures and videos of human activities obtained from the Internet. Robot experiences exhibit smaller morphological discrepancies and are directly applicable to policy learning but are typically scarce. Conversely, human data are more abundant but present significant morphological differences to humanoid robots. 

\subsubsection{Obtaining Robot Experience Data}
A reliable way to obtain robot experience data is to execute existing policies, either model-based or learning-based. However, collecting data on a physical robot requires a laborious setup of the environment and raises significant safety concerns. Therefore, conducting these executions in simulation is a more efficient approach, although the fidelity of the simulator will inevitably cause a sim-to-real gap. 

\begin{figure}[!t]
    \centering
    \includegraphics[width=\columnwidth, trim=0.0cm 0cm 0cm 0cm,clip]{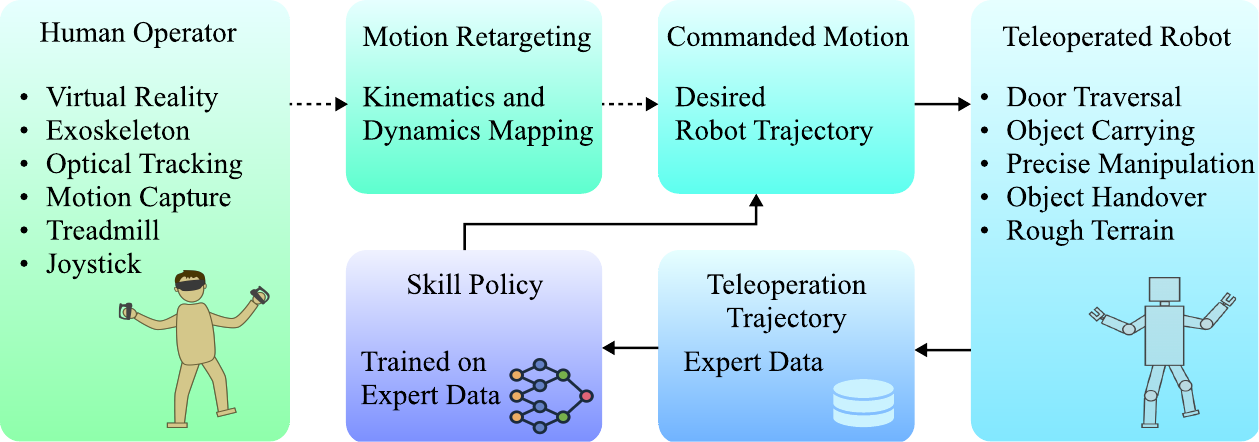}
    \caption{Control flow for learning from teleoperated demonstrations. Expert data is created from teleoperated trajectories (dashed lines), which in turn is used to train and deploy an autonomous skill policy (solid lines).}
    \label{fig:Teleoperation_control_flow}
    \vspace{-0.1in}
\end{figure}

Teleoperation is one of the most common ways to directly capture robot data commanded by human experts. A main advantage of teleoperation is its ability to provide smooth, natural, and precise trajectories for a wide range of tasks. Fig.~\ref{fig:Teleoperation_control_flow} outlines the control flow of teleoperation data used as a source of policy training. The first step of this process is generating the demonstration through teleoperation, represented by the top path of the control flow (dashed lines). Motion retargeting maps human measurements from the teleoperation device to the desired trajectories in the robot domain. Robot execution data collected from teleoperation can be used to train autonomous policies \moved{(solid lines)} that directly command the robot's motion without human intervention.


However, teleoperation for the \textit{full-body} of a humanoid has a number of limitations. Firstly, a majority of teleoperation systems capture only manipulation skills~\cite{diffusion_policy}. Generally, Virtual Reality (VR)-based teleoperation schemes cannot sense the operator's gait and are restricted to simply commanding walking speeds and directions via joysticks~\cite{jorgensen_2022_cockpit_vr, cisnero2024soro, lu2024mobiletelevision}. Full-body sensing, including human gaits, often requires additional equipment such as IMU suits~\cite{penco_2018_retarget, dallard2023ral} or exoskeletons~\cite{ramos2018humanoid}. These devices may be prohibitively expensive, bulky, complicated to maintain, and lack transparency and user-friendliness. Secondly, while teleoperation can generate versatile training data for a wide range of tasks, the utility of this data may be limited if the robot's kinematics fail to seamlessly retarget the reference human motion.  Additionally, retargeting dynamic tasks such as walking or pushing an object relies heavily on the dynamic model of the human demonstrator \cite{penco_2018_retarget} and requires meeting high synchronicity~\cite{dallard2023ral} and rich sensory feedback. 



Although both teleoperation data and policy execution data are recorded from robots, they exhibit distinct characteristics. During teleoperation, human instructors tend to provide diverse demonstrations even for the same task. Therefore, teleoperation data are often multimodal; that is, given a specific task, there exists a distribution of plausible ways to accomplish the task. In contrast, data from executing a single policy are often unimodal; that is, given an input, the output is often fixed. Different policy learning approaches have been proposed to address these multimodal and unimodal data features.

\subsubsection{Approaches to Learning from Robot Experience Data}
From unimodal policy execution data, which contains paired observations and actions, IL approaches are often used for policy distillation. Behavior Cloning (BC) casts IL as a supervised learning problem, which remains one of the most straightforward approaches for robot skill learning~\cite{xu2024humanvla,SuperPADL}. 
Another IL technique is Inverse Reinforcement Learning (IRL), which reconstructs rewards from the data in addition to training an RL policy. The IRL study in~\cite{wu2023infer} infers a generalizable reward of the expert demonstration for bipedal locomotion and then uses it to train an RL policy in unseen terrains. 

To capture the data multimodality and produce diverse future actions, the Action Chunking Transformer (ACT)~\cite{humanplus, cheng2024tv} is adopted to handle distribution shifts due to the compounding error inherent in naive BC. Diffusion policy~\cite{huang2024diffuseloco}, a BC method, shows the ability to acquire multimodal locomotion skills by learning from a large dataset collected from multiple expert policies. However, obtaining these skills requires large-scale versatile data, which motivates the scaling of data collection via teleoperation in industrial companies, \textit{e.g.}, Tesla and Toyota Research~\cite{diffusion_policy}. 

\revised{\textit{Key Takeaways:} Although collecting high-quality data demands considerable effort and resources, IL from robot experience remains a reliable method for attaining skills with expert-level performance. Industrial companies and research labs are increasingly focusing on scaling data collection to develop a broader range of diverse policies through IL. Especially, teleoperation is one of the most popular ways to collect humanoid robot experiences nowadays. For further reading on collecting robot experience data, we recommend the survey on humanoid robot teleoperation~\cite{teleop_humanoid_review}. We also find the survey on imitation learning of humanoid bipedal locomotion~\cite{ParraMoreno2023ImitationMF} a decent summary.}

\subsection{Skill Learning: Imitation from Human Data}
\label{sec:human}

While robot experiences can serve as a reliable data source, collecting loco-manipulation data directly from robots remains a formidable challenge. Collecting teleoperation data, even though it is one of the most commonly used approaches, is costly and tedious to scale. 
Additionally, gathering robot data by deploying existing model-based methods or trained policies presents additional difficulties. Deploying these methods on hardware raises safety concerns. On the other hand, data collected from a simulator or synthesized from a model suffers from sim-to-real gaps. Furthermore, model-based methods based on human knowledge (\textit{e.g.}, dynamics models, heuristic trajectories) generate consistent but similar behaviors, leading to limited data diversity. 

Learning from a large, diverse corpus of human data mitigates these challenges, as recording human data is more accessible and scalable due to the reflexive usage of loco-manipulation skills by humans in their daily lives. Furthermore, training policies to imitate human data can simplify the synthesis of loco-manipulation behaviors. Recent research efforts in 3D human motion data archival have surged in the vision and computer graphics communities. As shown in Fig.~\ref{fig:learn_from_human}, there are currently two primary approaches to \revised{acquiring} 3D human motion data: (i) recording directly from motion capture systems and (ii) reconstructing from 2D videos. 

\subsubsection{Obtaining Human Data}

\begin{figure}[!t]
    \centering
    \includegraphics[width=0.85\columnwidth]{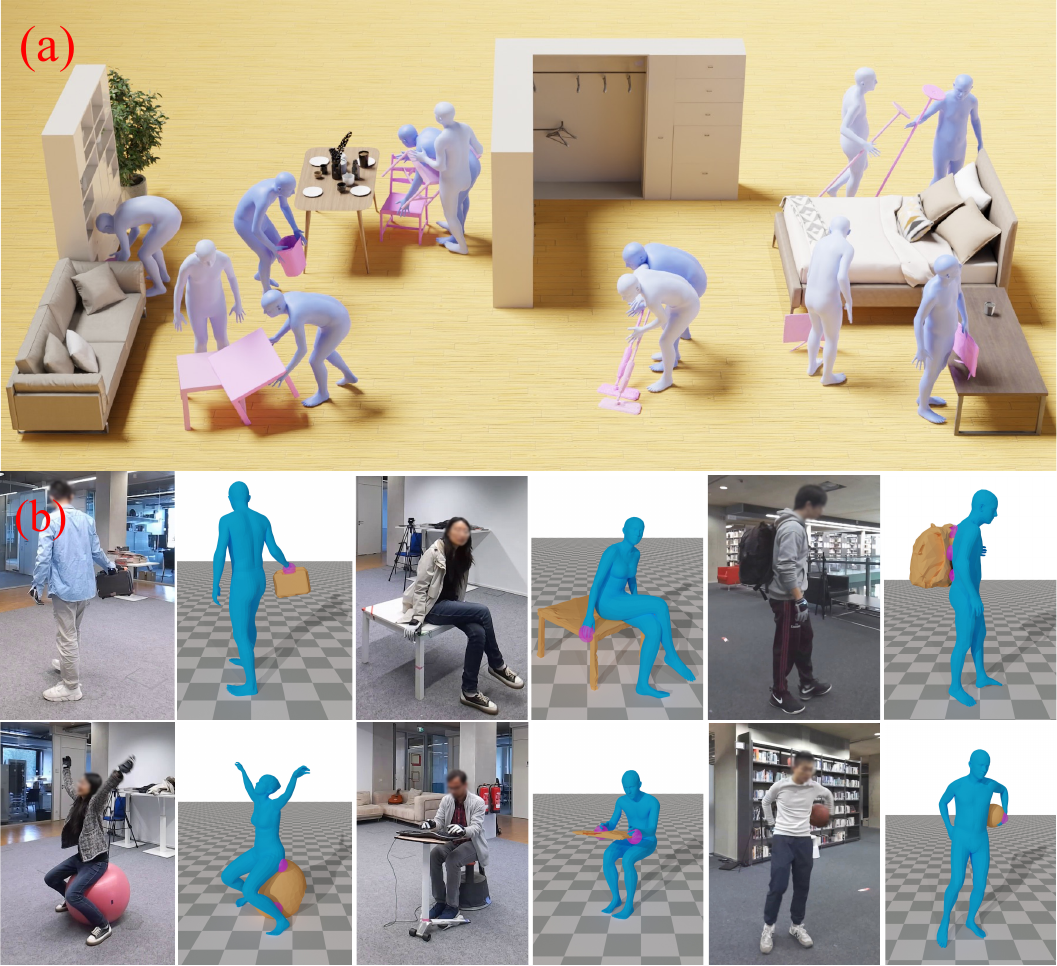}
    \caption{Motion capture of human data. (a) The human interacts with household items \cite{object_guided_motion_tog2023}. (b) The human interacts with items with whole-body manipulation \cite{behave}.}
    \label{fig:human_data}
    \vspace{-0.1in}
\end{figure}

Various tracking systems are used to obtain human motion data.  
As shown in Fig.~\ref{fig:human_data},~\cite{behave, object_guided_motion_tog2023} captures humans interacting with various objects while moving around. The following datasets, CMU~\cite{CMU_dataset}, SFU~\cite{SFU_dataset}, LAFAAN~\cite{LAFAAN_dataset}, and AMASS~\cite{AMASS}, are commonly used because they provide a wide variety of human motions.
However, motion capture data often require heavily instrumented environments and actors, making them expensive to scale, and indoor lab settings provide little exposure to outdoor activities.

Alternatively, videos and images obtained from the Internet offer a rich and diverse source of human motion data, including athletic performances, artistic dances, or daily chores. However, 
motion extracted from internet data is usually of lower quality, containing noise, jittery, and non-physical artifacts due to occlusion and motion blurs. 
Therefore, the reconstruction of accurate 3D human poses from 2D data remains an active research topic in the computer vision community. 
\begin{figure}[!t]
    \centering
    \includegraphics[width=0.98\columnwidth]{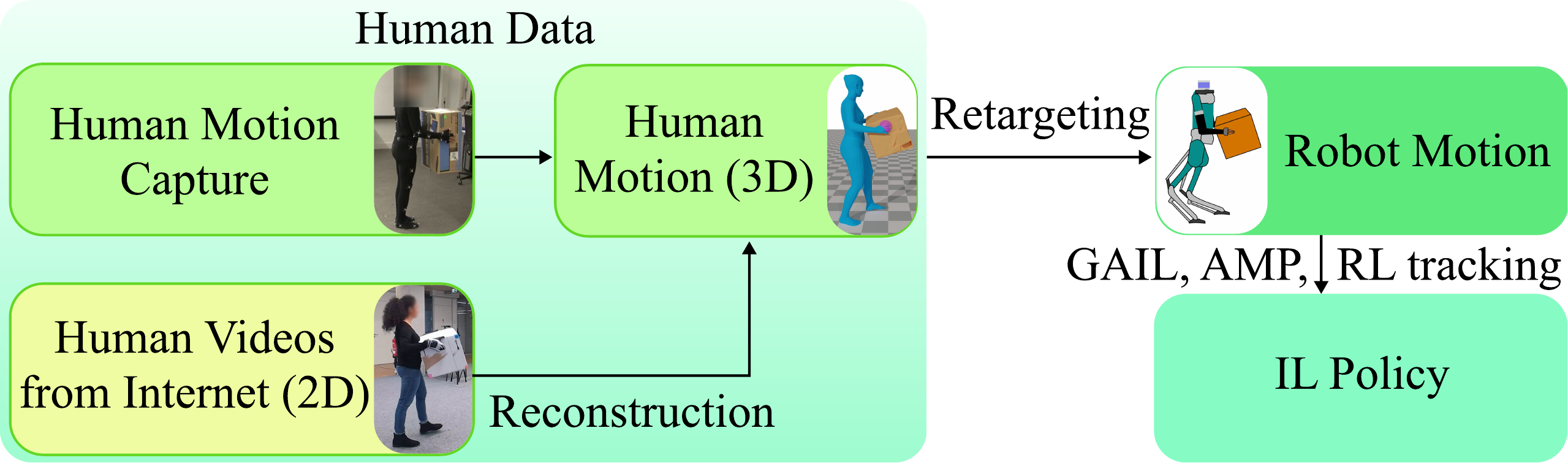}
    \caption{A pipeline for learning from human data. The 3D human motion data can be recorded from motion capture systems or reconstructed from 2D human videos. Robot motion is retargeted from 3D human motion for imitation learning.}
    \label{fig:learn_from_human}
    \vspace{-0.1in}
\end{figure}

Animation is also a widely explored approach to obtaining human motion data. Although the approach is effective in designing expressive motions for virtual human characters, the process often requires the use of sophisticated animation tools by professional animators~\cite{VarinThesis}, which makes it less scalable than motion capture or Internet videos. To address this limitation, researchers have been pursuing \textit{motion generation}, leveraging human data to generate diverse and realistic motions. \revised{Recent advances in generative models, particularly diffusion-based approaches for video generation such as Sora~\cite{sora} and Stable Video Diffusion~\cite{blattmann2023stable}, have significantly propelled progress in motion generation. A comprehensive survey~\cite{Motion_Generation_Survey} highlights the impressive capability of these models to generate realistic human motions.} However, our focus here is training IL policies to generate \textit{physically plausible} actions that achieve the motions demonstrated in human data. The training pipeline is shown in Fig.~\ref{fig:learn_from_human}.

\subsubsection{Challenges in Learning from Human Data}
Using offline human motion data (from any source) to train humanoids inevitably creates an embodiment gap both in observation and action spaces due to different body proportions, joint configurations, and mass distributions between most humanoids and humans. 
Closing this embodiment gap requires retargeting, which involves mapping the motion collected from a source skeletal model to a target robot model. \solution{Previous work in computer graphics and robotics has explored various retargeting strategies, such as joint-space~\cite{penco_2018_retarget, humanplus, cheng2024expressive} and task-space correspondence~\cite{cheng2024tv}, contact points~\cite{Otani2017Retarget}, fingertips~\cite{ fabisch_modular_2022}, gait synchronization~\cite{ramos2018humanoid,dallard2023ral}, and motion feasibility filter~\cite{he2024omnih2o, rouxel2022multicontact}.}
Developing systematic solutions for retargeting the entire human body, including dexterous hands, remains a critical topic for advancing humanoids.

Another correspondence problem is that the human data are only proprioception-based, which lacks sensory input and action output. 
Notably, these human data lack tactile or force measurement from interactions, which limits the capability of learning for loco-manipulation with rich physical interaction. 
\solution{To solve this issue, IL trains control policies that track references within a physical simulator.} Specifically, IL control policies accept state-only references instead of state-action pairs and output control signals. The physical simulator plays a key role in providing sensory input and validating the physical feasibility of the policy action.  


\subsubsection{Approaches to Learning from Human Data}
Examples of learned human-like motions include walking using Generative Adversarial Imitation Learning (GAIL)~\cite{gail_locomotion, gail_stand} \revised{and Adversarial Motion Prior (AMP)~\cite{AMP,tang2024humanmimic,zhang2024wholebody}, which is 
an extension of the GAIL}. Recently, RL-based motion imitation using motion capture data has successfully transferred to humanoid robots~\cite{cheng2024expressive, he2024omnih2o, humanplus,dugar2024DigitMHC}. However, most of these works achieve only relatively conservative loco-manipulation behaviors. Highly agile behaviors, as shown in DeepMimic~\cite{DeepMimic2018}, still exist only in simulators, and similar capabilities have yet to be replicated with real robots.

Human motion imitation can also achieve robust policies capable of rich interactions with objects in unstructured environments. For example,~\cite{peng2018sfv, yu2021human} demonstrates the learning of full-body gymnastic skills for humanoids in physics simulation, utilizing video reconstruction data and RL-based motion imitation.~\cite{xie_box_locomanip, hassan-box_carry} learns loco-manipulation policies enabling simulated humanoid characters to carry boxes. 
Furthermore,~\cite{physhoi} mimics motion capture data to achieve tasks such as playing basketball and grasping objects. However, many of these approaches rely on privileged information, such as the ground truth of object and ego poses, from the simulator, which limits their extension to real-world hardware. 


\revised{\textit{Key Takeaways:} Transferring skills from humans, especially through Internet-scale datasets, unlocks a broad range of loco-manipulation capabilities for humanoid robots. Although learning from human data holds great promise, a large body of what has been achieved in simulation has yet to be realized in real-world robots. Developing affordable and capable humanoid robots and high-fidelity simulators of real-world loco-manipulation scenarios can accelerate progress in this research direction. As we witness an increasing accessibility of humanoid robots in the market, we foresee that imitation from human data will enable humanoid robots to acquire a vast and diverse skill set. This skill set is fundamental to building humanoid foundation models discussed in Sec.~\ref{sec:foundation_model}.}

\subsection{Skill Learning: Hybrid Methods}
\label{sec:combine_method}

\begin{figure}[!t]
    \centering
    \includegraphics[width=0.85\columnwidth]{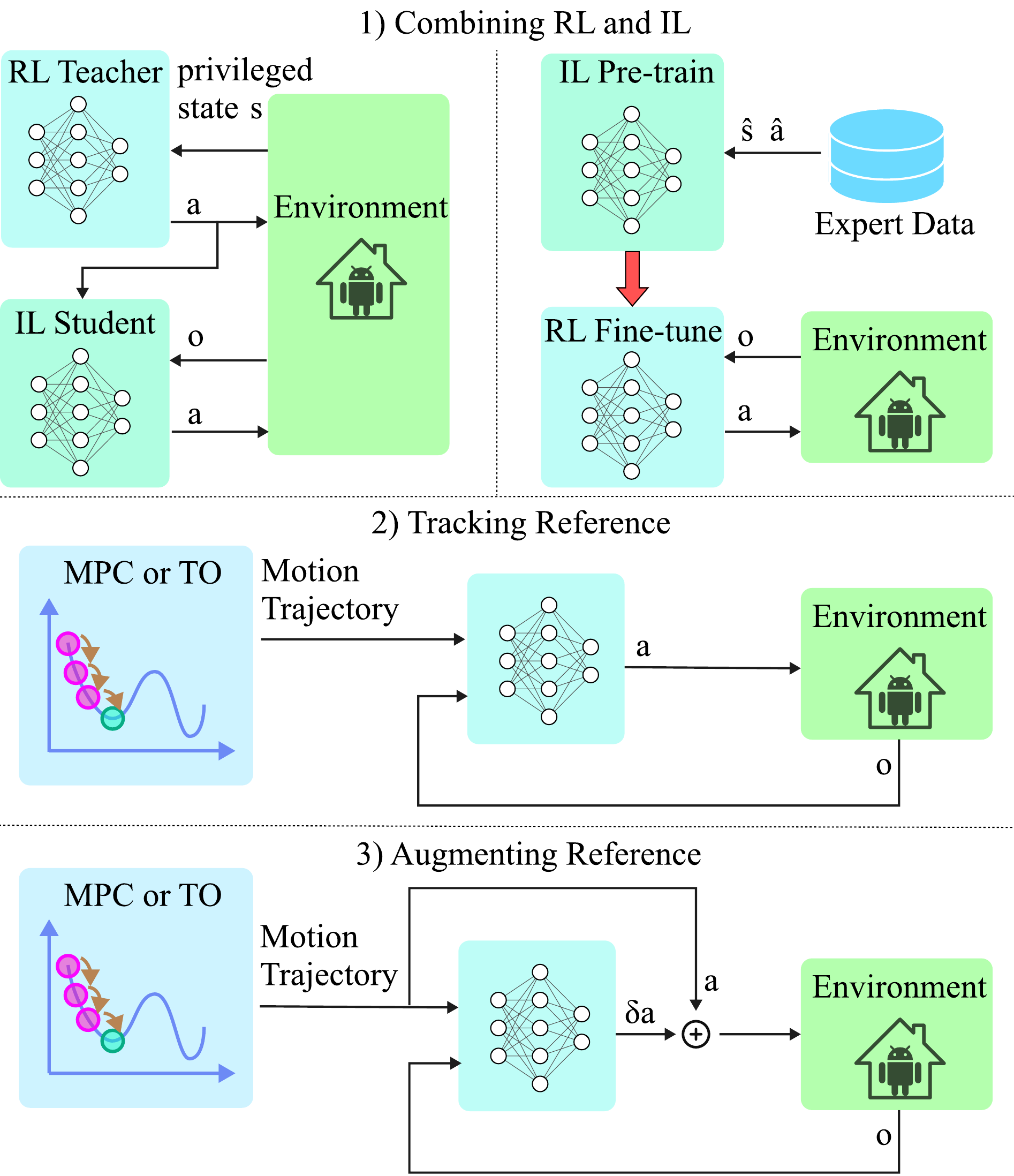}
    \caption{The frameworks of combining model-based methods and model-free methods for skill learning.}
    \label{fig:combine_methods}
    \vspace{-0.1in}
\end{figure}

Approaches that combine learning-based methods (IL and RL) and model-based methods are illustrated in Fig.~\ref{fig:combine_methods}.

\subsubsection{Combining Pure RL and IL} The combination of IL and pure RL without demonstration data has led to effective sim-to-real transfers. A two-stage teacher-student paradigm is widely adopted~\cite{humanoid_parkour_learning, transformer_digit_RL, wu2024learn}. In these works, a teacher policy is first trained from simulated privileged observation using pure RL. Then, a student policy clones the behavior of the teacher, achieving a similar performance using only partial observations. The trained student policy is readily deployable on hardware with onboard observations. Another two-stage paradigm~\cite{IFM_quad} reverses the order of the two policies. It first uses IL to pre-train an imitation policy from expert data and then uses an RL policy to fine-tune the imitation policy, achieving performance beyond the IL expert and adapting to different environments or tasks~\cite{Smith2023LearningAA}. 

\subsubsection{Learning to Track Trajectory Reference}
Combining model-based methods with learning-based methods is extensively explored.
Numerous works on quadruped robots~\cite{jenelten2024dtc, RL_MPC_RAL2023} use MPC to generate reference motions and use them as imitation rewards for RL-based motion imitation. 
However, calculating MPC online can prolong the training time and occasionally fail to find a feasible solution. Online RL that tracks \textit{offline-generated} trajectories avoids this problem and has achieved effective RL policies for versatile bipedal locomotion~\cite{GPS, li2023robust, marew2024soccerkicking}. Supervised learning is also used to mimic TO-generated offline motions. For example,~\cite{LearnCDM2020} allows humanoid robots to achieve brachiation on monkey bars. 
In addition to enhancing learning methods with model-informed trajectories, a learned policy can also suggest effective warm starts or hyperparameters for model-based methods, thus significantly reducing the iterations required for convergence to an optimal trajectory. 

\subsubsection{Learning to Augment Trajectory Reference} 
Rather than tracking the reference trajectory, augmenting the reference with a residual is another popular approach. Early work on dynamic movement primitives~\cite{Ijspeert_2002} modifies the reference trajectory by learning a task-specific force output, which achieves a humanoid racket swing. A milestone in trajectory augmentation is the demonstration of bipedal locomotion in~\cite{Xie_2018_Cassie}. Since then, there have been more studies on trajectory augmentation with successful dynamic locomotion~\cite{singh2023Access, OPT-Mimic}. Recently, loco-manipulation skills are also achieved through augmenting TO-based reference trajectory~\cite{liu2024opt2skill}. 
Beyond joint-level reference, a policy can also augment the task-space reference. For example,~\cite{VictorIROS2022NeuralAdaptation} learn a locomotion policy to adapt the foot placement reference derived from inverted pendulum models.

Both imitation and augmentation of the reference trajectory share benefits and drawbacks. 
As a benefit, references expedite the learning process and provide an effective way of learning complex skills. 
However, both methods rely on predefined trajectories and, therefore, have limited potential to learn emergent, diverse behaviors. 


\revised{\textit{Key Takeaways:} This subsection summarizes hybrid methods with successful humanoid hardware deployments. Hybrid methods that combine learning methods with and without demonstration are effective for learning complex and robust behaviors. We also show that model-based methods are not in conflict with learning-based methods, as they can provide guidance for efficient learning. Overall, hybrid methods that combine learning-based and model-based approaches have achieved superior efficiency, versatility, and performance in humanoid tasks that outperform single methods alone.}

\subsection{Representation and Composition for Versatile Skills}
\label{sec:skill_composition}
\begin{figure}[!t]
    \centering
    \includegraphics[width=0.95\columnwidth]{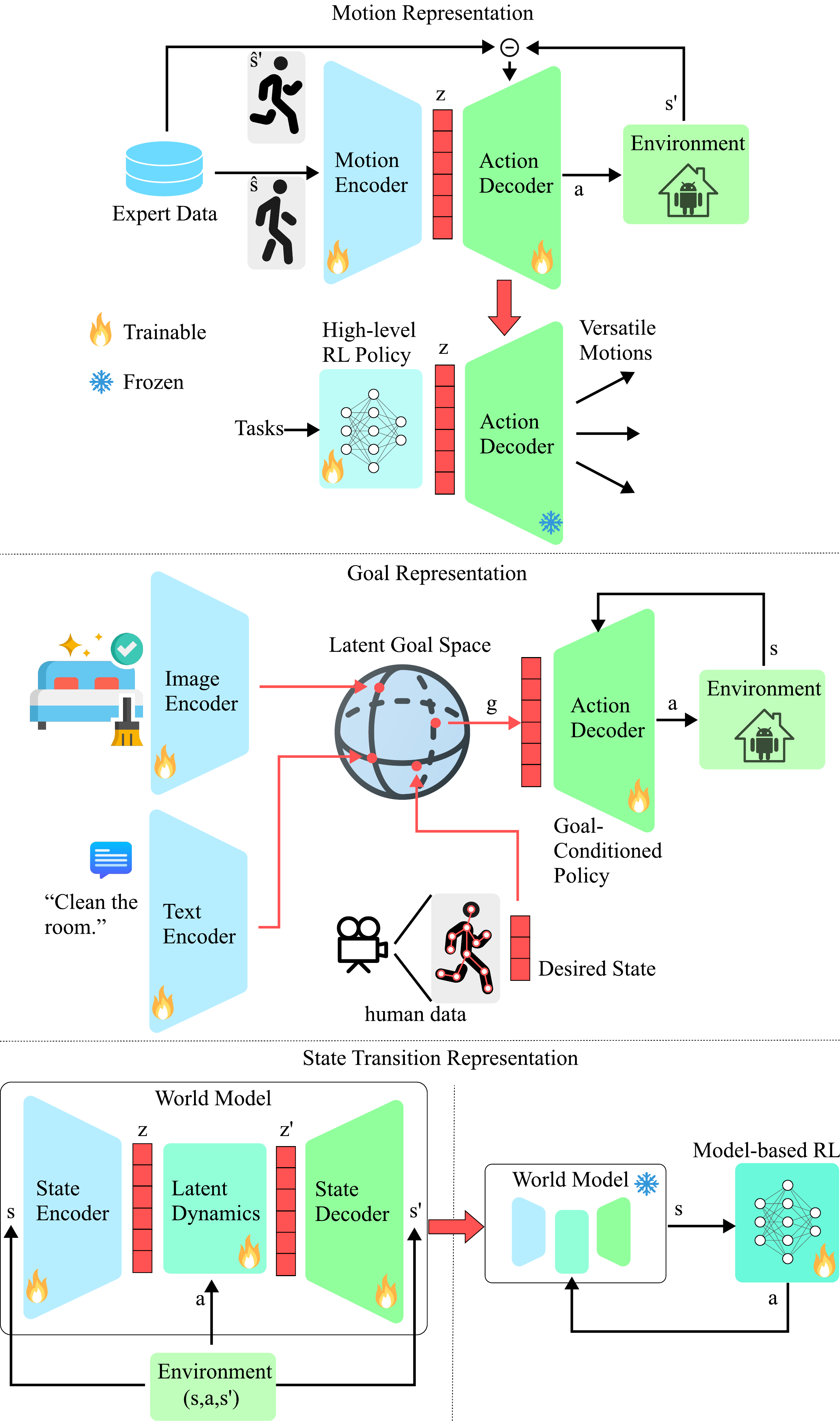}
    \caption{Implicit skill representations. (a)~Represent motions in latent space. (b)~Represent goals in latent space and instruct a goal-conditioned policy for execution. (c)~Represent state transitions with a world model and then use the learned world model for sample efficient training.}
    \label{fig:combine_skills}
    \vspace{-0.1in}
\end{figure}

A good representation of skills makes it easy to compose tasks. In general, a skill can be represented explicitly as a state-action trajectory that accomplishes a task or implicitly as a network structure and its learned weights. In this subsection, we explore implicit representations for skill composition to achieve versatile and multi-skill tasks. 
Recent learning-based approaches enable smooth transitions between multiple skills. Among them, Mixture of Experts (MOE) is widely used. MOE employs a hierarchical architecture; it first trains multiple distinct skills, implicitly encoded in low-level expert policies, and then learns a high-level policy to select~\cite{Coros09} or blend~\cite{RAL_MOE_Humanoid} these expert networks. This architecture allows for smooth transitions between skills and facilitates the completion of diverse tasks. 
However, MOE encounters expert imbalance issues favoring certain experts while degrading others, which could diminish the diversity provided by the experts. 
Instead of obtaining and then blending \textit{multiple} policies, structured representations improve memory efficiency and allow a \textit{single} policy to achieve multiple tasks. Next, we introduce three well-received structured representations: motion representation, goal representation, and state transition representation, all shown in Fig.~\ref{fig:combine_skills}.

\subsubsection{Motion Representation}
Motion representation extracts the essential features and temporal dependencies of high-dimensional long-horizon motions~\cite{li2024fld}. Specifically, motion representation encodes high-dimensional motions in a low-dimensional latent space. 
Such latent-space representations are commonly learned in an unsupervised manner using generative models such as Variational Autoencoders (VAEs)~\cite{VAE} and Generative Adversarial Networks (GANs)~\cite{GAN_Goodfellow}. Compared to VAEs, GANs have greater potential to generate realistic motions following the reference data distribution, but they are often difficult to train. 
The result of learning motion representation is often a model that can synthesize versatile motions given latent codes. Therefore, the generative model can be reused for new downstream tasks by pairing it with a high-level task-specific policy. For example, several studies~\cite{bohez2022imitaterepurposelearningreusable, Tessler2023CALM, luo2024universal} learn a high-level RL policy to efficiently use a reduced-dimensional latent space motion representation and allow simulated humanoid robots to follow a set of user-commanded tasks. 

\subsubsection{Goal Representation}
Another approach to learning a \textit{single} policy for multiple tasks is through the representation of goals. The goal is typically represented as a feature vector, which can be encoded from an image of the scene in its final state, a natural language instruction, or a desired state from observing human demonstrations. This goal representation is often paired with Goal-Conditioned Policies (GCPs)~\cite{ijcai2022GCRL_survey}. Unlike standard RL policies that achieve only one task, GCPs achieve multiple tasks within a single general policy conditioned on different goals. 
GCPs have demonstrated versatile humanoid skills using IL, such as diffusion-model based BC~\cite{huang2024diffuseloco} and RL with imitation objectives~\cite{cheng2024expressive, he2024omnih2o, VMP}. 

\subsubsection{State Transition Representation}
The latent space can also represent the transition dynamics of a Markov Decision Process (MDP). In this representation, data collected from the MDP are used to train a dynamics model that predicts the transition probability between abstract, compressed representations of MDP states. This learned dynamics model is often referred to as a world model~\cite{world_model_Ha}. 
Sampling from a world model yields imaginary data, which can be achieved efficiently in parallel. By leveraging the imaginary data, Model-Based RL (MBRL) achieves greater efficiency compared to a typical model-free setting, where interaction data must be obtained from a simulated or real environment. MBRL has shown success in agile motor skills on humanoids~\cite{Hester2010, Gu-RSS-24}. 
Another approach, TD-MPC2~\cite{hansen2024worldmodelsvisual}, uses the world model in an MPC fashion, planning actions that lead to imaginary trajectories with high scores. 
Beyond data efficiency, the world model can mitigate the sim-to-real gap by fine-tuning a small batch of real-world data~\cite{shi2024efficientmodelbasedapproachlearning}.


\revised{\textit{Key Takeaways}: Enabling robots to accomplish versatile skills and multiple tasks is one of the main trends in robot skill learning. Whereas obtaining and blending single-skill policies is widely explored, more recent methods put efforts into achieving multiple tasks in a \textit{single} policy. This requires a more structured representation of the skill motion, the task goal, and/or the environment dynamics. Latent space models, goal-conditioned policies, and world models are promising approaches toward this objective. However, many of these methods are still limited to the computer graphics community and yet to be implemented on humanoid hardware. }

\subsection{Learning for Humanoid Loco-manipulation} 

\begin{figure}
\centering
\includegraphics[width=0.35\textwidth]{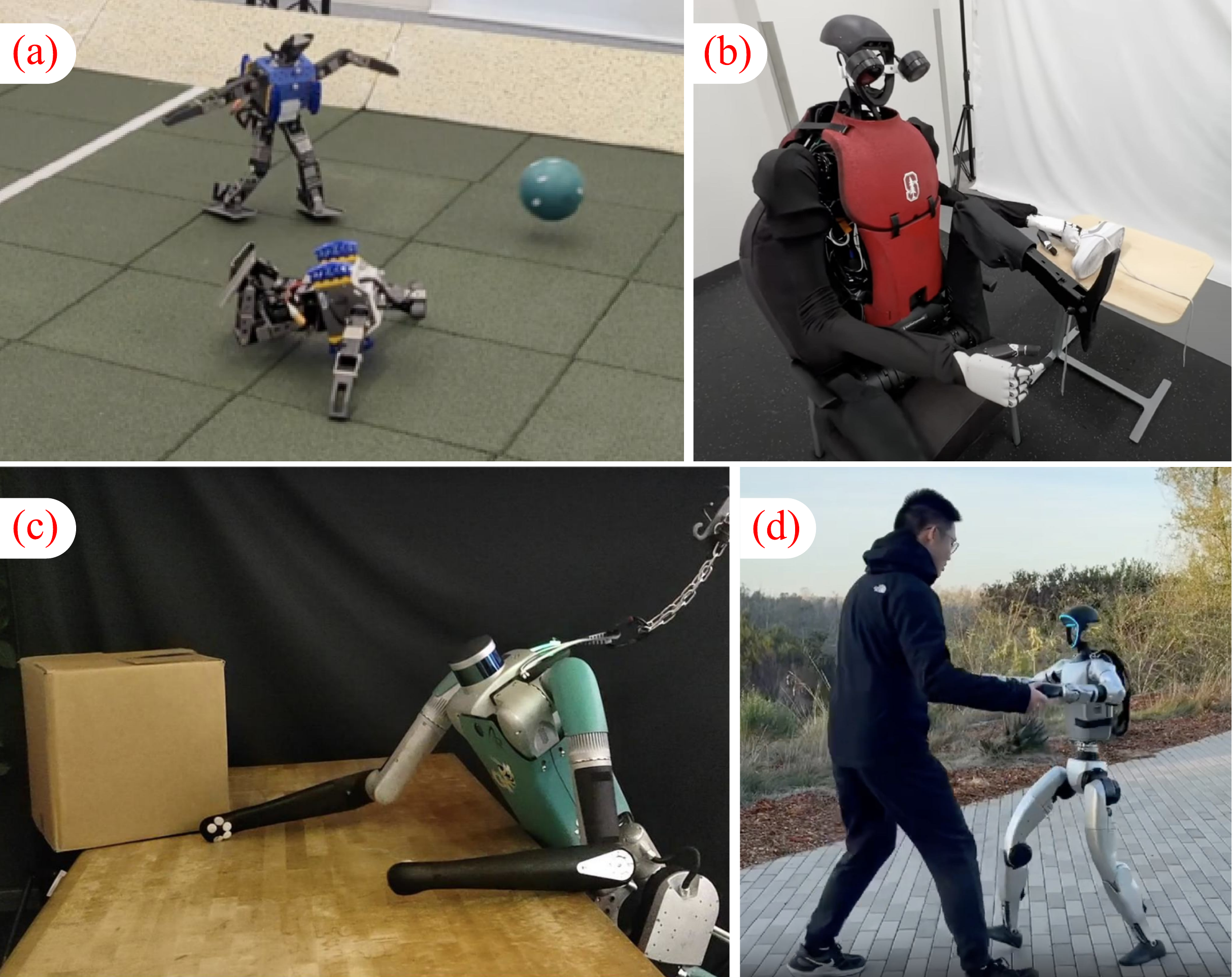}
\caption{Learning-based methods for humanoid loco-manipulation skills. (a) Getting up and chasing a ball~\cite{SR_deepmind_soccer}. (b) Tying shoelace~\cite{humanplus}. (c) Multi-contact box manipulation~\cite{liu2024opt2skill}. (d) Dancing with a human~\cite{ji2024exbody2}.}
\label{fig:learning_locomanip}
\vspace{-0.15in}
\end{figure}

Loco-manipulation skills are challenging for learning-based methods as they often struggle with achieving physically stable contact or precise contact forces. While many learning-based approaches, such as CooHOI~\cite{gao2024coohoi}, demonstrate humanoid loco-manipulation skills in simulation, the physical interactions with external environments or objects are often oversimplified. As a result, only a few studies have demonstrated sim-to-real transfer for loco-manipulation skills. 
A large part of these studies rely on sim-to-real RL, with a few examples shown in Fig.~\ref{fig:learning_locomanip}. However, sim-to-real RL for loco-manipulation typically involves complicated reward designs that are fine-tuned for specific environments and tasks~\cite{dao2023simtoreal, munn2024dynamic_throwing}, see Sec.~\ref{sec:RL}. To enable loco-manipulation tasks, these methods define the contact sequence either implicitly using reference trajectory~\cite{dao2023simtoreal, liu2024opt2skill} or explicitly via reward design~\cite{zhang2024wococo}, thus increasing the success rate of sim-to-real transfer. To address the uncertainty of the mass and other properties of the manipulated object, most RL approaches rely on domain randomization to provide robustness to object parameters~\cite{transformer_digit_RL}.

A hierarchical RL structure that manages distinct skills is a viable strategy for achieving various loco-manipulation tasks such as falling recovery and ball kicking~\cite{SR_deepmind_soccer}. However, hierarchical RL has scalability issues, see Sec.~\ref{sec:skill_composition}. On the other hand, IL methods, particularly those employing RL with motion imitation, have enabled humanoid loco-manipulation via teleoperation, as outlined in Sec.~\ref{sec:expert}. Although teleoperation is not autonomous, it is a crucial intermediate step to collect humanoid data. This has led to promising developments in \textit{autonomous} loco-manipulation skills~\cite{he2024omnih2o, humanplus}, with the potential for further expansion to more diverse loco-manipulation tasks. 

\revised{\textit{Key Takeaways}: Although learning-based methods for humanoid loco-manipulation are less developed than model-based methods, their significance should not be overlooked. 
Learning-based methods are potentially more robust, as they can adapt to unstructured scenarios that model-based methods struggle to explicitly address, such as recovering from a fall in an arbitrary configuration~\cite{fall_recovery}. 
Moreover, learning-based methods can find emergent behaviors that are challenging for model-based methods~\cite{Curiosity-schwarke23a}. 
The recent success of quadrupedal loco-manipulation~\cite{liu2024visualwholebodycontrollegged, he2024learningvisualquadrupedallocomanipulation,  portela2024learning} shows significant promise for humanoids. However, transferring these algorithms from quadruped to humanoid is challenging because of the more complex dynamics of humanoids, which require enhanced safety measures and precise balance control.}

\subsection{Challenge in Lack of Benchmarks for Loco-manipulation}
\label{sec:benchmark}
The development of humanoid loco-manipulation skills is still in its infancy compared to more established tasks such as humanoid locomotion~\cite{LocoMuJoCo} and tabletop manipulation~\cite{liu2023libero}. \revised{To advance this emerging field, there is a pressing need for large-scale, systematic benchmarks specifically designed for humanoid loco-manipulation, encompassing both simulation and real-world environments. Benchmarks that define loco-manipulation tasks and evaluation metrics can significantly accelerate research in this field.} 

For benchmarks in simulation, HumanoidBench~\cite{sferrazza2024humanoidbench} and Mimicking-Bench~\cite{liu2024mimickingbench} 
offer a list of loco-manipulation tasks\revised{, such as sitting on a piece of furniture and wiping a window.} SMPLOlympics~\cite{luo2024SMPLOlympics} presents a collection of sports as tasks for simulated humanoids. 
These benchmarks provide metrics for evaluating the performance of humanoid algorithms and help verify the reproducibility within MuJoCo~\cite{todorov2012mujoco} and Isaac Gym~\cite{Orbit}. \revised{However, these tasks represent only a subset of human skills, and the evaluation metrics remain highly task-specific. Expanding the benchmarks to include a broader range of real-world tasks, along with more standardized metrics, would enhance their utility and practicality. Moreover, the sim-to-real gap continues to limit the practical value and reliability of these simulation benchmarks: it remains uncertain how well performance in simulation translates to real-world deployment, although promising results have been demonstrated in zero-shot sim-to-real transfer.} 

For hardware benchmarks, the development of affordable and capable humanoid robots as standardized platforms for hardware evaluation can significantly accelerate the research. \solution{Research efforts on open-source humanoid hardware and software, such as Hector~\cite{li2023dynamic}, the MIT Humanoid~\cite{chignoli2021humanoid}, the Berkeley Humanoid~\cite{liao2024berkeleyhumanoid}, and the Duke Humanoid~\cite{xia2024duke} represent valuable contributions.} \revised{Additionally, rapid prototyping of diverse humanoid robots facilitates cross-embodiment generalization, while mass production of standardized robots streamlines the deployment of benchmarks, such as through the Robotic Grand Factory initiatives~\cite{Q_Humanoid}.} \revised{Complementing these efforts, the construction of remote-accessible real-world environments, as exemplified by the DARPA Robotics Challenge, Robotarium \cite{robotarium}, and TriFinger~\cite{TriFinger}, enables researchers to focus on advancing algorithms. Ideally, these physical environments share the same benchmark metrics as the simulators to facilitate sim-to-real transfer, while remaining highly customizable to support novel tasks.} 


\subsection{Challenge in Data Scarcity}
\label{sec:data_scarcity}
As discussed in Sec.~\ref{sec:expert}, the four data sources present a trade-off between quality and availability. 
The lack of high-quality large-scale robot data becomes a bottleneck for robot skill learning. \solution{To solve the bottleneck, much effort has been put into data scaling.} 
There is ongoing debate about whether scaling is the best approach to generalist humanoid robots.

The central question we must answer is which aspect of human motion we want robots to learn. Some humanoid tasks can be achieved simply by mimicking 3D human pose trajectories~\cite{dallard2023ral,humanplus,cheng2024expressive,he2024omnih2o,okami2024}, but a true general-purpose robot emerges from purposive learning: the ability to identify meaningful intentions from human data and adapt past experiences to new tasks or environments~\cite{Purposive_learning_Cheng2019}. Therefore, human data must teach the robot not only \emph{what} humans are doing, but also \emph{how} and \emph{why} they are doing it. Current data acquisition methods that capture human joint poses only enable learning what humans are doing. In this regard, imitation of human data at the trajectory level is not fundamentally generalizable due to the inevitable gap in morphology and in the surrounding environment. 

\revised{We argue that generalization in loco-manipulation is achieved by including the motion of the manipulated objects, enhanced with a greater variety of sensing modalities instead of data quantity scaling. \solution{To develop truly versatile and adaptive humanoids, human data should also include that of the manipulated objects, cognitive actions (\textit{e.g.}, trust, compete, collaborate) paired with multimodal observations (\textit{e.g.}, whole-body haptic sensing, egocentric images), so that humanoids can learn the `how' and `why'.} However, instrumenting the environment and the manipulated objects with force and tactile sensing is highly complex and difficult to scale. \solution{Potential approaches include leveraging vision-language models to extract information from videos~\cite{radford2021learning,alayrac2022flamingo, mao2024learning}, employing video generation models~\cite{ho2022imagen} or physics-based simulation~\cite{todorov2012mujoco, Orbit} to generate synthetic data, and adopting hybrid strategies that effectively combine these synthetic data sources with limited real-world multimodal data~\cite{gr00tn1, humanoidpolicyhuman}. Recent work aims to bridge the gap between human animation and humanoid applications by collecting human multimodal observations~\cite{NEURIPS2022_ActionSense,grauman2024egoexo4dunderstandingskilledhuman} and human kinetics~\cite{AddBiomechanics}.} Together with the rapid advancement in humanoid hardware, purposive learning with more informative human data will become the mainstream approach to achieve versatile and general-purpose humanoids.}

\section{Foundation Models for Humanoid Robots}
\label{sec:foundation_model}

Foundation Models (FMs) are large pre-trained models using Internet-scale data~\cite{bommasani2021opportunities}. Recent progress in FMs such as Large Language Models (LLMs) and Vision-Language Models (VLMs) has demonstrated groundbreaking capabilities in solving a wide range of downstream tasks (through in-context learning or fine-tuning), such as code generation and video understanding~\cite{openai2023gpt4}. The common sense reasoning capabilities of FMs have inspired many explorations of their applications in robotics~\cite{firoozi2023foundation,  Robots_via_Foundation_Models}. Despite this growing interest, research on FMs specifically for humanoid robots remains sparse. In this section, we first provide an overview of FMs in the context of general robotics (\textit{e.g.}, mobile manipulation, instead of humanoids) and then explore their potential applications to humanoid robots. 

\begin{figure}
\centering
\includegraphics[width=0.48\textwidth]{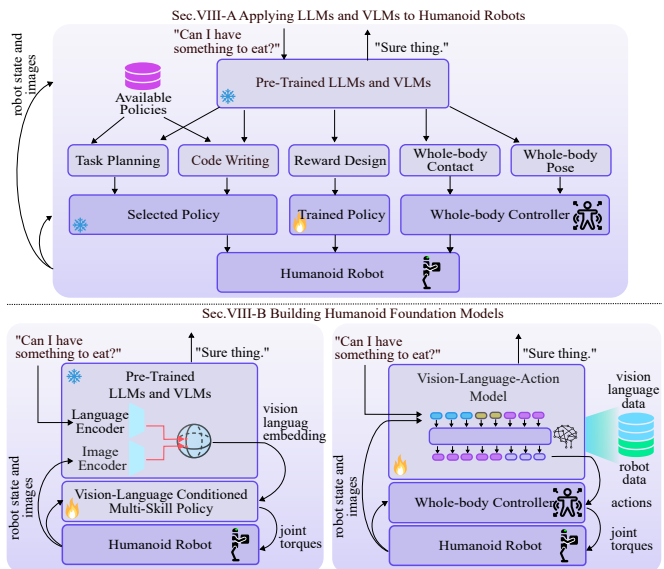}
\caption{Sec.~\ref{sec:foundation_model_for_humanoid} details different approaches to applying LLMs and VLMs in robotics tasks. These approaches prompt pre-trained LLMs and VLMs to generate task-relevant intermediate representations that can be executed by low-level policies or controllers. Sec.~\ref{sec:humanoid_foundation_model} presents two approaches to building humanoid foundation models. The first approach uses robot data to train multi-skill policy, and the second approach fine-tunes an existing VLM. Both approaches directly output actions for low-level humanoid control.}
\label{fig:FM_figure}
\vspace{-0.15in}
\end{figure}

Strategies to leverage FMs for robotics can generally be categorized into two, as shown in Fig.~\ref{fig:FM_figure}. The first strategy (Sec.~\ref{sec:foundation_model_for_humanoid}) elicits actionable knowledge from pre-trained LLMs/VLMs for robotic tasks, without additional model fine-tuning. The second strategy (Sec.~\ref{sec:humanoid_foundation_model}) collects abundant robotic data to fine-tune or co-train a \textit{Robot Foundation Model} (RFM) that generalizes to control tasks with common sense reasoning capability~\cite{brohan2022rt, brohan2023rt, jiang2023vima, kim2024openvla, gemini_robotics_2025, bu2025agibot}.

\subsection{Applying LLMs/VLMs to Humanoid Robots}
\label{sec:foundation_model_for_humanoid}
Applying LLMs/VLMs to humanoids is still a nascent field. Many studies have demonstrated the successful deployment of LLMs/VLMs across various robot embodiments, such as dexterous hands~\cite{ma2023eureka}, manipulators~\cite{liang2023code}, mobile manipulators~\cite{saycan2022arxiv}, quadrupedal robots~\cite{ma2024dreureka}, and bipedal robots~\cite{yao2024anybipeendtoendframeworktraining}. 


Among these works, a majority way of using LLMs/VLMs is to leverage pre-trained models without robot data. Although these pre-trained models have semantic understanding capability and context awareness, they often lack embodied knowledge and can prescribe actions that are ambiguous or non-admissible. Therefore, considerable research efforts have focused on task-planning mechanisms to enable the generation of admissible action plans. 
For example, SayCan~\cite{saycan2022arxiv} ranks available actions based on value functions obtained during the training of corresponding action policies for mobile manipulators. VLM-PC~\cite{chen2024VLM-PC} restricts GPT-4o to output plans with skills available only for quadruped navigation. 

\subsubsection{\revised{FMs for selecting low-level skills (task planning)}} Such task planning capability of FMs has advanced the complexity of tasks that a humanoid can accomplish. For instance, ~\cite{PADL} and~\cite{Prompt_Plan_Perform} use pre-trained LLMs to select skills and task goals for animated humanoid characters. OmniH2O~\cite{he2024omnih2o} employs GPT-4~\cite{openai2023gpt4} to select autonomous skills such as greeting a human. HYPERmotion~\cite{wang2024hypermotion} applies an LLM to construct task graphs that enable a hybrid wheeled-leg robot to execute complex loco-manipulation tasks. However, using FMs with a fixed skill set limits skill versatility. In addition, for complex behaviors such as those in humanoid loco-manipulation tasks, it is essential to allow FMs to author the detailed motions of low-level skills, instead of selecting from the existing, relatively-abstract low-level skills.


\subsubsection{\revised{FMs for authoring low-level skills}} Thus, many research efforts have focused on identifying the best bridge between FMs and low-level robot skills. For example, researchers have proposed to generate code~\cite{liang2023code, ProgPrompt2023} and reward functions~\cite{yu2023language, ma2024dreureka, yao2024anybipeendtoendframeworktraining} as intermediate representations for bipedal and quadrupedal robots. Compared to selecting the existing skills, these intermediate representations provide additional flexibility in adjusting the generated motion. Furthermore, FMs can generate whole-body poses~\cite{jiang2024harmon, kumar2023Words_into_Action} and whole-body contacts~\cite{xiao2024unified, totsila2024words2contact} for humanoid robots. These techniques allow users to intuitively direct a robot's behavior through expressive inputs such as natural language, images, or even gestures. 

\subsection{Building Humanoid Foundation Models}
\label{sec:humanoid_foundation_model}


While most FMs are developed in the vision or language domain, building them in the robotics domain for embodied intelligence is a natural extension. Similar to LLM and VLM, Robot Foundation Models (RFMs) are large models trained from Internet-scale robotic datasets. These RFMs often process multimodal inputs (\textit{e.g.}, egocentric images, and natural language as task description) \revised{and enable interaction with the physical world through robot action outputs. Therefore, they are also called Vision-Language-Action (VLA) models}. Leveraging the Internet-scale multimodal dataset, RFMs hold the promise of generalization across diverse tasks and provide a natural interface for human-robot interaction, both essential for real-world robot applications.

\subsubsection{\revised{Hierarchical vision-language-conditioned policy}} One popular approach to building RFMs is to create a multi-task sensorimotor control policy using a high-capacity model that can consume large amounts of robot data. The trained policy can perform a wide variety of low-level skills, even for multiple robot embodiments. To enable high-level reasoning/planning, this approach often employs a hierarchical framework combining a pre-trained LLM or VLM with the low-level control policy, initialized from scratch. During training, this setup aligns the semantic knowledge of the LLM and VLM with the physical behavior of the control policy, enabling cross-modal capabilities such as language-to-action. {Transformer~\cite{Attention_Need} is a common choice for such a low-level policy due to its scalability.} To enable interactive user commands, the low-level policy is conditioned on language and/or image inputs, often tokenized with pre-trained text or image encoders from LLMs or VLMs. For example, RT-1~\cite{brohan2022rt} and VIMA~\cite{jiang2023vima} both train a language-conditioned visuomotor policy with a large amount of manipulation data and have demonstrated the ability to perform a wide range of skills. However, successful implementations are limited to robots that have stable dynamics and a large amount of high-quality data collected with significant resources~\cite{brohan2023rt, black2024pi0, open_x_embodiment_rt_x_2023, kim2024openvla}. 

\revised{Applying such a hierarchical framework to humanoid robots is an exciting new direction. Recent works have extended RFMs to the humanoid upper body \cite{gr00tn1, bu2025agibot, gemini_robotics_2025}. For example, NVIDIA developed GR00T N1 \cite{gr00tn1}, a general-purpose FM for humanoid robots, and Figure~AI demonstrates Helix, a hierarchical VLA for dexterous manipulation and collaboration between two humanoid robots.} However, building an RFM for humanoid loco-manipulation remains a challenging endeavor: the inherent instability in locomotion and high-dimensional action space in dexterous hands makes it difficult to collect high-quality data efficiently. Therefore, VLA for humanoid loco-manipulation is shown only in simulation. For example, HumanVLA~\cite{xu2024humanvla} and SuperPADL~\cite{SuperPADL} train a humanoid action policy aligned with the latent space of pre-trained VLMs, enabling skills based on image and language inputs.

\subsubsection{\revised{End-to-end Vision-Language-Action model}} To leverage prior knowledge in FMs, another popular approach is to build an end-to-end VLA model, as exemplified by RT-2~\cite{brohan2023rt}, OpenVLA~\cite{kim2024openvla}, and Gato~\cite{reed2022generalist}. 
This approach treats robot data (\textit{i.e.}, observations and actions) as tokens in the language model's vocabulary, allowing direct fine-tuning or co-training with existing VLMs. Unlike the previous approach, VLA outputs actions as tokens directly without relying on trainable low-level policies. The VLA model not only generates robot actions for diverse skills, but also retains semantic reasoning abilities in the language and vision domain, enhancing its generalization capacity compared to models trained in single domains. For example, RT-2~\cite{brohan2023rt} represents the actions as a string of numbers similar to the tokens from the pre-trained vision and language tokenizer of the base VLMs. However, representing actions with stringified numbers can be token-inefficient with high degrees-of-freedom robots such as humanoids. 

A key research question in building RFMs is the design of effective algorithms and model architectures. Although most of today's RFMs are based on autoregressive transformer models~\cite{Attention_Need}, their computational inefficiency over long sequences poses a significant challenge for both training and inference. This has driven the exploration of alternative models that are both efficient and high-capacity, such as state-space models~\cite{gu2023mamba}. 

The key to the success of training an RFM, especially in the case of VLA, heavily depends on the choice of input and output representations. Most RFMs take as input a combination of task descriptions in language, visual observations of the surrounding environment, and the history of robot states. Outputs typically consist of robot actions, which are derived from either a multi-task policy (the first approach) or an end-to-end VLA model (the second approach).
There are variations in the VLA model where its token outputs specify more than just actions. For example, Octo~\cite{octo} and RDT-1B~\cite{liu2024rdt1b} use the token output for a diffusion denoising process. $\pi_0$~\cite{black2024pi0} maps a learned token to robot actions through a diffusion head, enabling high-frequency control (up to $50$~Hz for a bimanual manipulator with a wheelbase). GR-2~\cite{cheang2024gr2} predicts tokens that represent future images and actions; thus it functions as both a world model and a visuomotor policy. For humanoid robots, input and output are not yet well defined. For tasks involving rich physical interactions, the force feedback is as crucial as egocentric visual input. Given the unstable dynamics of humanoid robots, a more practical approach is to use end-effector and body poses as action outputs from the RFM and adopt additional low-level policies to ensure balance and safety via high-frequency feedback control.
Incorporating robotic data as a new modality into state-of-the-art FMs would require significantly more data. Hence, designing effective input/output representation for humanoid loco-manipulation tasks still remains an open research question.


\subsection{Opportunities and Challenges in Foundation Models}
Integrating Foundation Models (FMs) into humanoid robots offers distinct opportunities and challenges. On the opportunity side, since the majority of data used to train FMs is collected by humans, the knowledge embedded in these models is inherently biased towards human-like embodiments. Consequently, humanoid robots could potentially utilize existing knowledge in FMs more effectively due to a smaller embodiment gap. This advantage extends beyond planning and control capabilities to include interactions with humans using natural modalities such as language and gestures. 


A major challenge in applying FMs to humanoids arises from the high inference cost. Running large foundation models using only onboard computing is not feasible due to the limited power and computation, which hampers real-time hardware control. \solution{To address this challenge, several solutions have been proposed. One effective strategy involves adopting a decentralized hierarchy \cite{gemini_robotics_2025}, where FMs operate over the cloud and provide only high-level decisions at a lower frequency, while another controller remains onboard and manages real-time task execution.} However, the inference delay and internet latency may impede the control performance. Another approach is to enhance the speed of the computing platform and the efficiency of FMs. \solution{For instance, NVIDIA introduced Jetson Thor, an onboard computing platform designed for humanoid robots. 
Google proposed SARA-RT~\cite{leal2023sarart}, which accelerates the model speed without compromising its quality.}

The training of FM is also resource and time consuming. For example, training the LLaMA model took $34$~days on $992$~NVIDIA~A100-80B GPUs~\cite{touvron2023llama}, which incurs high cost, high energy consumption, as well as carbon dioxide emission. As FMs continue to scale up, the training cost increases further. \solution{A promising approach to maintaining a reasonable cost for training robotics FMs is to leverage parameter-efficient fine-tuning techniques. For example, OpenVLA leverages the LoRA technique to fine-tune an FM with a robotics dataset~\cite{kim2024openvla}.}

A critical component of the humanoid foundation model that has yet to be developed is a scaling law, similar to the training of large language models~\cite{kaplan2020scaling}. The scaling law provides guidance on how we should scale up model, compute, and data, to meet the desired performance in the most efficient way. A major research effort focuses on scaling the robotics dataset. Open X-Embodiment extends the idea to a much larger robotics dataset across various robot embodiments and tasks~\cite{open_x_embodiment_rt_x_2023}. Recent work has already explored a data scaling law for robot manipulation, with a focus on generalization capabilities~\cite{lin2024data}, as well as model scaling behaviors for zero-shot capabilities in action selection~\cite{nasiriany2024pivot}, which marks important initial steps towards this direction. 

\revised{Beyond computational limitations, FMs face additional challenges when deployed in humanoid systems. Safety constraints are particularly concerning, as models trained on internet data may generate unsafe actions when controlling physical robots that are inherently unstable. \solution{Establishing a thorough evaluation procedure to assess the safety level of humanoid robots will be essential}. Moreover, ethical considerations arise from biases present in human-generated training data, which can lead to unfair or discriminatory robot behaviors across different user demographics. These biases may be amplified in embodied systems that interact directly with humans in diverse settings. \solution{Federated learning approaches could help address some of these ethical concerns by enabling model training across distributed devices while preserving data privacy and potentially reducing centralized data biases \cite{kim2024addressing}.}} 

\revised{\textit{Key Takeaways}: With rapid progress in FMs and humanoid robots, a plethora of research studies on humanoid embodied intelligence is anticipated. The first method outlined in Sec.~\ref{sec:foundation_model_for_humanoid} provides a more accessible way to leverage VLMs and, thus, has been the popular approach taken by researchers so far, but VLMs as is lack a deep understanding of robot actions. 
In the near future, we expect increased research efforts devoted to VLA models described in Sec.~\ref{sec:humanoid_foundation_model} to address this issue. It takes great effort to train RFMs from (i) a large amount of real robot data (\textit{e.g.}, Gemini-Robotics \cite{gemini_robotics_2025, bu2025agibot}) or (ii) a combination of robot data with synthetic data from simulation or video generation models (\textit{e.g.}, GR00T N1 \cite{gr00tn1}). With the development of RFMs, we expect to have FMs that can well understand robot action and the physical world, enabling rapid deployment of robotic controllers achieved with little to no additional robot data. 
For further reading of FMs for robotics, we recommend the survey in~\cite{firoozi2023foundation, Robots_via_Foundation_Models}. Since most of the FMs are built with the Transformer backbones, please refer to~\cite{Attention_Need} for a comprehensive mathematical foundation.}

\section{Additional Discussions}
\label{sec:discussion}


\subsection{Model-based Methods Versus Learning-based Methods}

\revised{Based on the studies in this survey, we provide a clear comparison between model-based and learning-based methods in Table.~\ref{tab:model_vs_learning}. 
In this table, we compare the performance in several aspects: the ability to incorporate multi-modal sensor input, the control robustness, the motion accuracy, the real-time feasibility, the versatility, and the generalizability of each method. 
Additionally, we evaluate the state-of-the-art capability in several humanoid skills, including locomotion, loco-manipulation, and whole-body multi-contact control. We list representative papers that achieved these skills as supporting evidence for this evaluation. Note that many algorithm performances reported are anecdotal and highly depend on their implementation. Therefore, obtaining a concrete numerical comparison is challenging. This is mentioned in Sec.~\ref{sec:benchmark} challenges in lack of appropriate benchmark.}

\revised{A key takeaway from this comparison is that learning-based methods exhibit superior robustness via RL and versatility via IL, while model-based approaches offer advantages in motion accuracy. The emerging paradigm of VLA holds promise for enhanced generalizability; however, its effectiveness in mastering humanoid skills has yet to be convincingly demonstrated.}

\begin{table}[]
\caption{Comparison between model-based and learning-based methods}
\begin{tabular}{p{25mm} cccc}
\hline
\hline
\multicolumn{1}{l|}{} & \multicolumn{1}{c|}{Model-based} & \multicolumn{3}{c}{Learning-based} \\
\multicolumn{1}{l|}{} & \multicolumn{1}{c|}{MPC, WBC} & RL & IL & VLA \\ \hline\hline
Performance measure &  &  &  &  \\ \hline
\multicolumn{1}{l|}{\begin{tabular}[c]{@{}l@{}}Flexibility to incorporate\\multi-modal sensor \end{tabular}} & \multicolumn{1}{c|}{low} & medium & high & very high \\
\multicolumn{1}{l|}{Robustness} & \multicolumn{1}{c|}{high} & very high & low & very low \\
\multicolumn{1}{l|}{Motion accuracy} & \multicolumn{1}{c|}{very high} & medium & medium & very low \\
\multicolumn{1}{l|}{Real-time feasibility} & \multicolumn{1}{c|}{high} & high & high & low \\
\multicolumn{1}{l|}{Versatility} & \multicolumn{1}{c|}{high} & medium & high & high \\
\multicolumn{1}{l|}{Generalizability} & \multicolumn{1}{c|}{medium} & low & low & high \\ \hline\hline
\multicolumn{5}{l}{Achieved performance of humanoid skills} \\ \hline
\multicolumn{1}{l|}{Locomotion} & \multicolumn{1}{c|}{\begin{tabular}[c]{@{}c@{}}very high\\ \cite{khazoom2024tailoring, wensing_thesis}\end{tabular}} & \begin{tabular}[c]{@{}c@{}}very high\\\cite{humanoid_parkour_learning, Cassie_dash}\end{tabular} & \begin{tabular}[c]{@{}c@{}}medium\\ \cite{huang2024diffuseloco}\end{tabular} & NA \\
\multicolumn{1}{l|}{\begin{tabular}[c]{@{}l@{}}Loco-manipulation\\ (hand and foot contact)\end{tabular}} & \multicolumn{1}{c|}{\begin{tabular}[c]{@{}c@{}}high\\ \cite{li2023dynamic, CarpentierTRO}\end{tabular}} & \begin{tabular}[c]{@{}c@{}}medium\\ \cite{dao2022sim, zhang2024wococo}\end{tabular} & \begin{tabular}[c]{@{}c@{}}low\\ 
 \cite{humanplus}\end{tabular} & NA \\
\multicolumn{1}{l|}{\begin{tabular}[c]{@{}c@{}} Whole-body \\multi-contact\end{tabular}} & \multicolumn{1}{c|}{\begin{tabular}[c]{@{}c@{}}medium\\ \cite{lengagne2013generation}\end{tabular}} & \begin{tabular}[c]{@{}c@{}}high\\ \cite{zhuang2025embracecollisionshumanoidshadowing}\end{tabular} & NA & NA \\ \hline\hline
\end{tabular}
\label{tab:model_vs_learning}
\vspace{-0.4cm}
\end{table}

\subsection{Hardware Limitations for Humanoid Robots}


\revised{Advancements in control, planning, and learning for loco-manipulation necessitate robot hardware that has high-power-density actuators, strong yet lightweight limbs, and accurate sensing. These requirements are often limitations to hardware performance, which drives hardware designers to push the boundary of humanoid physical capabilities.}

\revised{The traditional approach to increase torque capacity is through high gear reduction \cite{ficht2021bipedal} and the inclusion of elastic elements to be impact-resilient \cite{kim_dynamic_2020}. However, \solution{the last decade has witnessed advancements in electric motor design such as Quasi-Direct Drive (QDD) actuators (less than 10:1 gear ratio) achieving dynamic and accurate motions \cite{liao2024berkeleyhumanoid, li2023dynamic}.} Among the many advantages QDD actuators offer for achieving dynamic motions are their backdrivability and high force control bandwidth, due to reduced actuator backlash and friction \cite{chignoli2021humanoid}. However, achieving high torque requires a high current, which in turn causes strain on power electronics and overheating. \solution{Solving this problem involves manufacturing with materials to enhance heat dissipation \cite{ficht2021bipedal} and continuing to improve controller performance for energy efficiency~\cite{Locomotion_efficiency_survey_2018}}}.

\revised{Another key element of hardware design is the robot’s inertial properties. In building robots capable of gymnastic-style maneuvers, many designs place mass away from end-effectors \cite{chignoli2021humanoid, li2023dynamic, xia2024duke}. The MIT Humanoid, for example, contains around $75$\% of its mass in the torso \cite{chignoli2021humanoid}. This is often achieved through belt drives or parallel mechanisms to remotely actuate joints such as the ankle or knee \cite{li2023dynamic, xia2024duke}. While this significantly reduces rotational inertia and the torque demand for agile maneuvers, there are a number of notable drawbacks. Four-bar transmissions often reduce joint range of motion, while belt drives may require additional modeling and maintenance \cite{ficht2021bipedal}. \solution{Specialized mechanisms for remote actuation that allow a full range of motion can potentially address this problem \cite{park2023design}.}}

\revised{Battery technology remains one of the primary hardware limitations in the development of sustained, autonomous humanoid robots. Limited battery capacity restricts both operational time and the feasibility of executing high-power, full-body motions in untethered scenarios. \solution{One promising direction is the development of more compact and integrated battery designs. For example, embedding battery cells directly into the structural components of the robot chassis can increase effective energy capacity without compromising weight distribution or form factor. In parallel, solid-state batteries have become commercially available with enhanced energy density and safety profiles \cite{machin2024advancements}. This has increased the operation times of humanoid robots, such as Apptronik's Apollo (4hrs) \cite{Apptronik} and Unitree's G1 (2hrs) \cite{H1}.} Nevertheless, the continued reliance on tethered power in laboratory settings \cite{Nadia_door_auto, humanoid_shuffle, xia2024duke, DS_iCub} indicates that battery technology remains a limitation.}  

\subsection{The Future Applications and Challenges 
of Humanoid Robots in the Society}

\revised{Humanoid robots with loco-manipulation capabilities will become an essential part of our lives and a pillar of the future economy, similar to how smartphones and autonomous driving have transformed our lives. }

\revised{Given a sufficient amount of time, humanoid robots will be increasingly autonomous and capable of performing general human-level tasks across a wide range of domains. The components discussed in this paper---sensing, planning, control, and learning---coordinate in real time to execute complex, goal-directed behaviors. This integration enables humanoid robots to take on tedious, hazardous, and physically demanding tasks.
\begin{itemize}
\item In outdoor environments, they could serve as first responders during national crises, providing medical care and emergency support; monitor controlled settings in civil and mechanical infrastructure; deliver and sort building materials at construction sites; explore subterranean or collapsed structures with harsh and unpredictable terrain; and assist in agricultural tasks such as planting crops. 
\item In factories and warehouses, humanoid robots have the potential to address the growing demand for flexible manufacturing by performing tasks such as delivery, assembly, and inspection, helping to offset the impact of a declining industrial workforce.
\item In human-centered settings such as households and hospitals, they have the potential to carry laundry, load dishwashers, and move patients between beds, supporting the needs of an aging population and improving their quality of life.
\end{itemize}}

\revised{However, significant challenges remain in integrating humanoid robots into all aspects of society. To be fully accepted, these robots must be physically and mentally reliable, highlighting the core technical challenges of achieving robust and safe mobility and manipulation. Their intentions must also be well predicted and communicated with humans, which calls for advancements in effective human-robot interaction. Beyond the technical hurdles, economic and ethical concerns also pose major obstacles. Humanoid robots are still far from being commercially viable due to high costs, and the absence of legal and regulatory policies further hinders their appearance in our daily lives.}

\section{Conclusion}
Humanoid robotics is advancing at an unprecedented pace, as seen in recent groundbreaking innovations from both industry and academia. \revised{In this survey, we go over foundational and state-of-the-art methods for humanoid loco-manipulation, from both model-based and learning-based perspectives. Although humanoid robots still face significant technical challenges, their agility, safety, reliability, and versatility have been improved significantly. More opportunities are emerging through the exploration of new paradigms.} Physically, advanced observers (vision and whole-body tactile sensing) and estimators are emerging for contact-rich whole-body loco-manipulation. Cognitively, foundation models grounded on humanoid robots have great potential to unlock the ability of open-world understanding and the development of generalized intelligent agents. More importantly, an integration of foundation models with planning and control policies presents promising capabilities. In the foreseeable future, the cost of humanoid robots will continue to decrease, making them more accessible; their physical capability (hardware intelligence) and cognitive intelligence will significantly advance. We look forward to humanoid robots that are responsive, purposeful, and fully capable of human-like loco-manipulation tasks in the upcoming decade. 
\bibliographystyle{IEEEtranN}
{\footnotesize \bibliography{zhaoyuan_refs}}

\begin{thebibliography}{424}
\providecommand{\natexlab}[1]{#1}
\providecommand{\url}[1]{#1}
\csname url@samestyle\endcsname
\providecommand{\newblock}{\relax}
\providecommand{\bibinfo}[2]{#2}
\providecommand{\BIBentrySTDinterwordspacing}{\spaceskip=0pt\relax}
\providecommand{\BIBentryALTinterwordstretchfactor}{4}
\providecommand{\BIBentryALTinterwordspacing}{\spaceskip=\fontdimen2\font plus
\BIBentryALTinterwordstretchfactor\fontdimen3\font minus \fontdimen4\font\relax}
\providecommand{\BIBforeignlanguage}[2]{{%
\expandafter\ifx\csname l@#1\endcsname\relax
\typeout{** WARNING: IEEEtranN.bst: No hyphenation pattern has been}%
\typeout{** loaded for the language `#1'. Using the pattern for}%
\typeout{** the default language instead.}%
\else
\language=\csname l@#1\endcsname
\fi
#2}}
\providecommand{\BIBdecl}{\relax}
\BIBdecl

\bibitem[Samadi et~al.(2021)Samadi, Roux, Tanguy, Caron, and Kheddar]{humanoid_shuffle}
S.~Samadi, J.~Roux, A.~Tanguy, S.~Caron, and A.~Kheddar, ``Humanoid control under interchangeable fixed and sliding unilateral contacts,'' \emph{IEEE Robotics and Automation Letters}, vol.~6, no.~2, pp. 4032--4039, 2021.

\bibitem[Li et~al.(2023{\natexlab{a}})Li, Ma, Kolt, Shah, and Nguyen]{li2023dynamic}
J.~Li, J.~Ma, O.~Kolt, M.~Shah, and Q.~Nguyen, ``Dynamic loco-manipulation on hector: Humanoid for enhanced control and open-source research,'' \emph{arXiv preprint arXiv:2312.11868}, 2023.

\bibitem[Henze et~al.(2016)Henze, Roa, and Ott]{henze2016passivity}
B.~Henze, M.~A. Roa, and C.~Ott, ``Passivity-based whole-body balancing for torque-controlled humanoid robots in multi-contact scenarios,'' \emph{The International Journal of Robotics Research}, vol.~35, no.~12, pp. 1522--1543, 2016.

\bibitem[Figueroa et~al.(2020)Figueroa, Faraji, Koptev, and Billard]{DS_iCub}
N.~Figueroa, S.~Faraji, M.~Koptev, and A.~Billard, ``A dynamical system approach for adaptive grasping, navigation and co-manipulation with humanoid robots,'' in \emph{IEEE International Conference on Robotics and Automation}, 2020, pp. 7676--7682.

\bibitem[Calvert et~al.(2024)Calvert, Penco, Anderson, Bialek, Chatterjee, Mishra, Clark, Bertrand, and Griffin]{Nadia_door_auto}
D.~Calvert, L.~Penco, D.~Anderson, T.~Bialek, A.~Chatterjee, B.~Mishra, G.~Clark, S.~Bertrand, and R.~Griffin, ``A behavior architecture for fast humanoid robot door traversals,'' \emph{arXiv preprint arXiv:2411.03532}, 2024.

\bibitem[Khazoom et~al.(2024)Khazoom, Hong, Chignoli, Stanger-Jones, and Kim]{khazoom2024tailoring}
C.~Khazoom, S.~Hong, M.~Chignoli, E.~Stanger-Jones, and S.~Kim, ``Tailoring solution accuracy for fast whole-body model predictive control of legged robots,'' \emph{IEEE Robotics and Automation Letters}, vol.~9, no.~12, pp. 11\,074--11\,081, 2024.

\bibitem[Goswami and Vadakkepat(2018)]{humanoid_reference}
A.~Goswami and P.~Vadakkepat, \emph{Humanoid Robotics: A Reference}, 1st~ed.\hskip 1em plus 0.5em minus 0.4em\relax Springer Publishing Company, Incorporated, 2018.

\bibitem[Zhao et~al.(2024)Zhao, Yu, Han, Chen, Qiu, and Huang]{BHR-WI_Compliant}
L.~Zhao, Z.~Yu, L.~Han, X.~Chen, X.~Qiu, and Q.~Huang, ``Compliant motion control of wheel-legged humanoid robot on rough terrains,'' \emph{IEEE/ASME Transactions on Mechatronics}, vol.~29, no.~3, pp. 1949--1959, 2024.

\bibitem[Wensing et~al.(2024)Wensing, Posa, Hu, Escande, Mansard, and Prete]{Wensing_survey_TRO}
P.~M. Wensing, M.~Posa, Y.~Hu, A.~Escande, N.~Mansard, and A.~D. Prete, ``Optimization-based control for dynamic legged robots,'' \emph{IEEE Transactions on Robotics}, vol.~40, pp. 43--63, 2024.

\bibitem[Chi et~al.(2024)Chi, Xu, Feng, Cousineau, Du, Burchfiel, Tedrake, and Song]{diffusion_policy}
C.~Chi, Z.~Xu, S.~Feng, E.~Cousineau, Y.~Du, B.~Burchfiel, R.~Tedrake, and S.~Song, ``Diffusion policy: Visuomotor policy learning via action diffusion,'' \emph{The International Journal of Robotics Research}, vol.~0, no.~0, p. 02783649241273668, 2024.

\bibitem[Firoozi et~al.(2024)Firoozi, Tucker, Tian, Majumdar, Sun, Liu, Zhu, Song, Kapoor, Hausman, Ichter, Driess, Wu, Lu, and Schwager]{firoozi2023foundation}
R.~Firoozi \emph{et~al.}, ``Foundation models in robotics: Applications, challenges, and the future,'' \emph{The International Journal of Robotics Research}, vol.~0, no.~0, p. 02783649241281508, 2024.

\bibitem[Hu et~al.(2023)Hu, Xie, Jain, Francis, Patrikar, Keetha, Kim, Xie, Zhang, Zhao, Chong, Wang, Sycara, Johnson-Roberson, Batra, Wang, Scherer, Kira, Xia, and Bisk]{Robots_via_Foundation_Models}
Y.~Hu \emph{et~al.}, ``Toward general-purpose robots via foundation models: A survey and meta-analysis,'' \emph{CoRR}, vol. abs/2312.08782, 2023.

\bibitem[Dahiya et~al.(2009)Dahiya, Metta, Valle, and Sandini]{dahiya2009tactile}
R.~S. Dahiya, G.~Metta, M.~Valle, and G.~Sandini, ``Tactile sensing—from humans to humanoids,'' \emph{IEEE Transactions on Robotics}, vol.~26, no.~1, pp. 1--20, 2009.

\bibitem[Sotaro~Katayama and Tazaki(2023)]{MPC_survey}
M.~M. Sotaro~Katayama and Y.~Tazaki, ``Model predictive control of legged and humanoid robots: models and algorithms,'' \emph{Advanced Robotics}, vol.~37, no.~5, pp. 298--315, 2023.

\bibitem[Bao et~al.(2024)Bao, Humphreys, Peng, and Zhou]{DRL_Legged_Survey}
L.~Bao, J.~Humphreys, T.~Peng, and C.~Zhou, ``Deep reinforcement learning for bipedal locomotion: A brief survey,'' 2024.

\bibitem[Ha et~al.(0)Ha, Lee, van~de Panne, Xie, Yu, and Khadiv]{ha2024learningbased}
S.~Ha, J.~Lee, M.~van~de Panne, Z.~Xie, W.~Yu, and M.~Khadiv, ``Learning-based legged locomotion: State of the art and future perspectives,'' \emph{The International Journal of Robotics Research}, vol.~0, no.~0, p. 02783649241312698, 0.

\bibitem[Gams et~al.(2022)Gams, Petri{\v{c}}, Nemec, and Ude]{Gams2022}
A.~Gams, T.~Petri{\v{c}}, B.~Nemec, and A.~Ude, ``Manipulation learning on humanoid robots,'' \emph{Current Robotics Reports}, vol.~3, no.~3, pp. 97--109, 2022.

\bibitem[Parra-Moreno et~al.(2023)Parra-Moreno, Yanguas-Rojas, and Mojica-Nava]{ParraMoreno2023ImitationMF}
J.~Parra-Moreno, D.~Yanguas-Rojas, and E.~Mojica-Nava, ``Imitation methods for bipedal locomotion of humanoid robots: A survey,'' \emph{IEEE Colombian Conference on Automatic Control}, pp. 1--6, 2023.

\bibitem[Tong et~al.(2024)Tong, Liu, and Zhang]{humanoid_survey_IJAS}
Y.~Tong, H.~Liu, and Z.~Zhang, ``Advancements in humanoid robots: A comprehensive review and future prospects,'' \emph{IEEE/CAA Journal of Automatica Sinica}, vol.~11, no.~2, pp. 301--328, 2024.

\bibitem[Wieber et~al.(2016)Wieber, Tedrake, and Kuindersma]{wieber2016shr}
P.-B. Wieber, R.~Tedrake, and S.~Kuindersma, ``Modeling and control of legged robots,'' in \emph{handbook of robotics}.\hskip 1em plus 0.5em minus 0.4em\relax Springer, 2016, pp. 1203--1234.

\bibitem[Carpentier and Wieber(2021)]{carpentier2021crr}
J.~Carpentier and P.-B. Wieber, ``Recent progress in legged robots locomotion control,'' \emph{Current Robotics Reports}, vol.~2, no.~3, pp. 231--238, 2021.

\bibitem[Khan and Mandava(2023)]{khan2023robotica}
M.~S. Khan and R.~K. Mandava, ``A review on gait generation of the biped robot on various terrains,'' \emph{Robotica}, vol.~41, no.~6, pp. 1888--1930, 2023.

\bibitem[Westervelt et~al.(2018)Westervelt, Grizzle, Chevallereau, Choi, and Morris]{Westervelt2018}
E.~R. Westervelt, J.~W. Grizzle, C.~Chevallereau, J.~H. Choi, and B.~Morris, \emph{Feedback Control of Dynamic Bipedal Robot Locomotion}.\hskip 1em plus 0.5em minus 0.4em\relax CRC Press, 2018.

\bibitem[Kajita et~al.(2014)Kajita, Hirukawa, Harada, and Yokoi]{kajita2014book}
S.~Kajita, H.~Hirukawa, K.~Harada, and K.~Yokoi, \emph{Introduction to Humanoid Robotics}.\hskip 1em plus 0.5em minus 0.4em\relax Springer Berlin, Heidelberg, 2014.

\bibitem[McGeer(1990)]{passive_walking_McGeer}
T.~McGeer, ``Passive walking with knees,'' in \emph{Proceedings., IEEE International Conference on Robotics and Automation}, 1990, pp. 1640--1645 vol.3.

\bibitem[Collins et~al.(2005)Collins, Ruina, Tedrake, and Wisse]{Passive-Walkers}
S.~Collins, A.~Ruina, R.~Tedrake, and M.~Wisse, ``Efficient bipedal robots based on passive-dynamic walkers,'' \emph{Science}, vol. 307, no. 5712, pp. 1082--1085, 2005.

\bibitem[Kajita et~al.(2003)Kajita, Kanehiro, Kaneko, Fujiwara, Harada, Yokoi, and Hirukawa]{Kajita2003}
S.~Kajita, F.~Kanehiro, K.~Kaneko, K.~Fujiwara, K.~Harada, K.~Yokoi, and H.~Hirukawa, ``Biped walking pattern generation by using preview control of zero-moment point,'' in \emph{IEEE International Conference on Robotics and Automation}, vol.~2, 2003, pp. 1620--1626.

\bibitem[Kim et~al.(2020)Kim, Jorgensen, Lee, Ahn, Luo, and Sentis]{kim_dynamic_2020}
D.~Kim, S.~J. Jorgensen, J.~Lee, J.~Ahn, J.~Luo, and L.~Sentis, ``Dynamic locomotion for passive-ankle biped robots and humanoids using whole-body locomotion control,'' \emph{The International Journal of Robotics Research}, vol.~39, no.~8, pp. 936--956, 2020.

\bibitem[Li et~al.(2024{\natexlab{a}})Li, Kolt, and Nguyen]{li2024continuous}
J.~Li, O.~Kolt, and Q.~Nguyen, ``Continuous dynamic bipedal jumping via adaptive-model optimization,'' \emph{arXiv preprint arXiv:2404.11807}, 2024.

\bibitem[Wensing(2014)]{wensing_thesis}
P.~Wensing, ``\BIBforeignlanguage{English}{Optimization and control of dynamic humanoid running and jumping},'' Ph.D. dissertation, The Ohio State University, 2014.

\bibitem[Pardis et~al.(2024)Pardis, Chignoli, and Kim]{pardis2024probabilistic}
S.~Pardis, M.~Chignoli, and S.~Kim, ``Probabilistic homotopy optimization for dynamic motion planning,'' in \emph{International Conference on Intelligent Robots and Systems}, 2024, pp. 12\,856--12\,863.

\bibitem[Englsberger et~al.(2020)Englsberger, Dietrich, Mesesan, Garofalo, Ott, and Albu-Sch{\"a}ffer]{englsberger2020mptc}
J.~Englsberger, A.~Dietrich, G.-A. Mesesan, G.~Garofalo, C.~Ott, and A.~O. Albu-Sch{\"a}ffer, ``Mptc-modular passive tracking controller for stack of tasks based control frameworks,'' \emph{Robotics: Science and Systems}, 2020.

\bibitem[Agravante et~al.(2019)Agravante, Cherubini, Sherikov, Wieber, and Kheddar]{agravante2019human}
D.~J. Agravante, A.~Cherubini, A.~Sherikov, P.-B. Wieber, and A.~Kheddar, ``Human-humanoid collaborative carrying,'' \emph{IEEE Transactions on Robotics}, vol.~35, no.~4, pp. 833--846, 2019.

\bibitem[Morimoto and Atkeson(2007)]{Atkeson_Biped_Learn_2007}
J.~Morimoto and C.~G. Atkeson, ``Learning biped locomotion,'' \emph{IEEE Robotics and Automation Magazine}, vol.~14, no.~2, pp. 41--51, 2007.

\bibitem[Nakanishi et~al.(2004)Nakanishi, Morimoto, Endo, Cheng, Schaal, and Kawato]{Nakanishi2004_LearnBiped}
J.~Nakanishi, J.~Morimoto, G.~Endo, G.~Cheng, S.~Schaal, and M.~Kawato, ``Learning from demonstration and adaptation of biped locomotion,'' \emph{Robotics and Autonomous Systems}, vol.~47, no.~2, pp. 79--91, 2004.

\bibitem[Calandra et~al.(2015)Calandra, Seyfarth, Peters, and Deisenroth]{Calandra2015_BO_Biped}
R.~Calandra, A.~Seyfarth, J.~Peters, and M.~P. Deisenroth, ``Bayesian optimization for learning gaits under uncertainty: An experimental comparison on a dynamic bipedal walker,'' \emph{Annals of Mathematics and Artificial Intelligence}, vol.~76, no. 1–2, p. 5–23, 2015.

\bibitem[Crowley et~al.(2023)Crowley, Dao, Duan, Green, Hurst, and Fern]{Cassie_dash}
D.~Crowley, J.~Dao, H.~Duan, K.~Green, J.~Hurst, and A.~Fern, ``Optimizing bipedal locomotion for the 100m dash with comparison to human running,'' in \emph{IEEE International Conference on Robotics and Automation}, 2023, pp. 12\,205--12\,211.

\bibitem[Li et~al.(2024{\natexlab{b}})Li, Peng, Abbeel, Levine, Berseth, and Sreenath]{li2023robust}
Z.~Li, X.~B. Peng, P.~Abbeel, S.~Levine, G.~Berseth, and K.~Sreenath, ``Reinforcement learning for versatile, dynamic, and robust bipedal locomotion control,'' \emph{The International Journal of Robotics Research}, vol.~0, no.~0, p. 02783649241285161, 2024.

\bibitem[Siekmann et~al.(2022)Siekmann, Green, Warila, Fern, and Hurst]{siekmann2023blind}
J.~Siekmann, K.~Green, J.~Warila, A.~Fern, and J.~Hurst, ``Blind bipedal stair traversal via sim-to-real reinforcement learning,'' in \emph{Robotics: Science and Systems}, 2022.

\bibitem[Zhuang et~al.(2024)Zhuang, Yao, and Zhao]{humanoid_parkour_learning}
Z.~Zhuang, S.~Yao, and H.~Zhao, ``Humanoid parkour learning,'' in \emph{Annual Conference on Robot Learning}, 2024.

\bibitem[Dao et~al.(2022)Dao, Green, Duan, Fern, and Hurst]{dao2022sim}
J.~Dao, K.~Green, H.~Duan, A.~Fern, and J.~Hurst, ``Sim-to-real learning for bipedal locomotion under unsensed dynamic loads,'' in \emph{International Conference on Robotics and Automation}, 2022, pp. 10\,449--10\,455.

\bibitem[Radosavovic et~al.(2024)Radosavovic, Xiao, Zhang, Darrell, Malik, and Sreenath]{transformer_digit_RL}
I.~Radosavovic, T.~Xiao, B.~Zhang, T.~Darrell, J.~Malik, and K.~Sreenath, ``Real-world humanoid locomotion with reinforcement learning,'' \emph{Science Robotics}, vol.~9, no.~89, 2024.

\bibitem[Kashiri et~al.(2018)Kashiri, Abate, Abram, Albu-Schaffer, Clary, Daley, Faraji, Furnemont, Garabini, Geyer, Grabowski, Hurst, Malzahn, Mathijssen, Remy, Roozing, Shahbazi, Simha, Song, Smit-Anseeuw, Stramigioli, Vanderborght, Yesilevskiy, and Tsagarakis]{Locomotion_efficiency_survey_2018}
N.~Kashiri \emph{et~al.}, ``An overview on principles for energy efficient robot locomotion,'' \emph{Frontiers in Robotics and AI}, vol.~5, 2018.

\bibitem[Reher et~al.(2016)Reher, Cousineau, Hereid, Hubicki, and Ames]{DURUS}
J.~Reher, E.~A. Cousineau, A.~Hereid, C.~M. Hubicki, and A.~D. Ames, ``Realizing dynamic and efficient bipedal locomotion on the humanoid robot durus,'' in \emph{IEEE International Conference on Robotics and Automation}, 2016, pp. 1794--1801.

\bibitem[Heremans et~al.(2019)Heremans, Vijayakumar, Bouri, Dehez, and Ronsse]{lock_ankle}
F.~Heremans, S.~Vijayakumar, M.~Bouri, B.~Dehez, and R.~Ronsse, ``\BIBforeignlanguage{en}{Bio-inspired design and validation of the efficient lockable spring ankle ({ELSA}) prosthesis},'' \emph{\BIBforeignlanguage{en}{IEEE Int Conf Rehabil Robot}}, vol. 2019, pp. 411--416, Jun. 2019.

\bibitem[Xia et~al.(2024)Xia, Li, Lee, Scutari, and Chen]{xia2024duke}
B.~Xia, B.~Li, J.~Lee, M.~Scutari, and B.~Chen, ``The duke humanoid: Design and control for energy efficient bipedal locomotion using passive dynamics,'' \emph{arXiv preprint arXiv:2409.19795}, 2024.

\bibitem[Yang et~al.(2020)Yang, Yuan, Heng, Komura, and Li]{RAL_MOE_Humanoid}
C.~Yang, K.~Yuan, S.~Heng, T.~Komura, and Z.~Li, ``Learning natural locomotion behaviors for humanoid robots using human bias,'' \emph{IEEE Robotics and Automation Letters}, vol.~5, no.~2, pp. 2610--2617, 2020.

\bibitem[Huang and Grizzle(2023)]{huang_cassie_navigation}
J.-K. Huang and J.~W. Grizzle, ``Efficient anytime clf reactive planning system for a bipedal robot on undulating terrain,'' \emph{IEEE Transactions on Robotics}, vol.~39, no.~3, pp. 2093--2110, 2023.

\bibitem[Muenprasitivej et~al.(2024)Muenprasitivej, Jiang, Shamsah, Coogan, and Zhao]{muenprasitivej2024bipedal}
K.~Muenprasitivej, J.~Jiang, A.~Shamsah, S.~Coogan, and Y.~Zhao, ``Bipedal safe navigation over uncertain rough terrain: Unifying terrain mapping and locomotion stability,'' \emph{IEEE/RSJ International Conference on Intelligent Robots and Systems}, 2024.

\bibitem[McCrory et~al.(2023)McCrory, Mishra, Griffin, Pratt, and Sevil]{mccrory2023bipedal}
S.~McCrory, B.~Mishra, R.~Griffin, J.~Pratt, and H.~E. Sevil, ``Bipedal navigation planning over rough terrain using traversability models,'' in \emph{SoutheastCon}, 2023, pp. 89--95.

\bibitem[Li et~al.(2023{\natexlab{b}})Li, Zeng, Chen, and Sreenath]{li2023autonomous}
Z.~Li, J.~Zeng, S.~Chen, and K.~Sreenath, ``Autonomous navigation of underactuated bipedal robots in height-constrained environments,'' \emph{The International Journal of Robotics Research}, vol.~42, no.~8, pp. 565--585, 2023.

\bibitem[Cheng et~al.(2024{\natexlab{a}})Cheng, Ji, Yang, Zou, Kautz, Bıyık, Yin, Liu, and Wang]{cheng2024navila}
A.-C. Cheng, Y.~Ji, Z.~Yang, X.~Zou, J.~Kautz, E.~Bıyık, H.~Yin, S.~Liu, and X.~Wang, ``Navila: Legged robot vision-language-action model for navigation,'' \emph{arXiv preprint arXiv:2412.04453}, 2024.

\bibitem[Acosta and Posa(2023)]{Acosta_Cassie}
B.~Acosta and M.~Posa, ``Bipedal walking on constrained footholds with mpc footstep control,'' in \emph{IEEE-RAS International Conference on Humanoid Robots}, 2023, pp. 1--8.

\bibitem[Duan et~al.(2024)Duan, Pandit, Gadde, Van~Marum, Dao, Kim, and Fern]{duan2023learning}
H.~Duan, B.~Pandit, M.~S. Gadde, B.~Van~Marum, J.~Dao, C.~Kim, and A.~Fern, ``Learning vision-based bipedal locomotion for challenging terrain,'' in \emph{IEEE International Conference on Robotics and Automation}, 2024, pp. 56--62.

\bibitem[Griffin et~al.(2019)Griffin, Wiedebach, McCrory, Bertrand, Lee, and Pratt]{griffin2019astar}
R.~J. Griffin, G.~Wiedebach, S.~McCrory, S.~Bertrand, I.~Lee, and J.~Pratt, ``Footstep planning for autonomous walking over rough terrain,'' in \emph{IEEE-RAS International Conference on Humanoid Robots}, 2019, pp. 9--16.

\bibitem[Peng et~al.(2023)Peng, Donca, Castillo, and Hereid]{peng2023safe}
C.~Peng, O.~Donca, G.~Castillo, and A.~Hereid, ``Safe bipedal path planning via control barrier functions for polynomial shape obstacles estimated using logistic regression,'' in \emph{IEEE International Conference on Robotics and Automation}, 2023, pp. 3649--3655.

\bibitem[Zhao et~al.(2016)Zhao, Topcu, and Sentis]{Zhao_CDC2016}
Y.~Zhao, U.~Topcu, and L.~Sentis, ``High-level planner synthesis for whole-body locomotion in unstructured environments,'' in \emph{IEEE 55th Conference on Decision and Control}, 2016, pp. 6557--6564.

\bibitem[Shamsah et~al.(2023)Shamsah, Gu, Warnke, Hutchinson, and Zhao]{shamsah2023TRO}
A.~Shamsah, Z.~Gu, J.~Warnke, S.~Hutchinson, and Y.~Zhao, ``Integrated task and motion planning for safe legged navigation in partially observable environments,'' \emph{IEEE Transactions on Robotics}, vol.~39, no.~6, pp. 4913--4934, 2023.

\bibitem[Shamsah et~al.(2025)Shamsah, Agarwal, Katta, Raju, Kousik, and Zhao]{shamsah2024socially}
A.~Shamsah, K.~Agarwal, N.~Katta, A.~Raju, S.~Kousik, and Y.~Zhao, ``Socially acceptable bipedal robot navigation via social zonotope network model predictive control,'' \emph{IEEE Transactions on Automation Science and Engineering}, 2025.

\bibitem[Zhu et~al.(2025)Zhu, Raju, Shamsah, Wu, Hutchinson, and Zhao]{zhu2025emobipednav}
W.~Zhu, A.~Raju, A.~Shamsah, A.~Wu, S.~Hutchinson, and Y.~Zhao, ``Emobipednav: Emotion-aware social navigation for bipedal robots with deep reinforcement learning,'' \emph{arXiv preprint arXiv:2503.12538}, 2025.

\bibitem[Gibson et~al.(2022)Gibson, Dosunmu-Ogunbi, Gong, and Grizzle]{gibson2022terrain}
G.~Gibson, O.~Dosunmu-Ogunbi, Y.~Gong, and J.~Grizzle, ``Terrain-adaptive, alip-based bipedal locomotion controller via model predictive control and virtual constraints,'' in \emph{IEEE/RSJ International Conference on Intelligent Robots and Systems}, 2022, pp. 6724--6731.

\bibitem[Xia et~al.(2011)Xia, Xiong, and Chen]{global_step_planner_TMECH}
Z.~Xia, J.~Xiong, and K.~Chen, ``Global navigation for humanoid robots using sampling-based footstep planners,'' \emph{IEEE/ASME Transactions on Mechatronics}, vol.~16, no.~4, pp. 716--723, 2011.

\bibitem[Shamsah et~al.(2024)Shamsah, Jiang, Yoon, Coogan, and Zhao]{shamsah2024terrain_predictivecontrol}
A.~Shamsah, J.~Jiang, Z.~Yoon, S.~Coogan, and Y.~Zhao, ``Terrain-aware model predictive control of heterogeneous bipedal and aerial robot coordination for search and rescue tasks,'' 2024.

\bibitem[Jiang et~al.(2023{\natexlab{a}})Jiang, Coogan, and Zhao]{jiang2023abstraction}
J.~Jiang, S.~Coogan, and Y.~Zhao, ``Abstraction-based planning for uncertainty-aware legged navigation,'' \emph{IEEE Open Journal of Control Systems}, vol.~2, pp. 221--234, 2023.

\bibitem[Ahmadi et~al.(2021)Ahmadi, Xiong, and Ames]{ahmadi2021risk}
M.~Ahmadi, X.~Xiong, and A.~D. Ames, ``Risk-averse control via cvar barrier functions: Application to bipedal robot locomotion,'' \emph{IEEE Control Systems Letters}, vol.~6, pp. 878--883, 2021.

\bibitem[Smith et~al.(2012)Smith, Karayiannidis, Nalpantidis, Gratal, Qi, Dimarogonas, and Kragic]{smith2012dual}
C.~Smith, Y.~Karayiannidis, L.~Nalpantidis, X.~Gratal, P.~Qi, D.~V. Dimarogonas, and D.~Kragic, ``Dual arm manipulation—a survey,'' \emph{Robotics and Autonomous systems}, vol.~60, no.~10, pp. 1340--1353, 2012.

\bibitem[Pollard(1994)]{pollard1994wholehand}
N.~S. Pollard, ``Parallel methods for synthesizing whole-hand grasps from generalized prototypes,'' Ph.D. dissertation, Massachusetts Institute of Technology, Cambridge, MA, 1994.

\bibitem[Salisbury(1987)]{salisbury1987whole}
J.~K. Salisbury, ``Whole-arm manipulation,'' in \emph{Proceeding of the International Symposium of Robotics Research}.\hskip 1em plus 0.5em minus 0.4em\relax The MIT Press, 1987.

\bibitem[Townsend and Salisbury(1993)]{townsend1993mechanical}
W.~T. Townsend and J.~K. Salisbury, ``Mechanical design for whole-arm manipulation,'' in \emph{Robots and Biological Systems: Towards a New Bionics}.\hskip 1em plus 0.5em minus 0.4em\relax Springer, 1993, pp. 153--164.

\bibitem[Cheng et~al.(2023{\natexlab{a}})Cheng, Patil, Temel, Kroemer, and Mason]{cheng2023enhancing}
X.~Cheng, S.~Patil, Z.~Temel, O.~Kroemer, and M.~T. Mason, ``Enhancing dexterity in robotic manipulation via hierarchical contact exploration,'' \emph{IEEE Robotics and Automation Letters}, vol.~9, no.~1, pp. 390--397, 2023.

\bibitem[Goncalves et~al.(2022)Goncalves, Kuppuswamy, Beaulieu, Uttamchandani, Tsui, and Alspach]{Punyo}
A.~Goncalves, N.~Kuppuswamy, A.~Beaulieu, A.~Uttamchandani, K.~M. Tsui, and A.~Alspach, ``Punyo-1: Soft tactile-sensing upper-body robot for large object manipulation and physical human interaction,'' in \emph{IEEE International Conference on Soft Robotics}, 2022, pp. 844--851.

\bibitem[Stasse and Righetti(2020)]{stasse2020whole}
O.~Stasse and L.~Righetti, \emph{Whole-Body Manipulation}.\hskip 1em plus 0.5em minus 0.4em\relax Springer, 2020, pp. 1--9.

\bibitem[Cheng et~al.(2019{\natexlab{a}})Cheng, Dean-Leon, Bergner, Rogelio Guadarrama~Olvera, Leboutet, and Mittendorfer]{cheng2019robotskin}
G.~Cheng, E.~Dean-Leon, F.~Bergner, J.~Rogelio Guadarrama~Olvera, Q.~Leboutet, and P.~Mittendorfer, ``A comprehensive realization of robot skin: Sensors, sensing, control, and applications,'' \emph{Proceedings of the IEEE}, vol. 107, no.~10, pp. 2034--2051, 2019.

\bibitem[Gupta et~al.(2022)Gupta, Mao, Bhatia, Cheng, King, Hou, and Mason]{gupta2022extrinsic}
A.~Gupta, Y.~Mao, A.~Bhatia, X.~Cheng, J.~King, Y.~Hou, and M.~T. Mason, ``Extrinsic dexterous manipulation with a direct-drive hand: A case study,'' in \emph{IEEE/RSJ International Conference on Intelligent Robots and Systems}, 2022, pp. 4660--4667.

\bibitem[Hsiao and Lozano-Perez(2006)]{hsiao2006imitation}
K.~Hsiao and T.~Lozano-Perez, ``Imitation learning of whole-body grasps,'' in \emph{IEEE/RSJ International Conference on Intelligent Robots and Systems}, 2006, pp. 5657--5662.

\bibitem[Yuan et~al.(2019)Yuan, Hang, Song, Kragic, Wang, and Stork]{yuan2019reinforcement}
W.~Yuan, K.~Hang, H.~Song, D.~Kragic, M.~Y. Wang, and J.~A. Stork, ``Reinforcement learning in topology-based representation for human body movement with whole arm manipulation,'' in \emph{IEEE International Conference on Robotics and Automation}, 2019, pp. 2153--2160.

\bibitem[Zhang et~al.(2023)Zhang, Barreiros, and Onol]{zhang2023plan}
M.~Zhang, J.~Barreiros, and A.~O. Onol, ``Plan-guided reinforcement learning for whole-body manipulation,'' in \emph{IROS Workshop on Leveraging Models for Contact-Rich Manipulation}, 2023.

\bibitem[Florek-Jasińska et~al.(2014)Florek-Jasińska, Wimböck, and Ott]{WAM_Justin}
M.~Florek-Jasińska, T.~Wimböck, and C.~Ott, ``Humanoid compliant whole arm dexterous manipulation: Control design and experiments,'' in \emph{IEEE/RSJ International Conference on Intelligent Robots and Systems}, 2014, pp. 1616--1621.

\bibitem[Murooka et~al.(2015)Murooka, Nozawa, Kakiuchi, Okada, and Inaba]{Murooka_heavy_push}
M.~Murooka, S.~Nozawa, Y.~Kakiuchi, K.~Okada, and M.~Inaba, ``Whole-body pushing manipulation with contact posture planning of large and heavy object for humanoid robot,'' in \emph{IEEE International Conference on Robotics and Automation}, 2015, pp. 5682--5689.

\bibitem[Cheng et~al.(2023{\natexlab{b}})Cheng, Kumar, and Pathak]{legAsManip}
X.~Cheng, A.~Kumar, and D.~Pathak, ``Legs as manipulator: Pushing quadrupedal agility beyond locomotion,'' in \emph{IEEE International Conference on Robotics and Automation}, 2023, pp. 5106--5112.

\bibitem[He et~al.(2024{\natexlab{a}})He, Lei, Ze, Sreenath, Li, and Xu]{he2024learningvisualquadrupedallocomanipulation}
Z.~He, K.~Lei, Y.~Ze, K.~Sreenath, Z.~Li, and H.~Xu, ``Learning visual quadrupedal loco-manipulation from demonstrations,'' 2024.

\bibitem[Arm et~al.(2024)Arm, Mittal, Kolvenbach, and Hutter]{2024pedipulate}
P.~Arm, M.~Mittal, H.~Kolvenbach, and M.~Hutter, ``Pedipulate: Enabling manipulation skills using a quadruped robot’s leg,'' in \emph{IEEE International Conference on Robotics and Automation}, 2024, pp. 5717--5723.

\bibitem[Ferrolho et~al.(2023)Ferrolho, Ivan, Merkt, Havoutis, and Vijayakumar]{Ferrolho2023}
H.~Ferrolho, V.~Ivan, W.~Merkt, I.~Havoutis, and S.~Vijayakumar, ``Roloma: robust loco-manipulation for quadruped robots with arms,'' \emph{Autonomous Robots}, vol.~47, no.~8, p. 1463–1481, 2023.

\bibitem[Sleiman et~al.(2023)Sleiman, Farshidian, and Hutter]{HutterScience_locomanip}
J.-P. Sleiman, F.~Farshidian, and M.~Hutter, ``Versatile multicontact planning and control for legged loco-manipulation,'' \emph{Science Robotics}, vol.~8, no.~81, 2023.

\bibitem[Portela et~al.(2024)Portela, Margolis, Ji, and Agrawal]{portela2024learning}
T.~Portela, G.~B. Margolis, Y.~Ji, and P.~Agrawal, ``Learning force control for legged manipulation,'' in \emph{International Conference on Robotics and Automation}, 2024, pp. 15\,366--15\,372.

\bibitem[Liu et~al.(2024{\natexlab{a}})Liu, Chen, Cheng, Ji, Qiu, Yang, and Wang]{liu2024visualwholebodycontrollegged}
M.~Liu, Z.~Chen, X.~Cheng, Y.~Ji, R.~Qiu, R.~Yang, and X.~Wang, ``Visual whole-body control for legged loco-manipulation,'' \emph{The 8th Conference on Robot Learning}, 2024.

\bibitem[Zhang et~al.(2018)Zhang, McCarthy, Jow, Lee, Chen, Goldberg, and Abbeel]{DRACO_teleop}
T.~Zhang, Z.~McCarthy, O.~Jow, D.~Lee, X.~Chen, K.~Goldberg, and P.~Abbeel, ``Deep imitation learning for complex manipulation tasks from virtual reality teleoperation,'' in \emph{IEEE International Conference on Robotics and Automation}, 2018, pp. 5628--5635.

\bibitem[Rakita et~al.(2019)Rakita, Mutlu, Gleicher, and Hiatt]{SR_humanoid_bimanual}
D.~Rakita, B.~Mutlu, M.~Gleicher, and L.~M. Hiatt, ``Shared control–based bimanual robot manipulation,'' \emph{Science Robotics}, vol.~4, no.~30, 2019.

\bibitem[Seo et~al.(2023)Seo, Han, Sim, Bang, Gonzalez, Sentis, and Zhu]{seo2023deep}
M.~Seo, S.~Han, K.~Sim, S.~H. Bang, C.~Gonzalez, L.~Sentis, and Y.~Zhu, ``Deep imitation learning for humanoid loco-manipulation through human teleoperation,'' \emph{IEEE-RAS International Conference on Humanoid Robots}, pp. 1--8, 2023.

\bibitem[Jorgensen et~al.(2020)Jorgensen, Vedantam, Gupta, Cappel, and Sentis]{jorgensen2019finding}
S.~J. Jorgensen, M.~Vedantam, R.~Gupta, H.~Cappel, and L.~Sentis, ``Finding locomanipulation plans quickly in the locomotion constrained manifold,'' in \emph{IEEE International Conference on Robotics and Automation}, 2020, pp. 6611--6617.

\bibitem[Murooka et~al.(2021{\natexlab{a}})Murooka, Chappellet, Tanguy, Benallegue, Kumagai, Morisawa, Kanehiro, and Kheddar]{Murooka_Pattern}
M.~Murooka, K.~Chappellet, A.~Tanguy, M.~Benallegue, I.~Kumagai, M.~Morisawa, F.~Kanehiro, and A.~Kheddar, ``Humanoid loco-manipulations pattern generation and stabilization control,'' \emph{IEEE Robotics and Automation Letters}, vol.~6, no.~3, pp. 5597--5604, 2021.

\bibitem[Vaz et~al.(2023)Vaz, Kosanovic, and Oh]{Vaz2023}
J.~C. Vaz, N.~Kosanovic, and P.~Oh, ``Art: Avatar robotics telepresence—the future of humanoid material handling loco-manipulation,'' \emph{Intelligent Service Robotics}, vol.~17, no.~2, p. 237–250, 2023.

\bibitem[Chappellet et~al.(2023)Chappellet, Murooka, Caron, Kanehiro, and Kheddar]{chappellet2023humanoid}
K.~Chappellet, M.~Murooka, G.~Caron, F.~Kanehiro, and A.~Kheddar, ``Humanoid loco-manipulations using combined fast dense 3d tracking and slam with wide-angle depth-images,'' \emph{IEEE Transactions on Automation Science and Engineering}, 2023.

\bibitem[Vaillant et~al.(2016)Vaillant, Kheddar, Audren, Keith, Brossette, Escande, Bouyarmane, Kaneko, Morisawa, Gergondet, et~al.]{vaillant2016auro}
J.~Vaillant \emph{et~al.}, ``Multi-contact vertical ladder climbing with an hrp-2 humanoid,'' \emph{Autonomous Robots}, vol.~40, no.~3, pp. 561--580, 2016.

\bibitem[Ferrari et~al.(2023)Ferrari, Rossini, Ruscelli, Laurenzi, Oriolo, Tsagarakis, and Hoffman]{ferrari2023multi}
P.~Ferrari, L.~Rossini, F.~Ruscelli, A.~Laurenzi, G.~Oriolo, N.~G. Tsagarakis, and E.~M. Hoffman, ``Multi-contact planning and control for humanoid robots: Design and validation of a complete framework,'' \emph{Robotics and Autonomous Systems}, vol. 166, p. 104448, 2023.

\bibitem[Kheddar and Billard(2011)]{kheddar2011robio}
A.~Kheddar and A.~Billard, ``A tactile matrix for whole-body humanoid haptic sensing and safe interaction,'' in \emph{IEEE International Conference on Robotics and Biomimetics}, 2011, pp. 1433--1438.

\bibitem[Jain et~al.(2013)Jain, Killpack, Edsinger, and Kemp]{jain2013reaching}
A.~Jain, M.~D. Killpack, A.~Edsinger, and C.~C. Kemp, ``Reaching in clutter with whole-arm tactile sensing,'' \emph{The International Journal of Robotics Research}, vol.~32, no.~4, pp. 458--482, 2013.

\bibitem[Gao et~al.(2022)Gao, Dai, and Nathan]{gao2022tactile}
S.~Gao, Y.~Dai, and A.~Nathan, ``Tactile and vision perception for intelligent humanoids,'' \emph{Advanced Intelligent Systems}, vol.~4, no.~2, p. 2100074, 2022.

\bibitem[Kappassov et~al.(2015)Kappassov, Corrales, and Perdereau]{kappassov2015tactile}
Z.~Kappassov, J.-A. Corrales, and V.~Perdereau, ``Tactile sensing in dexterous robot hands,'' \emph{Robotics and Autonomous Systems}, vol.~74, pp. 195--220, 2015.

\bibitem[Al-Handarish et~al.(2020)Al-Handarish, Omisore, Igbe, Han, Li, Du, Zhang, Wang, et~al.]{al2020tactilesurvey}
Y.~Al-Handarish, O.~M. Omisore, T.~Igbe, S.~Han, H.~Li, W.~Du, J.~Zhang, L.~Wang \emph{et~al.}, ``A survey of tactile-sensing systems and their applications in biomedical engineering,'' \emph{Advances in Materials Science and Engineering}, vol. 2020, 2020.

\bibitem[Li et~al.(2013)Li, Sch{\"u}rmann, Haschke, and Ritter]{li2013control}
Q.~Li, C.~Sch{\"u}rmann, R.~Haschke, and H.~J. Ritter, ``A control framework for tactile servoing.'' in \emph{Robotics: Science and systems}, 2013.

\bibitem[Veiga et~al.(2020{\natexlab{a}})Veiga, Edin, and Peters]{veiga2020grip}
F.~Veiga, B.~Edin, and J.~Peters, ``Grip stabilization through independent finger tactile feedback control,'' \emph{Sensors}, vol.~20, no.~6, p. 1748, 2020.

\bibitem[Song et~al.(2013)Song, Liu, Althoefer, Nanayakkara, and Seneviratne]{song2013efficient}
X.~Song, H.~Liu, K.~Althoefer, T.~Nanayakkara, and L.~D. Seneviratne, ``Efficient break-away friction ratio and slip prediction based on haptic surface exploration,'' \emph{IEEE Transactions on Robotics}, vol.~30, no.~1, pp. 203--219, 2013.

\bibitem[P.~Mittendorfer and Cheng(2015)]{WBM_tactile_gordon}
E.~Y. P.~Mittendorfer and G.~Cheng, ``Realizing whole-body tactile interactions with a self-organizing, multi-modal artificial skin on a humanoid robot,'' \emph{Advanced Robotics}, vol.~29, no.~1, pp. 51--67, 2015.

\bibitem[Sommer et~al.(2014)Sommer, Li, and Billard]{sommer2014bimanual}
N.~Sommer, M.~Li, and A.~Billard, ``Bimanual compliant tactile exploration for grasping unknown objects,'' in \emph{IEEE International Conference on Robotics and Automation}, 2014, pp. 6400--6407.

\bibitem[Hogan et~al.(2020)Hogan, Ballester, Dong, and Rodriguez]{hogan2020tactile}
F.~R. Hogan, J.~Ballester, S.~Dong, and A.~Rodriguez, ``Tactile dexterity: Manipulation primitives with tactile feedback,'' in \emph{IEEE International Conference on Robotics and Automation}, 2020, pp. 8863--8869.

\bibitem[Kaboli and Cheng(2018)]{kaboli2018robust}
M.~Kaboli and G.~Cheng, ``Robust tactile descriptors for discriminating objects from textural properties via artificial robotic skin,'' \emph{IEEE Transactions on Robotics}, vol.~34, no.~4, pp. 985--1003, 2018.

\bibitem[Tian et~al.(2019)Tian, Ebert, Jayaraman, Mudigonda, Finn, Calandra, and Levine]{tian2019manipulation}
S.~Tian, F.~Ebert, D.~Jayaraman, M.~Mudigonda, C.~Finn, R.~Calandra, and S.~Levine, ``Manipulation by feel: Touch-based control with deep predictive models,'' in \emph{IEEE International Conference on Robotics and Automation}, 2019, pp. 818--824.

\bibitem[Lambeta et~al.(2020)Lambeta, Chou, Tian, Yang, Maloon, Most, Stroud, Santos, Byagowi, Kammerer, et~al.]{lambeta2020digit}
M.~Lambeta \emph{et~al.}, ``Digit: A novel design for a low-cost compact high-resolution tactile sensor with application to in-hand manipulation,'' \emph{IEEE Robotics and Automation Letters}, vol.~5, no.~3, pp. 3838--3845, 2020.

\bibitem[Van~Hoof et~al.(2015)Van~Hoof, Hermans, Neumann, and Peters]{van2015learning}
H.~Van~Hoof, T.~Hermans, G.~Neumann, and J.~Peters, ``Learning robot in-hand manipulation with tactile features,'' in \emph{IEEE-RAS International Conference on Humanoid Robots}, 2015, pp. 121--127.

\bibitem[Melnik et~al.(2021)Melnik, Lach, Plappert, Korthals, Haschke, and Ritter]{melnik2021using}
A.~Melnik, L.~Lach, M.~Plappert, T.~Korthals, R.~Haschke, and H.~Ritter, ``Using tactile sensing to improve the sample efficiency and performance of deep deterministic policy gradients for simulated in-hand manipulation tasks,'' \emph{Frontiers in Robotics and AI}, vol.~8, p. 538773, 2021.

\bibitem[Lin et~al.(2024{\natexlab{a}})Lin, Zhang, Li, Qi, Yi, Levine, and Malik]{lin2024learning}
T.~Lin, Y.~Zhang, Q.~Li, H.~Qi, B.~Yi, S.~Levine, and J.~Malik, ``Learning visuotactile skills with two multifingered hands,'' \emph{arXiv preprint arXiv:2404.16823}, 2024.

\bibitem[Fu et~al.(2024{\natexlab{a}})Fu, Datta, Huang, Panitch, Drake, Ortiz, Mukadam, Lambeta, Calandra, and Goldberg]{fu2024touch}
L.~Fu, G.~Datta, H.~Huang, W.~C.-H. Panitch, J.~Drake, J.~Ortiz, M.~Mukadam, M.~Lambeta, R.~Calandra, and K.~Goldberg, ``A touch, vision, and language dataset for multimodal alignment,'' in \emph{Forty-first International Conference on Machine Learning}, 2024.

\bibitem[Yu et~al.(2024{\natexlab{a}})Yu, Kelvin, Xiao, Duan, and Soh]{Yu-RSS-24}
S.~Yu, L.~Kelvin, A.~Xiao, J.~Duan, and H.~Soh, ``{Octopi: Object Property Reasoning with Large Tactile-Language Models},'' in \emph{Proceedings of Robotics: Science and Systems}, 2024.

\bibitem[Chebotar et~al.(2014)Chebotar, Kroemer, and Peters]{chebotar2014learning}
Y.~Chebotar, O.~Kroemer, and J.~Peters, ``Learning robot tactile sensing for object manipulation,'' in \emph{2014 IEEE/RSJ International Conference on Intelligent Robots and Systems}, 2014, pp. 3368--3375.

\bibitem[Van~Hoof et~al.(2016)Van~Hoof, Chen, Karl, van~der Smagt, and Peters]{van2016stable}
H.~Van~Hoof, N.~Chen, M.~Karl, P.~van~der Smagt, and J.~Peters, ``Stable reinforcement learning with autoencoders for tactile and visual data,'' in \emph{IEEE/RSJ international conference on intelligent robots and systems}, 2016, pp. 3928--3934.

\bibitem[Wang et~al.(2022)Wang, Lambeta, Chou, and Calandra]{wang2022tacto}
S.~Wang, M.~Lambeta, P.-W. Chou, and R.~Calandra, ``Tacto: A fast, flexible, and open-source simulator for high-resolution vision-based tactile sensors,'' \emph{IEEE Robotics and Automation Letters}, vol.~7, no.~2, pp. 3930--3937, 2022.

\bibitem[Lin et~al.(2022)Lin, Lloyd, Church, and Lepora]{lin2022tactile}
Y.~Lin, J.~Lloyd, A.~Church, and N.~F. Lepora, ``Tactile gym 2.0: Sim-to-real deep reinforcement learning for comparing low-cost high-resolution robot touch,'' \emph{IEEE Robotics and Automation Letters}, vol.~7, no.~4, pp. 10\,754--10\,761, 2022.

\bibitem[Akinola et~al.(2025)Akinola, Xu, Carius, Fox, and Narang]{akinola2024tacsl}
I.~Akinola, J.~Xu, J.~Carius, D.~Fox, and Y.~Narang, ``Tacsl: A library for visuotactile sensor simulation and learning,'' \emph{IEEE Transactions on Robotics}, p. 1–17, 2025.

\bibitem[Guadarrama-Olvera et~al.(2018)Guadarrama-Olvera, Bergner, Dean, and Cheng]{guadarrama2018enhancing}
J.~R. Guadarrama-Olvera, F.~Bergner, E.~Dean, and G.~Cheng, ``Enhancing biped locomotion on unknown terrain using tactile feedback,'' in \emph{IEEE-RAS International Conference on Humanoid Robots}, 2018, pp. 1--9.

\bibitem[Guo et~al.(2020)Guo, Blaise, Molnar, Coholich, Padte, Zhao, and Hammond]{xiaofeng_foot}
X.~Guo, B.~Blaise, J.~Molnar, J.~Coholich, S.~Padte, Y.~Zhao, and F.~L. Hammond, ``Soft foot sensor design and terrain classification for dynamic legged locomotion,'' in \emph{IEEE International Conference on Soft Robotics}, 2020, pp. 550--557.

\bibitem[Suwanratchatamanee et~al.(2009)Suwanratchatamanee, Matsumoto, and Hashimoto]{suwanratchatamanee2009simple}
K.~Suwanratchatamanee, M.~Matsumoto, and S.~Hashimoto, ``A simple tactile sensing foot for humanoid robot and active ground slope recognition,'' in \emph{IEEE International Conference on Mechatronics}, 2009, pp. 1--6.

\bibitem[Veiga et~al.(2020{\natexlab{b}})Veiga, Akrour, and Peters]{veiga2020hierarchical}
F.~Veiga, R.~Akrour, and J.~Peters, ``Hierarchical tactile-based control decomposition of dexterous in-hand manipulation tasks,'' \emph{Frontiers in Robotics and AI}, vol.~7, p. 521448, 2020.

\bibitem[Maiolino et~al.(2013)Maiolino, Maggiali, Cannata, Metta, and Natale]{maiolino2013flexible}
P.~Maiolino, M.~Maggiali, G.~Cannata, G.~Metta, and L.~Natale, ``A flexible and robust large scale capacitive tactile system for robots,'' \emph{IEEE Sensors Journal}, vol.~13, no.~10, pp. 3910--3917, 2013.

\bibitem[Dean-Leon et~al.(2019)Dean-Leon, Guadarrama-Olvera, Bergner, and Cheng]{dean2019whole}
E.~Dean-Leon, J.~R. Guadarrama-Olvera, F.~Bergner, and G.~Cheng, ``Whole-body active compliance control for humanoid robots with robot skin,'' in \emph{IEEE International Conference on Robotics and Automation}, 2019, pp. 5404--5410.

\bibitem[Kobayashi et~al.(2022)Kobayashi, Dean-Leon, Guadarrama-Olvera, Bergner, and Cheng]{tactile_dance}
T.~Kobayashi, E.~Dean-Leon, J.~R. Guadarrama-Olvera, F.~Bergner, and G.~Cheng, ``Whole-body multicontact haptic human–humanoid interaction based on leader–follower switching: A robot dance of the “box step”,'' \emph{Advanced Intelligent Systems}, vol.~4, no.~2, p. 2100038, 2022.

\bibitem[Yuan et~al.(2017)Yuan, Dong, and Adelson]{yuan2017gelsight}
W.~Yuan, S.~Dong, and E.~H. Adelson, ``Gelsight: High-resolution robot tactile sensors for estimating geometry and force,'' \emph{Sensors}, vol.~17, no.~12, p. 2762, 2017.

\bibitem[Hirai et~al.(1998)Hirai, Hirose, Haikawa, and Takenaka]{Honda_Humanoid}
K.~Hirai, M.~Hirose, Y.~Haikawa, and T.~Takenaka, ``The development of honda humanoid robot,'' in \emph{IEEE International Conference on Robotics and Automation}, vol.~2, 1998, pp. 1321--1326.

\bibitem[Dafarra et~al.(2024)Dafarra, Pattacini, Romualdi, Rapetti, Grieco, Darvish, Milani, Valli, Sorrentino, Viceconte, Scalzo, Traversaro, Sartore, Elobaid, Guedelha, Herron, Leonessa, Draicchio, Metta, Maggiali, and Pucci]{Dafarra_2024}
S.~Dafarra \emph{et~al.}, ``icub3 avatar system: Enabling remote fully immersive embodiment of humanoid robots,'' \emph{Science Robotics}, vol.~9, no.~86, 2024.

\bibitem[Kuindersma et~al.(2016)Kuindersma, Deits, Fallon, Valenzuela, Dai, Permenter, Koolen, Marion, and Tedrake]{Kuindersma2016}
S.~Kuindersma, R.~Deits, M.~Fallon, A.~Valenzuela, H.~Dai, F.~Permenter, T.~Koolen, P.~Marion, and R.~Tedrake, ``Optimization-based locomotion planning, estimation, and control design for the atlas humanoid robot,'' \emph{Autonomous Robots}, vol.~40, no.~3, pp. 429--455, 2016.

\bibitem[Xu et~al.(2021)Xu, Wang, Hao, Zhao, Lin, Jin, and Ding]{xu2021flexible}
Y.~Xu, Z.~Wang, W.~Hao, W.~Zhao, W.~Lin, B.~Jin, and N.~Ding, ``A flexible multimodal sole sensor for legged robot sensing complex ground information during locomotion,'' \emph{Sensors}, vol.~21, no.~16, p. 5359, 2021.

\bibitem[Tyler et~al.(2023)Tyler, Malhotra, Montague, Zhao, HammondIII, and Zhao]{TYLER2023Foot}
T.~Tyler, V.~Malhotra, A.~Montague, Z.~Zhao, F.~L. HammondIII, and Y.~Zhao, ``Integrating reconfigurable foot design, multi-modal contact sensing, and terrain classification for bipedal locomotion,'' \emph{IFAC Modeling, Estimation and Control Conference}, vol.~56, no.~3, pp. 523--528, 2023.

\bibitem[Nori et~al.(2015)Nori, Traversaro, Eljaik, Romano, Del~Prete, and Pucci]{nori2015icub}
F.~Nori, S.~Traversaro, J.~Eljaik, F.~Romano, A.~Del~Prete, and D.~Pucci, ``icub whole-body control through force regulation on rigid non-coplanar contacts,'' \emph{Frontiers in Robotics and AI}, vol.~2, p.~6, 2015.

\bibitem[Kheddar et~al.(2019)Kheddar, Caron, Gergondet, Comport, Tanguy, Ott, Henze, Mesesan, Englsberger, Roa, Wieber, Chaumette, Spindler, Oriolo, Lanari, Escande, Chappellet, Kanehiro, and Rabaté]{AirBus}
A.~Kheddar \emph{et~al.}, ``Humanoid robots in aircraft manufacturing: The airbus use cases,'' \emph{IEEE Robotics Automation Magazine}, vol.~26, no.~4, pp. 30--45, 2019.

\bibitem[Wong et~al.(2022)Wong, Samadi, Suleiman, and Kheddar]{wong2022thmc}
C.~Y. Wong, S.~Samadi, W.~Suleiman, and A.~Kheddar, ``Touch semantics for intuitive physical manipulation of humanoids,'' \emph{IEEE Transactions on Human-Machine Systems}, vol.~52, no.~6, pp. 1111--1121, 2022.

\bibitem[Harada et~al.(2010)Harada, Hattori, Hirukawa, Morisawa, Kajita, and Yoshida]{collision_free_walking_TMECH}
K.~Harada, S.~Hattori, H.~Hirukawa, M.~Morisawa, S.~Kajita, and E.~Yoshida, ``Two-stage time-parametrized gait planning for humanoid robots,'' \emph{IEEE/ASME Transactions on Mechatronics}, vol.~15, no.~5, pp. 694--703, 2010.

\bibitem[Armleder et~al.(2024)Armleder, Dean-Leon, Bergner, Guadarrama~Olvera, and Cheng]{armleder2024tactile}
S.~Armleder, E.~Dean-Leon, F.~Bergner, J.~R. Guadarrama~Olvera, and G.~Cheng, ``Tactile-based negotiation of unknown objects during navigation in unstructured environments with movable obstacles,'' \emph{Advanced Intelligent Systems}, vol.~6, no.~3, p. 2300621, 2024.

\bibitem[Ohmura and Kuniyoshi(2007)]{ohmura2007humanoid}
Y.~Ohmura and Y.~Kuniyoshi, ``Humanoid robot which can lift a 30kg box by whole body contact and tactile feedback,'' in \emph{IEEE/RSJ International Conference on Intelligent Robots and Systems}, 2007, pp. 1136--1141.

\bibitem[Natale and Cannata(2017)]{natale2017tactile}
L.~Natale and G.~Cannata, ``Tactile sensing,'' in \emph{Humanoid Robotics: A Reference}.\hskip 1em plus 0.5em minus 0.4em\relax Springer, 2017, pp. 110--1.

\bibitem[Lengagne et~al.(2013)Lengagne, Vaillant, Yoshida, and Kheddar]{lengagne2013generation}
S.~Lengagne, J.~Vaillant, E.~Yoshida, and A.~Kheddar, ``Generation of whole-body optimal dynamic multi-contact motions,'' \emph{The International Journal of Robotics Research}, vol.~32, no. 9-10, pp. 1104--1119, 2013.

\bibitem[Ponton et~al.(2016)Ponton, Herzog, Schaal, and Righetti]{ponton2016convex}
B.~Ponton, A.~Herzog, S.~Schaal, and L.~Righetti, ``A convex model of humanoid momentum dynamics for multi-contact motion generation,'' in \emph{IEEE-RAS International Conference on Humanoid Robots}, 2016, pp. 842--849.

\bibitem[Bolotnikova et~al.(2020)Bolotnikova, Courtois, and Kheddar]{bolotnikova2020ral}
A.~Bolotnikova, S.~Courtois, and A.~Kheddar, ``Multi-contact planning on humans for physical assistance by humanoid,'' \emph{IEEE Robotics and Automation Letters}, vol.~5, no.~1, pp. 135--142, 2020.

\bibitem[Bouyarmane et~al.(2019{\natexlab{a}})Bouyarmane, Caron, Escande, and Kheddar]{Bouyarmane2019}
K.~Bouyarmane, S.~Caron, A.~Escande, and A.~Kheddar, ``Multi-contact motion planning and control,'' in \emph{Humanoid Robotics: A Reference}.\hskip 1em plus 0.5em minus 0.4em\relax Springer, 2019, pp. 1763--1804.

\bibitem[Posa et~al.(2014)Posa, Cantu, and Tedrake]{posa2014direct}
M.~Posa, C.~Cantu, and R.~Tedrake, ``A direct method for trajectory optimization of rigid bodies through contact,'' \emph{The International Journal of Robotics Research}, vol.~33, no.~1, pp. 69--81, 2014.

\bibitem[Mordatch et~al.(2012)Mordatch, Todorov, and Popovi{\'c}]{mordatch2012discovery}
I.~Mordatch, E.~Todorov, and Z.~Popovi{\'c}, ``Discovery of complex behaviors through contact-invariant optimization,'' \emph{ACM Transactions on Graphics}, vol.~31, no.~4, pp. 1--8, 2012.

\bibitem[Zhu et~al.(2023)Zhu, Meduri, and Righetti]{zhu2023efficient}
H.~Zhu, A.~Meduri, and L.~Righetti, ``Efficient object manipulation planning with monte carlo tree search,'' in \emph{IEEE/RSJ International Conference on Intelligent Robots and Systems}, 2023, pp. 10\,628--10\,635.

\bibitem[Hauser et~al.(2005)Hauser, Bretl, and Latombe]{contact-before-motion}
K.~Hauser, T.~Bretl, and J.-C. Latombe, ``Non-gaited humanoid locomotion planning,'' in \emph{IEEE-RAS International Conference on Humanoid Robots}, 2005, pp. 7--12.

\bibitem[Ferrari et~al.(2019)Ferrari, Scianca, Lanari, and Oriolo]{ferrari2019integrated}
P.~Ferrari, N.~Scianca, L.~Lanari, and G.~Oriolo, ``An integrated motion planner/controller for humanoid robots on uneven ground,'' in \emph{European Control Conference}, 2019, pp. 1598--1603.

\bibitem[Amatucci et~al.(2022)Amatucci, Kim, Hwangbo, and Park]{amatucci2022monte}
L.~Amatucci, J.-H. Kim, J.~Hwangbo, and H.-W. Park, ``Monte carlo tree search gait planner for non-gaited legged system control,'' in \emph{IEEE International Conference on Robotics and Automation}, 2022, pp. 4701--4707.

\bibitem[Taouil et~al.(2024)Taouil, Amatucci, Khadiv, Dai, Barasuol, Turrisi, and Semini]{taouil2024non}
I.~Taouil, L.~Amatucci, M.~Khadiv, A.~Dai, V.~Barasuol, G.~Turrisi, and C.~Semini, ``Non-gaited legged locomotion with monte-carlo tree search and supervised learning,'' \emph{IEEE Robotics and Automation Letters}, pp. 1--8, 2024.

\bibitem[Ravi et~al.(2024)Ravi, Dh{\'e}din, Jordana, Zhu, Meduri, Righetti, Sch{\"o}lkopf, and Khadiv]{ravi2024efficient}
A.~K.~C. Ravi, V.~Dh{\'e}din, A.~Jordana, H.~Zhu, A.~Meduri, L.~Righetti, B.~Sch{\"o}lkopf, and M.~Khadiv, ``Efficient search and learning for agile locomotion on stepping stones,'' \emph{arXiv preprint arXiv:2403.03639}, 2024.

\bibitem[Murooka et~al.(2021{\natexlab{b}})Murooka, Kumagai, Morisawa, Kanehiro, and Kheddar]{Murooka_Search}
M.~Murooka, I.~Kumagai, M.~Morisawa, F.~Kanehiro, and A.~Kheddar, ``Humanoid loco-manipulation planning based on graph search and reachability maps,'' \emph{IEEE Robotics and Automation Letters}, vol.~6, no.~2, pp. 1840--1847, 2021.

\bibitem[Escande et~al.(2013)Escande, Kheddar, and Miossec]{escande2013planning}
A.~Escande, A.~Kheddar, and S.~Miossec, ``Planning contact points for humanoid robots,'' \emph{Robotics and Autonomous Systems}, vol.~61, no.~5, pp. 428--442, 2013.

\bibitem[Janson et~al.(2017)Janson, Schmerling, and Pavone]{janson2017monte}
L.~Janson, E.~Schmerling, and M.~Pavone, ``Monte carlo motion planning for robot trajectory optimization under uncertainty,'' in \emph{Robotics Research: Volume 2}.\hskip 1em plus 0.5em minus 0.4em\relax Springer, 2017, pp. 343--361.

\bibitem[Li et~al.(2021{\natexlab{a}})Li, Miao, Qureshi, and Yip]{li2021mpc}
L.~Li, Y.~Miao, A.~H. Qureshi, and M.~C. Yip, ``Mpc-mpnet: Model-predictive motion planning networks for fast, near-optimal planning under kinodynamic constraints,'' \emph{IEEE Robotics and Automation Letters}, vol.~6, no.~3, pp. 4496--4503, 2021.

\bibitem[Brossette et~al.(2018)Brossette, Escande, and Kheddar]{brossette2018tro}
S.~Brossette, A.~Escande, and A.~Kheddar, ``Multicontact postures computation on manifolds,'' \emph{IEEE Transactions on Robotics}, vol.~34, no.~5, pp. 1252--1265, 2018.

\bibitem[Abi-Farraj et~al.(2019)Abi-Farraj, Henze, Ott, Giordano, and Roa]{TORO_push}
F.~Abi-Farraj, B.~Henze, C.~Ott, P.~R. Giordano, and M.~A. Roa, ``Torque-based balancing for a humanoid robot performing high-force interaction tasks,'' \emph{IEEE Robotics and Automation Letters}, vol.~4, no.~2, pp. 2023--2030, 2019.

\bibitem[Rouxel et~al.(2022)Rouxel, Yuan, Wen, and Li]{rouxel2022multicontact}
Q.~Rouxel, K.~Yuan, R.~Wen, and Z.~Li, ``Multicontact motion retargeting using whole-body optimization of full kinematics and sequential force equilibrium,'' \emph{IEEE/ASME Transactions on Mechatronics}, vol.~27, no.~5, pp. 4188--4198, 2022.

\bibitem[Farnioli et~al.(2015)Farnioli, Gabiccini, and Bicchi]{force_dist_optim}
E.~Farnioli, M.~Gabiccini, and A.~Bicchi, ``Optimal contact force distribution for compliant humanoid robots in whole-body loco-manipulation tasks,'' in \emph{IEEE International Conference on Robotics and Automation}, 2015, pp. 5675--5681.

\bibitem[Li and Nguyen(2023{\natexlab{a}})]{Li_pose_optim}
J.~Li and Q.~Nguyen, ``Kinodynamic pose optimization for humanoid loco-manipulation,'' in \emph{IEEE-RAS International Conference on Humanoid Robots}, 2023, pp. 1--8.

\bibitem[Chatzinikolaidis et~al.(2020)Chatzinikolaidis, You, and Li]{chatzinikolaidis2020contact}
I.~Chatzinikolaidis, Y.~You, and Z.~Li, ``Contact-implicit trajectory optimization using an analytically solvable contact model for locomotion on variable ground,'' \emph{IEEE Robotics and Automation Letters}, vol.~5, no.~4, pp. 6357--6364, 2020.

\bibitem[Manchester and Kuindersma(2020)]{manchester2020variational}
Z.~Manchester and S.~Kuindersma, ``Variational contact-implicit trajectory optimization,'' in \emph{Robotics Research: The International Symposium}.\hskip 1em plus 0.5em minus 0.4em\relax Springer, 2020, pp. 985--1000.

\bibitem[Posa et~al.(2016)Posa, Kuindersma, and Tedrake]{posa2016optimization}
M.~Posa, S.~Kuindersma, and R.~Tedrake, ``Optimization and stabilization of trajectories for constrained dynamical systems,'' in \emph{IEEE International Conference on Robotics and Automation}, 2016, pp. 1366--1373.

\bibitem[Mastalli et~al.(2016)Mastalli, Havoutis, Focchi, Caldwell, and Semini]{mastalli2016hierarchical}
C.~Mastalli, I.~Havoutis, M.~Focchi, D.~G. Caldwell, and C.~Semini, ``Hierarchical planning of dynamic movements without scheduled contact sequences,'' in \emph{IEEE International Conference on Robotics and Automation}, 2016, pp. 4636--4641.

\bibitem[Chen et~al.(2023)Chen, Chen, Dong, Yu, and Huang]{chen2023online}
H.~Chen, X.~Chen, C.~Dong, Z.~Yu, and Q.~Huang, ``Online running pattern generation for humanoid robot with direct collocation of reference-tracking dynamics,'' \emph{IEEE/ASME Transactions on Mechatronics}, pp. 2091--2102, 2023.

\bibitem[Tassa et~al.(2012{\natexlab{a}})Tassa, Erez, and Todorov]{tassa2012synthesis}
Y.~Tassa, T.~Erez, and E.~Todorov, ``Synthesis and stabilization of complex behaviors through online trajectory optimization,'' in \emph{2012 IEEE/RSJ International Conference on Intelligent Robots and Systems}, 2012, pp. 4906--4913.

\bibitem[Carius et~al.(2018)Carius, Ranftl, Koltun, and Hutter]{carius2018trajectory}
J.~Carius, R.~Ranftl, V.~Koltun, and M.~Hutter, ``Trajectory optimization with implicit hard contacts,'' \emph{IEEE Robotics and Automation Letters}, vol.~3, no.~4, pp. 3316--3323, 2018.

\bibitem[Le~Cleac'h et~al.(2024)Le~Cleac'h, Howell, Yang, Lee, Zhang, Bishop, Schwager, and Manchester]{le2024fast}
S.~Le~Cleac'h, T.~A. Howell, S.~Yang, C.-Y. Lee, J.~Zhang, A.~Bishop, M.~Schwager, and Z.~Manchester, ``Fast contact-implicit model predictive control,'' \emph{IEEE Transactions on Robotics}, vol.~40, pp. 1617--1629, 2024.

\bibitem[Kong et~al.(2023)Kong, Li, Council, and Johnson]{kong2023hybrid}
N.~J. Kong, C.~Li, G.~Council, and A.~M. Johnson, ``Hybrid i{L}{Q}{R} model predictive control for contact implicit stabilization on legged robots,'' \emph{IEEE Transactions on Robotics}, vol.~39, no.~6, pp. 4712--4727, 2023.

\bibitem[Aydinoglu et~al.(2024)Aydinoglu, Wei, Huang, and Posa]{aydinoglu2023consensus}
A.~Aydinoglu, A.~Wei, W.-C. Huang, and M.~Posa, ``Consensus complementarity control for multi-contact mpc,'' \emph{IEEE Transactions on Robotics}, pp. 1--18, 2024.

\bibitem[Kurtz et~al.(2023)Kurtz, Castro, {\"O}nol, and Lin]{kurtz2023inverse}
V.~Kurtz, A.~Castro, A.~{\"O}. {\"O}nol, and H.~Lin, ``Inverse dynamics trajectory optimization for contact-implicit model predictive control,'' \emph{arXiv preprint arXiv:2309.01813}, 2023.

\bibitem[Esteban et~al.(2025)Esteban, Kurtz, Ghansah, and Ames]{esteban2025reduced}
S.~A. Esteban, V.~Kurtz, A.~B. Ghansah, and A.~D. Ames, ``Reduced-order model guided contact-implicit model predictive control for humanoid locomotion,'' \emph{arXiv preprint arXiv:2502.15630}, 2025.

\bibitem[Margolis et~al.(2022)Margolis, Chen, Paigwar, Fu, Kim, Kim, and Agrawal]{pmlr-v164-margolis22a}
G.~B. Margolis, T.~Chen, K.~Paigwar, X.~Fu, D.~Kim, S.~b. Kim, and P.~Agrawal, ``Learning to jump from pixels,'' in \emph{Proceedings of the Conference on Robot Learning}, vol. 164, 2022, pp. 1025--1034.

\bibitem[Tsounis et~al.(2020)Tsounis, Alge, Lee, Farshidian, and Hutter]{tsounis2020deepgait}
V.~Tsounis, M.~Alge, J.~Lee, F.~Farshidian, and M.~Hutter, ``Deepgait: Planning and control of quadrupedal gaits using deep reinforcement learning,'' \emph{IEEE Robotics and Automation Letters}, vol.~5, no.~2, pp. 3699--3706, 2020.

\bibitem[Lin et~al.(2019)Lin, Ponton, Righetti, and Berenson]{lin2019efficient}
Y.-C. Lin, B.~Ponton, L.~Righetti, and D.~Berenson, ``Efficient humanoid contact planning using learned centroidal dynamics prediction,'' in \emph{IEEE International Conference on Robotics and Automation}, 2019, pp. 5280--5286.

\bibitem[Xu et~al.(2023)Xu, Li, Wang, and Gui]{interdiff}
S.~Xu, Z.~Li, Y.-X. Wang, and L.-Y. Gui, ``Interdiff: Generating 3d human-object interactions with physics-informed diffusion,'' in \emph{Proceedings of the IEEE/CVF International Conference on Computer Vision}, 2023, pp. 14\,928--14\,940.

\bibitem[Li et~al.(2023{\natexlab{c}})Li, Wu, and Liu]{object_guided_motion_tog2023}
J.~Li, J.~Wu, and C.~K. Liu, ``Object motion guided human motion synthesis,'' \emph{ACM Transactions on Graphics}, vol.~42, no.~6, pp. 1--11, 2023.

\bibitem[Bahl et~al.(2023)Bahl, Mendonca, Chen, Jain, and Pathak]{bahl2023affordances}
S.~Bahl, R.~Mendonca, L.~Chen, U.~Jain, and D.~Pathak, ``Affordances from human videos as a versatile representation for robotics,'' in \emph{Proceedings of the IEEE/CVF Conference on Computer Vision and Pattern Recognition}, 2023, pp. 13\,778--13\,790.

\bibitem[Chignoli et~al.(2021)Chignoli, Kim, Stanger-Jones, and Kim]{chignoli2021humanoid}
M.~Chignoli, D.~Kim, E.~Stanger-Jones, and S.~Kim, ``The mit humanoid robot: Design, motion planning, and control for acrobatic behaviors,'' in \emph{IEEE-RAS International Conference on Humanoid Robots}, 2021, pp. 1--8.

\bibitem[Romualdi et~al.(2022)Romualdi, Dafarra, L'Erario, Sorrentino, Traversaro, and Pucci]{romualdi2022online}
G.~Romualdi, S.~Dafarra, G.~L'Erario, I.~Sorrentino, S.~Traversaro, and D.~Pucci, ``Online non-linear centroidal mpc for humanoid robot locomotion with step adjustment,'' in \emph{International Conference on Robotics and Automation}, 2022, pp. 10\,412--10\,419.

\bibitem[Elobaid et~al.(2023)Elobaid, Romualdi, Nava, Rapetti, Mohamed, and Pucci]{elobaid2023online}
M.~Elobaid, G.~Romualdi, G.~Nava, L.~Rapetti, H.~A.~O. Mohamed, and D.~Pucci, ``Online non-linear centroidal mpc for humanoid robots payload carrying with contact-stable force parametrization,'' in \emph{IEEE International Conference on Robotics and Automation}, 2023, pp. 12\,233--12\,239.

\bibitem[Li and Nguyen(2021)]{li2021force}
J.~Li and Q.~Nguyen, ``Force-and-moment-based model predictive control for achieving highly dynamic locomotion on bipedal robots,'' in \emph{IEEE Conference on Decision and Control}, 2021, pp. 1024--1030.

\bibitem[Henze et~al.(2014)Henze, Ott, and Roa]{henze2014posture}
B.~Henze, C.~Ott, and M.~A. Roa, ``Posture and balance control for humanoid robots in multi-contact scenarios based on model predictive control,'' in \emph{IEEE/RSJ International Conference on Intelligent Robots and Systems}, 2014, pp. 3253--3258.

\bibitem[Bang et~al.(2025)Bang, Gonzalez, Moore, Kang, Seo, Gupta, and Sentis]{bang2024rpc}
S.~H. Bang, C.~Gonzalez, G.~Moore, D.~H. Kang, M.~Seo, R.~Gupta, and L.~Sentis, ``Rpc: A modular framework for robot planning, control, and deployment,'' in \emph{IEEE/SICE International Symposium on System Integration}, 2025, pp. 1142--1148.

\bibitem[Penco et~al.(2019)Penco, Scianca, Modugno, Lanari, Oriolo, and Ivaldi]{penco2019multimode}
L.~Penco, N.~Scianca, V.~Modugno, L.~Lanari, G.~Oriolo, and S.~Ivaldi, ``A multimode teleoperation framework for humanoid loco-manipulation: An application for the icub robot,'' \emph{IEEE Robotics Automation Magazine}, vol.~26, no.~4, pp. 73--82, 2019.

\bibitem[Dai et~al.(2014)Dai, Valenzuela, and Tedrake]{Dai_centroidal_kinematics}
H.~Dai, A.~Valenzuela, and R.~Tedrake, ``Whole-body motion planning with centroidal dynamics and full kinematics,'' in \emph{2014 IEEE-RAS International Conference on Humanoid Robots}, 2014, pp. 295--302.

\bibitem[Polverini et~al.(2020)Polverini, Laurenzi, Hoffman, Ruscelli, and Tsagarakis]{polverini2020multi}
M.~P. Polverini, A.~Laurenzi, E.~M. Hoffman, F.~Ruscelli, and N.~G. Tsagarakis, ``Multi-contact heavy object pushing with a centaur-type humanoid robot: Planning and control for a real demonstrator,'' \emph{IEEE Robotics and Automation Letters}, vol.~5, no.~2, pp. 859--866, 2020.

\bibitem[Dantec et~al.(2021)Dantec, Budhiraja, Roig, Lembono, Saurel, Stasse, Fernbach, Tonneau, Vijayakumar, Calinon, Taix, and Mansard]{Memmo_2021}
E.~Dantec \emph{et~al.}, ``Whole body model predictive control with a memory of motion: Experiments on a torque-controlled talos,'' in \emph{International Conference on Robotics and Automation}, 2021, pp. 8202--8208.

\bibitem[Audren et~al.(2014)Audren, Vaillant, Kheddar, Escande, Kaneko, and Yoshida]{audren2014iros}
H.~Audren, J.~Vaillant, A.~Kheddar, A.~Escande, K.~Kaneko, and E.~Yoshida, ``Model preview control in multi-contact motion-application to a humanoid robot,'' in \emph{IEEE/RSJ International Conference on Intelligent Robots and Systems}, 2014, pp. 4030--4035.

\bibitem[Orin et~al.(2013)Orin, Goswami, and Lee]{orin2013centroidal}
D.~E. Orin, A.~Goswami, and S.-H. Lee, ``Centroidal dynamics of a humanoid robot,'' \emph{Autonomous robots}, vol.~35, pp. 161--176, 2013.

\bibitem[Li et~al.(2021{\natexlab{b}})Li, Frei, and Wensing]{li2021model}
H.~Li, R.~J. Frei, and P.~M. Wensing, ``Model hierarchy predictive control of robotic systems,'' \emph{IEEE Robotics and Automation Letters}, vol.~6, no.~2, pp. 3373--3380, 2021.

\bibitem[Wang et~al.(2021)Wang, Kim, Vijayakumar, and Tonneau]{Wang_Multi-Fidelity}
J.~Wang, S.~Kim, S.~Vijayakumar, and S.~Tonneau, ``Multi-fidelity receding horizon planning for multi-contact locomotion,'' in \emph{IEEE-RAS International Conference on Humanoid Robots}, 2021, pp. 53--60.

\bibitem[Li et~al.(2024{\natexlab{c}})Li, Le, Ma, and Nguyen]{li2024adapting}
J.~Li, Z.~Le, J.~Ma, and Q.~Nguyen, ``Adapting gait frequency for posture-regulating humanoid push-recovery via hierarchical model predictive control,'' \emph{arXiv preprint arXiv:2409.14342}, 2024.

\bibitem[Budhiraja et~al.(2019)Budhiraja, Carpentier, and Mansard]{Budhiraja_Consensus}
R.~Budhiraja, J.~Carpentier, and N.~Mansard, ``Dynamics consensus between centroidal and whole-body models for locomotion of legged robots,'' in \emph{IEEE International Conference on Robotics and Automation}, 2019, pp. 6727--6733.

\bibitem[Herzog et~al.(2016)Herzog, Schaal, and Righetti]{herzog2016structured}
A.~Herzog, S.~Schaal, and L.~Righetti, ``Structured contact force optimization for kino-dynamic motion generation,'' in \emph{Proc. IEEE/RSJ Int. Conf. Intell. Robots Syst.}, 2016, pp. 2703--2710.

\bibitem[Mayne(1966)]{original_DDP}
D.~Mayne, ``A second-order gradient method for determining optimal trajectories of non-linear discrete-time systems,'' \emph{International Journal of Control}, vol.~3, no.~1, pp. 85--95, 1966.

\bibitem[Tassa et~al.(2012{\natexlab{b}})Tassa, Erez, and Todorov]{Tassa_iLQR}
Y.~Tassa, T.~Erez, and E.~Todorov, ``Synthesis and stabilization of complex behaviors through online trajectory optimization,'' in \emph{IEEE/RSJ International Conference on Intelligent Robots and Systems}, 2012, pp. 4906--4913.

\bibitem[Rall(1981)]{rall1981automatic}
L.~B. Rall, \emph{Automatic differentiation: Techniques and applications}.\hskip 1em plus 0.5em minus 0.4em\relax Springer, 1981.

\bibitem[R{\"o}smann et~al.(2018)R{\"o}smann, Kr{\"a}mer, Makarow, Hoffmann, and Bertram]{rosmann2018exploiting}
C.~R{\"o}smann, M.~Kr{\"a}mer, A.~Makarow, F.~Hoffmann, and T.~Bertram, ``Exploiting sparse structures in nonlinear model predictive control with hypergraphs,'' in \emph{IEEE/ASME International Conference on Advanced Intelligent Mechatronics}, 2018, pp. 1332--1337.

\bibitem[Vanroye et~al.(2023)Vanroye, Sathya, De~Schutter, and Decr{\'e}]{vanroye2023fatrop}
L.~Vanroye, A.~Sathya, J.~De~Schutter, and W.~Decr{\'e}, ``Fatrop: A fast constrained optimal control problem solver for robot trajectory optimization and control,'' in \emph{IEEE/RSJ International Conference on Intelligent Robots and Systems}, 2023, pp. 10\,036--10\,043.

\bibitem[Callens et~al.(2024)Callens, Gillis, Decr{\'e}, and Swevers]{callens2024adaptivenlp}
L.~Callens, J.~Gillis, W.~Decr{\'e}, and J.~Swevers, ``Adaptivenlp: a framework for efficient online adaptability in nlp structures for optimal control problems,'' in \emph{European Control Conference}, 2024, pp. 1358--1365.

\bibitem[Wilson(1963)]{wilsonSQP}
R.~B. Wilson, ``A simplicial method for convex programming,'' \emph{Ph.D. Thesis}, 1963.

\bibitem[Zhakatayev et~al.(2017)Zhakatayev, Rakhim, Adiyatov, Baimyshev, and Varol]{zhakatayev_successive_2017}
A.~Zhakatayev, B.~Rakhim, O.~Adiyatov, A.~Baimyshev, and H.~A. Varol, ``Successive linearization based model predictive control of variable stiffness actuated robots,'' in \emph{IEEE International Conference on Advanced Intelligent Mechatronics}, 2017, pp. 1774--1779.

\bibitem[Jerez et~al.(2011)Jerez, Kerrigan, and Constantinides]{jerez2011condensed}
J.~L. Jerez, E.~C. Kerrigan, and G.~A. Constantinides, ``A condensed and sparse qp formulation for predictive control,'' in \emph{2011 50th IEEE Conference on Decision and Control and European Control Conference}.\hskip 1em plus 0.5em minus 0.4em\relax IEEE, 2011, pp. 5217--5222.

\bibitem[Bishop et~al.(2024)Bishop, Zhang, Gurumurthy, Tracy, and Manchester]{bishop2023relu}
A.~Bishop, J.~Zhang, S.~Gurumurthy, K.~Tracy, and Z.~Manchester, ``Relu-qp: A gpu-accelerated quadratic programming solver for model-predictive control,'' in \emph{IEEE International Conference on Robotics and Automation}, 2024, pp. 13\,285--13\,292.

\bibitem[Da and Grizzle(2019)]{gait_library_IJRR}
X.~Da and J.~Grizzle, ``Combining trajectory optimization, supervised machine learning, and model structure for mitigating the curse of dimensionality in the control of bipedal robots,'' \emph{The International Journal of Robotics Research}, vol.~38, no.~9, pp. 1063--1097, 2019.

\bibitem[Diehl et~al.(2006)Diehl, Bock, Diedam, and Wieber]{Diehl2006}
M.~Diehl, H.~G. Bock, H.~Diedam, and P.-B. Wieber, ``Fast direct multiple shooting algorithms for optimal robot control,'' \emph{Fast motions in biomechanics and robotics: optimization and feedback control}, pp. 65--93, 2006.

\bibitem[Betts(2010)]{Betts}
J.~T. Betts, \emph{Practical Methods for Optimal Control and Estimation Using Nonlinear Programming, Second Edition}, 2010.

\bibitem[Dantec et~al.(2022)Dantec, Naveau, Fernbach, Villa, Saurel, Stasse, Taix, and Mansard]{WB_MPC_Dantec}
E.~Dantec, M.~Naveau, P.~Fernbach, N.~Villa, G.~Saurel, O.~Stasse, M.~Taix, and N.~Mansard, ``Whole-body model predictive control for biped locomotion on a torque-controlled humanoid robot,'' in \emph{IEEE-RAS International Conference on Humanoid Robots}, 2022, pp. 638--644.

\bibitem[Ding et~al.(2019)Ding, Pandala, and Park]{ding_real-time_2019}
Y.~Ding, A.~Pandala, and H.-W. Park, ``Real-time model predictive control for versatile dynamic motions in quadrupedal robots,'' in \emph{International Conference on Robotics and Automation}, 2019, pp. 8484--8490.

\bibitem[Williams et~al.(2016)Williams, Drews, Goldfain, Rehg, and Theodorou]{williams2016aggressive}
G.~Williams, P.~Drews, B.~Goldfain, J.~M. Rehg, and E.~A. Theodorou, ``Aggressive driving with model predictive path integral control,'' in \emph{IEEE International Conference on Robotics and Automation}, 2016, pp. 1433--1440.

\bibitem[Carius et~al.(2022)Carius, Ranftl, Farshidian, and Hutter]{Carius_MPPI2022}
J.~Carius, R.~Ranftl, F.~Farshidian, and M.~Hutter, ``Constrained stochastic optimal control with learned importance sampling: A path integral approach,'' \emph{The International Journal of Robotics Research}, vol.~41, no.~2, pp. 189--209, 2022.

\bibitem[Howell et~al.(2022)Howell, Gileadi, Tunyasuvunakool, Zakka, Erez, and Tassa]{howell2022}
T.~Howell, N.~Gileadi, S.~Tunyasuvunakool, K.~Zakka, T.~Erez, and Y.~Tassa, ``Predictive sampling: Real-time behaviour synthesis with mujoco,'' \emph{arXiv preprint arXiv:2212.00541}, 2022.

\bibitem[Todorov et~al.(2012)Todorov, Erez, and Tassa]{todorov2012mujoco}
E.~Todorov, T.~Erez, and Y.~Tassa, ``Mujoco: A physics engine for model-based control,'' in \emph{IEEE/RSJ International Conference on Intelligent Robots and Systems}, 2012, pp. 5026--5033.

\bibitem[Mittal et~al.(2023)Mittal, Yu, Yu, Liu, Rudin, Hoeller, Yuan, Singh, Guo, Mazhar, Mandlekar, Babich, State, Hutter, and Garg]{Orbit}
M.~Mittal \emph{et~al.}, ``Orbit: A unified simulation framework for interactive robot learning environments,'' \emph{IEEE Robotics and Automation Letters}, vol.~8, no.~6, p. 3740–3747, 2023.

\bibitem[Turrisi et~al.(2024)Turrisi, Modugno, Amatucci, Kanoulas, and Semini]{turrisi2024benefits}
G.~Turrisi, V.~Modugno, L.~Amatucci, D.~Kanoulas, and C.~Semini, ``On the benefits of gpu sample-based stochastic predictive controllers for legged locomotion,'' \emph{2024 IEEE/RSJ International Conference on Intelligent Robots and Systems}, pp. 13\,757--13\,764, 2024.

\bibitem[Adu-Bredu et~al.(2023)Adu-Bredu, Gibson, and Grizzle]{Fabrics_WBC}
A.~Adu-Bredu, G.~Gibson, and J.~Grizzle, ``Exploring kinodynamic fabrics for reactive whole-body control of underactuated humanoid robots,'' in \emph{IEEE/RSJ International Conference on Intelligent Robots and Systems}, 2023, pp. 10\,397--10\,404.

\bibitem[Carpentier and Mansard(2018)]{CarpentierTRO}
J.~Carpentier and N.~Mansard, ``Multicontact locomotion of legged robots,'' \emph{IEEE Transactions on Robotics}, vol.~34, no.~6, pp. 1441--1460, 2018.

\bibitem[Otani and Bouyarmane(2017)]{Otani2017Retarget}
K.~Otani and K.~Bouyarmane, ``Adaptive whole-body manipulation in human-to-humanoid multi-contact motion retargeting,'' in \emph{IEEE-RAS International Conference on Humanoid Robotics}, 2017, pp. 446--453.

\bibitem[Harada et~al.(2007)Harada, Kajita, Kanehiro, Fujiwara, Kaneko, Yokoi, and Hirukawa]{Harada_HPR_Table_TMECH}
K.~Harada, S.~Kajita, F.~Kanehiro, K.~Fujiwara, K.~Kaneko, K.~Yokoi, and H.~Hirukawa, ``Real-time planning of humanoid robot's gait for force-controlled manipulation,'' \emph{IEEE/ASME Transactions on Mechatronics}, vol.~12, no.~1, pp. 53--62, 2007.

\bibitem[Foster et~al.(2024)Foster, McCrory, DeBuys, Bertrand, and Griffin]{foster2024physically}
J.~Foster, S.~McCrory, C.~DeBuys, S.~Bertrand, and R.~J. Griffin, ``Physically consistent online inertial adaptation for humanoid loco-manipulation,'' \emph{2024 IEEE/RSJ International Conference on Intelligent Robots and Systems}, pp. 11\,278--11\,285, 2024.

\bibitem[Nozawa et~al.(2010)Nozawa, Ueda, Kakiuchi, Okada, and Inaba]{object_mass_estimation_iros2010}
S.~Nozawa, R.~Ueda, Y.~Kakiuchi, K.~Okada, and M.~Inaba, ``A full-body motion control method for a humanoid robot based on on-line estimation of the operational force of an object with an unknown weight,'' in \emph{2010 IEEE/RSJ International Conference on Intelligent Robots and Systems}, 2010, pp. 2684--2691.

\bibitem[Bang et~al.(2024)Bang, Lee, Gonzalez, and Sentis]{Bang2024VariableIM}
S.~H. Bang, J.~Lee, C.~Gonzalez, and L.~Sentis, ``Variable inertia model predictive control for fast bipedal maneuvers,'' in \emph{IEEE Conference on Decision and Control}, 2024, pp. 4334--4341.

\bibitem[Gao et~al.(2024{\natexlab{a}})Gao, Paredes, Gong, He, Hereid, and Gu]{gao2024timevaryingfootplacementcontrolunderactuated}
Y.~Gao, V.~Paredes, Y.~Gong, Z.~He, A.~Hereid, and Y.~Gu, ``Time-varying foot-placement control for underactuated humanoid walking on swaying rigid surfaces,'' \emph{arXiv preprint arXiv:2409.08371}, 2024.

\bibitem[Lefèvre et~al.(2025)Lefèvre, Chaki, Kawakami, Tanguy, Yoshiike, and Kheddar]{lefevre2025ral}
H.~Lefèvre, T.~Chaki, T.~Kawakami, A.~Tanguy, T.~Yoshiike, and A.~Kheddar, ``Humanoid-human sit-to-stand-to-sit assistance,'' \emph{IEEE Robotics and Automation Letters}, vol.~10, 2025.

\bibitem[Zhu et~al.(2018)Zhu, Navarro, Fraisse, Crosnier, and Cherubini]{zhu2018dual}
J.~Zhu, B.~Navarro, P.~Fraisse, A.~Crosnier, and A.~Cherubini, ``Dual-arm robotic manipulation of flexible cables,'' in \emph{IEEE/RSJ International Conference on Intelligent Robots and Systems}, 2018, pp. 479--484.

\bibitem[Saha and Isto(2007)]{saha2007manipulation}
M.~Saha and P.~Isto, ``Manipulation planning for deformable linear objects,'' \emph{IEEE Transactions on Robotics}, vol.~23, no.~6, pp. 1141--1150, 2007.

\bibitem[Qin et~al.(2023)Qin, Escande, Kanehiro, and Yoshida]{qin2023dual}
Y.~Qin, A.~Escande, F.~Kanehiro, and E.~Yoshida, ``Dual-arm mobile manipulation planning of a long deformable object in industrial installation,'' \emph{IEEE Robotics and Automation Letters}, vol.~8, no.~5, pp. 3039--3046, 2023.

\bibitem[Zhang et~al.(2024{\natexlab{a}})Zhang, Li, Hauser, and Li]{zhang2024adaptigraph}
K.~Zhang, B.~Li, K.~Hauser, and Y.~Li, ``Adaptigraph: Material-adaptive graph-based neural dynamics for robotic manipulation,'' \emph{arXiv preprint arXiv:2407.07889}, 2024.

\bibitem[Li and Nguyen(2023{\natexlab{b}})]{li2023multi}
J.~Li and Q.~Nguyen, ``Multi-contact mpc for dynamic loco-manipulation on humanoid robots,'' in \emph{American Control Conference}, 2023, pp. 1215--1220.

\bibitem[Hsu et~al.(2015)Hsu, Xu, and Ames]{CBF}
S.-C. Hsu, X.~Xu, and A.~D. Ames, ``Control barrier function based quadratic programs with application to bipedal robotic walking,'' in \emph{American Control Conference}, 2015, pp. 4542--4548.

\bibitem[Khatib(1987)]{khatib1987unified}
O.~Khatib, ``A unified approach for motion and force control of robot manipulators: The operational space formulation,'' \emph{IEEE Journal on Robotics and Automation}, vol.~3, no.~1, pp. 43--53, 1987.

\bibitem[Hopkins et~al.(2015)Hopkins, Hong, and Leonessa]{hopkins2015compliant}
M.~A. Hopkins, D.~W. Hong, and A.~Leonessa, ``Compliant locomotion using whole-body control and divergent component of motion tracking,'' in \emph{IEEE International Conference on Robotics and Automation}, 2015, pp. 5726--5733.

\bibitem[Koolen et~al.(2016)Koolen, Bertrand, Thomas, De~Boer, Wu, Smith, Englsberger, and Pratt]{koolen2016design}
T.~Koolen, S.~Bertrand, G.~Thomas, T.~De~Boer, T.~Wu, J.~Smith, J.~Englsberger, and J.~Pratt, ``Design of a momentum-based control framework and application to the humanoid robot atlas,'' \emph{International Journal of Humanoid Robotics}, vol.~13, no.~01, p. 1650007, 2016.

\bibitem[Marew et~al.(2023)Marew, Lvovsky, Yu, Sessions, and Kim]{RMP}
D.~Marew, M.~Lvovsky, S.~Yu, S.~Sessions, and D.~Kim, ``Integration of riemannian motion policy with whole-body control for collision-free legged locomotion,'' in \emph{IEEE-RAS International Conference on Humanoid Robots}, 2023, pp. 1--8.

\bibitem[Yu et~al.(2024{\natexlab{b}})Yu, Perera, Marew, and Kim]{yu2024learning}
S.~Yu, N.~Perera, D.~Marew, and D.~Kim, ``Learning generic and dynamic locomotion of humanoids across discrete terrains,'' in \emph{International Conference on Humanoid Robots}, 2024, pp. 1048--1055.

\bibitem[Mistry et~al.(2008)Mistry, Nakanishi, Cheng, and Schaal]{Mistry_humanoidIK}
M.~Mistry, J.~Nakanishi, G.~Cheng, and S.~Schaal, ``Inverse kinematics with floating base and constraints for full body humanoid robot control,'' in \emph{IEEE-RAS International Conference on Humanoid Robots}, 2008, pp. 22--27.

\bibitem[Darvish et~al.(2023)Darvish, Penco, Ramos, Cisneros, Pratt, Yoshida, Ivaldi, and Pucci]{TRO_teleop_survey}
K.~Darvish, L.~Penco, J.~Ramos, R.~Cisneros, J.~Pratt, E.~Yoshida, S.~Ivaldi, and D.~Pucci, ``Teleoperation of humanoid robots: A survey,'' \emph{IEEE Transactions on Robotics}, vol.~39, no.~3, pp. 1706--1727, 2023.

\bibitem[Marion et~al.(2018)Marion, Fallon, Deits, Valenzuela, D’Arpino, Izatt, Manuelli, Antone, Dai, Koolen, et~al.]{marion2018director}
P.~Marion \emph{et~al.}, ``Director: A user interface designed for robot operation with shared autonomy,'' \emph{The DARPA Robotics Challenge Finals: Humanoid Robots To The Rescue}, pp. 237--270, 2018.

\bibitem[Penco et~al.(2018)Penco, Clement, Modugno, Mingo~Hoffman, Nava, Pucci, Tsagarakis, Mouret, and Ivaldi]{penco_2018_retarget}
L.~Penco, B.~Clement, V.~Modugno, E.~Mingo~Hoffman, G.~Nava, D.~Pucci, N.~G. Tsagarakis, J.~B. Mouret, and S.~Ivaldi, ``Robust real-time whole-body motion retargeting from human to humanoid,'' in \emph{IEEE-RAS International Conference on Humanoid Robots}, 2018, pp. 425--432.

\bibitem[Oh et~al.(2021)Oh, Sim, Cho, Lee, and Oh]{Oh_retarget_TMECH}
J.~Oh, O.~Sim, B.~Cho, K.~Lee, and J.-H. Oh, ``Online delayed reference generation for a humanoid imitating human walking motion,'' \emph{IEEE/ASME Transactions on Mechatronics}, vol.~26, no.~1, pp. 102--112, 2021.

\bibitem[Jorgensen et~al.(2022)Jorgensen, Wonsick, Paterson, Watson, Chase, and Mehling]{jorgensen_2022_cockpit_vr}
S.~J. Jorgensen, M.~Wonsick, M.~Paterson, A.~Watson, I.~Chase, and J.~S. Mehling, ``Cockpit interface for locomotion and manipulation control of the nasa valkyrie humanoid in virtual reality (vr),'' White Paper, 2022.

\bibitem[Wallace et~al.(2020)Wallace, He, Vaz, Georgescu, and Oh]{wallace2020multimodal}
D.~Wallace, Y.~H. He, J.~C. Vaz, L.~Georgescu, and P.~Y. Oh, ``Multimodal teleoperation of heterogeneous robots within a construction environment,'' in \emph{IEEE/RSJ International Conference on Intelligent Robots and Systems}, 2020, pp. 2698--2705.

\bibitem[Abi-Farrajl et~al.(2018)Abi-Farrajl, Henze, Werner, Panzirsch, Ott, and Roa]{abi2018humanoid}
F.~Abi-Farrajl, B.~Henze, A.~Werner, M.~Panzirsch, C.~Ott, and M.~A. Roa, ``Humanoid teleoperation using task-relevant haptic feedback,'' in \emph{IEEE/RSJ International Conference on Intelligent Robots and Systems}, 2018, pp. 5010--5017.

\bibitem[Ramos and Kim(2018)]{ramos2018humanoid}
J.~Ramos and S.~Kim, ``Humanoid dynamic synchronization through whole-body bilateral feedback teleoperation,'' \emph{IEEE Transactions on Robotics}, vol.~34, no.~4, pp. 953--965, 2018.

\bibitem[Colin et~al.(2023)Colin, Sim, and Ramos]{colin2023bipedal}
G.~Colin, Y.~Sim, and J.~Ramos, ``Bipedal robot walking control using human whole-body dynamic telelocomotion,'' in \emph{IEEE International Conference on Robotics and Automation}, 2023, pp. 12\,191--12\,197.

\bibitem[Sentis(2007)]{Sentis2007}
L.~Sentis, ``Synthesis and control of whole-body behaviors in humanoid systems,'' Ph.D. dissertation, Stanford University, 2007.

\bibitem[Mistry et~al.(2010)Mistry, Buchli, and Schaal]{Mistry2010}
M.~Mistry, J.~Buchli, and S.~Schaal, ``Inverse dynamics control of floating base systems using orthogonal decomposition,'' in \emph{International Conference on Robotics and Automation}, 2010, pp. 3406--3412.

\bibitem[Righetti et~al.(2011)Righetti, Buchli, Mistry, and Schaal]{IDC_unify}
L.~Righetti, J.~Buchli, M.~Mistry, and S.~Schaal, ``Inverse dynamics control of floating-base robots with external constraints: A unified view,'' in \emph{2011 IEEE International Conference on Robotics and Automation}, 2011, pp. 1085--1090.

\bibitem[Nakanishi et~al.(2008)Nakanishi, Cory, Mistry, Peters, and Schaal]{mistry_OSC_2008}
J.~Nakanishi, R.~Cory, M.~Mistry, J.~Peters, and S.~Schaal, ``Operational space control: A theoretical and empirical comparison,'' \emph{The International Journal of Robotics Research}, vol.~27, no.~6, pp. 737--757, 2008.

\bibitem[Mansard et~al.(2009)Mansard, Khatib, and Kheddar]{SoT_unilateral}
N.~Mansard, O.~Khatib, and A.~Kheddar, ``A unified approach to integrate unilateral constraints in the stack of tasks,'' \emph{IEEE Transactions on Robotics}, vol.~25, no.~3, pp. 670--685, 2009.

\bibitem[Keith et~al.(2011)Keith, Wieber, Mansard, and Kheddar]{keith2011iros}
F.~Keith, P.-B. Wieber, N.~Mansard, and A.~Kheddar, ``Analysis of the discontinuities in prioritized tasks-space control under discreet task scheduling operations,'' in \emph{IEEE/RSJ International Conference on Intelligent Robots and Systems}, 2011, pp. 3887--3892.

\bibitem[Rocchi et~al.(2015)Rocchi, Hoffman, Caldwell, and Tsagarakis]{rocchi_opensot_2015}
A.~Rocchi, E.~M. Hoffman, D.~G. Caldwell, and N.~G. Tsagarakis, ``{OpenSoT}: {A} whole-body control library for the compliant humanoid robot {COMAN},'' in \emph{{IEEE} {International} {Conference} on {Robotics} and {Automation}}, 2015, pp. 6248--6253.

\bibitem[Escande et~al.(2014)Escande, Mansard, and Wieber]{Escande_HQP_humanoid}
A.~Escande, N.~Mansard, and P.-B. Wieber, ``Hierarchical quadratic programming: Fast online humanoid-robot motion generation,'' \emph{The International Journal of Robotics Research}, vol.~33, no.~7, pp. 1006--1028, 2014.

\bibitem[Bouyarmane and Kheddar(2011)]{bouyarmane_using_2011}
K.~Bouyarmane and A.~Kheddar, ``\BIBforeignlanguage{en}{Using a multi-objective controller to synthesize simulated humanoid robot motion with changing contact configurations},'' in \emph{\BIBforeignlanguage{en}{{IEEE}/{RSJ} {International} {Conference} on {Intelligent} {Robots} and {Systems}}}, 2011, pp. 4414--4419.

\bibitem[de~Lasa et~al.(2010)de~Lasa, Mordatch, and Hertzmann]{de_lasa_feature-based_2010}
M.~de~Lasa, I.~Mordatch, and A.~Hertzmann, ``Feature-based locomotion controllers,'' \emph{ACM Transactions On Graphics}, vol.~29, no.~4, pp. 131:1--131:10, 2010.

\bibitem[Pfeiffer et~al.(2018)Pfeiffer, Escande, and Kheddar]{pfeiffer_singularity_2018}
K.~Pfeiffer, A.~Escande, and A.~Kheddar, ``Singularity {Resolution} in {Equality} and {Inequality} {Constrained} {Hierarchical} {Task}-{Space} {Control} by {Adaptive} {Nonlinear} {Least} {Squares},'' \emph{IEEE Robotics and Automation Letters}, vol.~3, no.~4, pp. 3630--3637, 2018.

\bibitem[Bouyarmane and Kheddar(2018)]{bouyarmane2018tac}
K.~Bouyarmane and A.~Kheddar, ``On weight-prioritized multitask control of humanoid robots,'' \emph{IEEE Transactions on Automatic Control}, vol.~63, no.~6, pp. 1632--1647, 2018.

\bibitem[Djeha et~al.(2023)Djeha, Gergondet, and Kheddar]{djeha2023tro}
M.~Djeha, P.~Gergondet, and A.~Kheddar, ``Robust task-space quadratic programming for kinematic-controlled robots,'' \emph{IEEE Transactions on Robotics}, vol.~39, no.~5, pp. 3857--3874, 2023.

\bibitem[Feng et~al.(2015)Feng, Whitman, Xinjilefu, and Atkeson]{siyuan_JFR}
S.~Feng, E.~Whitman, X.~Xinjilefu, and C.~G. Atkeson, ``Optimization-based full body control for the darpa robotics challenge,'' \emph{Journal of Field Robotics}, vol.~32, no.~2, pp. 293--312, 2015.

\bibitem[Hirukawa et~al.(2006)Hirukawa, Hattori, Harada, Kajita, Kaneko, Kanehiro, Fujiwara, and Morisawa]{adios_ZMP}
H.~Hirukawa, S.~Hattori, K.~Harada, S.~Kajita, K.~Kaneko, F.~Kanehiro, K.~Fujiwara, and M.~Morisawa, ``A universal stability criterion of the foot contact of legged robots - adios zmp,'' in \emph{IEEE International Conference on Robotics and Automation}, 2006, pp. 1976--1983.

\bibitem[Righetti et~al.(2013)Righetti, Buchli, Mistry, Kalakrishnan, and Schaal]{force_dist_IDC_Ludovic}
L.~Righetti, J.~Buchli, M.~Mistry, M.~Kalakrishnan, and S.~Schaal, ``Optimal distribution of contact forces with inverse-dynamics control,'' \emph{The International Journal of Robotics Research}, vol.~32, no.~3, pp. 280--298, 2013.

\bibitem[Wieber()]{Nonholonomy_Wieber}
P.-B. Wieber, \emph{Holonomy and Nonholonomy in the Dynamics of Articulated Motion}.\hskip 1em plus 0.5em minus 0.4em\relax Springer Berlin Heidelberg, p. 411–425.

\bibitem[Laurenzi et~al.(2018)Laurenzi, Mingo~Hoffman, Parigi~Polverini, and Tsagarakis]{Post_Optimization_balance}
A.~Laurenzi, E.~Mingo~Hoffman, M.~Parigi~Polverini, and N.~G. Tsagarakis, ``Balancing control through post-optimization of contact forces,'' in \emph{IEEE-RAS International Conference on Humanoid Robots}, 2018, pp. 320--326.

\bibitem[Bouyarmane et~al.(2019{\natexlab{b}})Bouyarmane, Chappellet, Vaillant, and Kheddar]{QP_Multirobot_Kheddar}
K.~Bouyarmane, K.~Chappellet, J.~Vaillant, and A.~Kheddar, ``Quadratic programming for multirobot and task-space force control,'' \emph{IEEE Transactions on Robotics}, vol.~35, no.~1, pp. 64--77, 2019.

\bibitem[Ji et~al.(2024)Ji, Peng, Liu, Li, Yang, Cheng, and Wang]{ji2024exbody2}
M.~Ji, X.~Peng, F.~Liu, J.~Li, G.~Yang, X.~Cheng, and X.~Wang, ``Exbody2: Advanced expressive humanoid whole-body control,'' \emph{arXiv preprint arXiv:2412.13196}, 2024.

\bibitem[He et~al.(2024{\natexlab{b}})He, Luo, He, Xiao, Zhang, Zhang, Kitani, Liu, and Shi]{he2024omnih2o}
T.~He, Z.~Luo, X.~He, W.~Xiao, C.~Zhang, W.~Zhang, K.~M. Kitani, C.~Liu, and G.~Shi, ``Omnih2o: Universal and dexterous human-to-humanoid whole-body teleoperation and learning,'' in \emph{Annual Conference on Robot Learning}, 2024.

\bibitem[Fu et~al.(2024{\natexlab{b}})Fu, Zhao, Wu, Wetzstein, and Finn]{humanplus}
Z.~Fu, Q.~Zhao, Q.~Wu, G.~Wetzstein, and C.~Finn, ``Humanplus: Humanoid shadowing and imitation from humans,'' in \emph{Annual Conference on Robot Learning}, 2024.

\bibitem[Wang et~al.(2023{\natexlab{a}})Wang, Dehio, Tanguy, and Kheddar]{wang2023ijrr}
Y.~Wang, N.~Dehio, A.~Tanguy, and A.~Kheddar, ``Impact-aware task-space quadratic-programming control,'' \emph{The International Journal of Robotics Research}, vol.~42, no.~14, pp. 1265--1282, 2023.

\bibitem[Patel et~al.(2019)Patel, Shield, Kazi, Johnson, and Biegler]{patel2019contact}
A.~Patel, S.~L. Shield, S.~Kazi, A.~M. Johnson, and L.~T. Biegler, ``Contact-implicit trajectory optimization using orthogonal collocation,'' \emph{IEEE Robotics and Automation Letters}, vol.~4, no.~2, pp. 2242--2249, 2019.

\bibitem[Aydinoglu and Posa(2022)]{aydinoglu2022real}
A.~Aydinoglu and M.~Posa, ``Real-time multi-contact model predictive control via admm,'' in \emph{International Conference on Robotics and Automation}, 2022, pp. 3414--3421.

\bibitem[{Sharanya Puthige Venkatesh}(2024)]{Venkatesh_Penn_thesis}
{Sharanya Puthige Venkatesh}, ``Approximating global mpc for contact rich manipulation using local feedback,'' Master's thesis, University of Pennsylvania, 2024.

\bibitem[Plancher and Kuindersma(2020)]{plancher2020performance}
B.~Plancher and S.~Kuindersma, ``A performance analysis of parallel differential dynamic programming on a gpu,'' in \emph{the 13th Workshop on the Algorithmic Foundations of Robotics}.\hskip 1em plus 0.5em minus 0.4em\relax Springer, 2020, pp. 656--672.

\bibitem[Adabag et~al.(2024)Adabag, Atal, Gerard, and Plancher]{adabag2024mpcgpu}
E.~Adabag, M.~Atal, W.~Gerard, and B.~Plancher, ``Mpcgpu: Real-time nonlinear model predictive control through preconditioned conjugate gradient on the gpu,'' in \emph{IEEE International Conference on Robotics and Automation}, 2024, pp. 9787--9794.

\bibitem[Kong et~al.(2024)Kong, Payne, Zhu, and Johnson]{kong2024saltation}
N.~J. Kong, J.~J. Payne, J.~Zhu, and A.~M. Johnson, ``Saltation matrices: The essential tool for linearizing hybrid dynamical systems,'' \emph{Proceedings of the IEEE}, 2024.

\bibitem[Suh et~al.(2022)Suh, Pang, and Tedrake]{suh2022bundled}
H.~J.~T. Suh, T.~Pang, and R.~Tedrake, ``Bundled gradients through contact via randomized smoothing,'' \emph{IEEE Robotics and Automation Letters}, vol.~7, no.~2, pp. 4000--4007, 2022.

\bibitem[Hammoud et~al.(2022)Hammoud, Jordana, and Righetti]{hammoud2022irisc}
B.~Hammoud, A.~Jordana, and L.~Righetti, ``irisc: Iterative risk sensitive control for nonlinear systems with imperfect observations,'' in \emph{2022 American Control Conference (ACC)}.\hskip 1em plus 0.5em minus 0.4em\relax IEEE, 2022, pp. 3550--3557.

\bibitem[Chen et~al.(2024{\natexlab{a}})Chen, Li, Cheng, Hovakimyan, and Nguyen]{chen2024autotuning}
Q.~Chen, J.~Li, S.~Cheng, N.~Hovakimyan, and Q.~Nguyen, ``Autotuning bipedal locomotion mpc with grfm-net for efficient sim-to-real transfer,'' \emph{arXiv preprint arXiv:2409.15710}, 2024.

\bibitem[Sartore et~al.(2024)Sartore, Rando, Romualdi, Molinari, Rosasco, and Pucci]{sartore2024automatic}
C.~Sartore, M.~Rando, G.~Romualdi, C.~Molinari, L.~Rosasco, and D.~Pucci, ``Automatic gain tuning for humanoid robots walking architectures using gradient-free optimization techniques,'' in \emph{IEEE-RAS International Conference on Humanoid Robots}, 2024, pp. 996--1003.

\bibitem[Jiang et~al.(2024{\natexlab{a}})Jiang, He, Wang, Cheng, Sang, and Zhou]{Jiang2024skill_learning}
R.~Jiang, B.~He, Z.~Wang, X.~Cheng, H.~Sang, and Y.~Zhou, ``Robot skill learning and the data dilemma it faces: a systematic review,'' \emph{Robotic Intelligence and Automation}, vol.~44, no.~2, pp. 270--286, 2024.

\bibitem[Zhang et~al.(2024{\natexlab{b}})Zhang, Xiao, He, and Shi]{zhang2024wococo}
C.~Zhang, W.~Xiao, T.~He, and G.~Shi, ``Wococo: Learning whole-body humanoid control with sequential contacts,'' in \emph{Annual Conference on Robot Learning}, 2024.

\bibitem[Haarnoja et~al.(2024)Haarnoja, Moran, Lever, Huang, Tirumala, Humplik, Wulfmeier, Tunyasuvunakool, Siegel, Hafner, Bloesch, Hartikainen, Byravan, Hasenclever, Tassa, Sadeghi, Batchelor, Casarini, Saliceti, Game, Sreendra, Patel, Gwira, Huber, Hurley, Nori, Hadsell, and Heess]{SR_deepmind_soccer}
T.~Haarnoja \emph{et~al.}, ``Learning agile soccer skills for a bipedal robot with deep reinforcement learning,'' \emph{Science Robotics}, vol.~9, no.~89, 2024.

\bibitem[Huang et~al.(2024)Huang, Chi, Wang, Li, Peng, Shao, Nikolic, and Sreenath]{huang2024diffuseloco}
X.~Huang, Y.~Chi, R.~Wang, Z.~Li, X.~B. Peng, S.~Shao, B.~Nikolic, and K.~Sreenath, ``Diffuseloco: Real-time legged locomotion control with diffusion from offline datasets,'' in \emph{Annual Conference on Robot Learning}, 2024.

\bibitem[Schulman et~al.(2017)Schulman, Wolski, Dhariwal, Radford, and Klimov]{PPO}
J.~Schulman, F.~Wolski, P.~Dhariwal, A.~Radford, and O.~Klimov, ``Proximal policy optimization algorithms,'' \emph{arXiv preprint:1707.06347}, 2017.

\bibitem[Haarnoja et~al.(2018)Haarnoja, Zhou, Abbeel, and Levine]{SAC}
T.~Haarnoja, A.~Zhou, P.~Abbeel, and S.~Levine, ``Soft actor-critic: Off-policy maximum entropy deep reinforcement learning with a stochastic actor.'' in \emph{International Conference on Machine Learning}, vol.~80, 2018, pp. 1856--1865.

\bibitem[Sferrazza et~al.(2024)Sferrazza, Huang, Lin, Lee, and Abbeel]{sferrazza2024humanoidbench}
C.~Sferrazza, D.-M. Huang, X.~Lin, Y.~Lee, and P.~Abbeel, ``{HumanoidBench: Simulated Humanoid Benchmark for Whole-Body Locomotion and Manipulation},'' in \emph{Robotics: Science and Systems}, 2024.

\bibitem[Mnih et~al.(2016)Mnih, Badia, Mirza, Graves, Lillicrap, Harley, Silver, and Kavukcuoglu]{A3C}
V.~Mnih, A.~P. Badia, M.~Mirza, A.~Graves, T.~P. Lillicrap, T.~Harley, D.~Silver, and K.~Kavukcuoglu, ``Asynchronous methods for deep reinforcement learning,'' in \emph{International Conference on Machine Learning}, 2016.

\bibitem[Fujimoto et~al.(2018)Fujimoto, Hoof, and Meger]{TD3}
S.~Fujimoto, H.~Hoof, and D.~Meger, ``Addressing function approximation error in actor-critic methods,'' in \emph{International Conference on Machine Learning}, 2018, pp. 1582--1591.

\bibitem[Lillicrap et~al.(2015)Lillicrap, Hunt, Pritzel, Heess, Erez, Tassa, Silver, and Wierstra]{DDPG}
T.~P. Lillicrap, J.~J. Hunt, A.~Pritzel, N.~M.~O. Heess, T.~Erez, Y.~Tassa, D.~Silver, and D.~Wierstra, ``Continuous control with deep reinforcement learning,'' \emph{CoRR}, vol. abs/1509.02971, 2015.

\bibitem[Xie et~al.(2020)Xie, Clary, Dao, Morais, Hurst, and van~de Panne]{pmlr-v100-xie20a}
Z.~Xie, P.~Clary, J.~Dao, P.~Morais, J.~Hurst, and M.~van~de Panne, ``Learning locomotion skills for cassie: Iterative design and sim-to-real,'' in \emph{the Conference on Robot Learning}, vol. 100, 2020, pp. 317--329.

\bibitem[Schwarke et~al.(2023)Schwarke, Klemm, Boon, Bjelonic, and Hutter]{Curiosity-schwarke23a}
C.~Schwarke, V.~Klemm, M.~v.~d. Boon, M.~Bjelonic, and M.~Hutter, ``Curiosity-driven learning of joint locomotion and manipulation tasks,'' in \emph{Conference on Robot Learning}, vol. 229, 2023, pp. 2594--2610.

\bibitem[Kim et~al.(2024{\natexlab{a}})Kim, Oh, Lee, Choi, Ji, Jung, Youm, and Hwangbo]{reward_constraint_RL}
Y.~Kim, H.~Oh, J.~Lee, J.~Choi, G.~Ji, M.~Jung, D.~Youm, and J.~Hwangbo, ``Not only rewards but also constraints: Applications on legged robot locomotion,'' \emph{IEEE Transactions on Robotics}, vol.~40, pp. 2984--3003, 2024.

\bibitem[Lee et~al.(2020)Lee, Hwangbo, Wellhausen, Koltun, and Hutter]{Hutter_SR2020}
J.~Lee, J.~Hwangbo, L.~Wellhausen, V.~Koltun, and M.~Hutter, ``Learning quadrupedal locomotion over challenging terrain,'' \emph{Science Robotics}, vol.~5, no.~47, 2020.

\bibitem[Tan et~al.(2016)Tan, Xie, Boots, and Liu]{Tan_systemID}
J.~Tan, Z.~Xie, B.~Boots, and C.~K. Liu, ``Simulation-based design of dynamic controllers for humanoid balancing,'' in \emph{IEEE/RSJ International Conference on Intelligent Robots and Systems}, 2016, pp. 2729--2736.

\bibitem[Li et~al.(2024{\natexlab{d}})Li, Li, Fu, and Wu]{li2024learningagilebipedalmotions}
Y.~Li, J.~Li, W.~Fu, and Y.~Wu, ``Learning agile bipedal motions on a quadrupedal robot,'' in \emph{IEEE International Conference on Robotics and Automation}, 2024, pp. 9735--9742.

\bibitem[He et~al.(2025)He, Gao, Xiao, Zhang, Wang, Wang, Luo, He, Sobanbab, Pan, Yi, Qu, Kitani, Hodgins, Fan, Zhu, Liu, and Shi]{asap}
T.~He \emph{et~al.}, ``Asap: Aligning simulation and real-world physics for learning agile humanoid whole-body skills,'' \emph{arXiv preprint arXiv:2502.01143}, 2025.

\bibitem[Hsu et~al.(2022)Hsu, Ren, Nguyen, Majumdar, and Fisac]{hsuren2022slr}
K.-C. Hsu, A.~Z. Ren, D.~P. Nguyen, A.~Majumdar, and J.~F. Fisac, ``Sim-to-lab-to-real: Safe reinforcement learning with shielding and generalization guarantees,'' \emph{Artificial Intelligence}, p. 103811, 2022.

\bibitem[Brunke et~al.(2022)Brunke, Greeff, Hall, Yuan, Zhou, Panerati, and Schoellig]{SafeRLRoboticsSurvey}
L.~Brunke, M.~Greeff, A.~W. Hall, Z.~Yuan, S.~Zhou, J.~Panerati, and A.~P. Schoellig, ``Safe learning in robotics: From learning-based control to safe reinforcement learning,'' \emph{Annual Review of Control, Robotics, and Autonomous Systems}, vol.~5, no. Volume 5, 2022, pp. 411--444, 2022.

\bibitem[Ibarz et~al.(2021)Ibarz, Tan, Finn, Kalakrishnan, Pastor, and Levine]{RL_Robot_Levine}
J.~Ibarz, J.~Tan, C.~Finn, M.~Kalakrishnan, P.~Pastor, and S.~Levine, ``How to train your robot with deep reinforcement learning: lessons we have learned,'' \emph{The International Journal of Robotics Research}, vol.~40, no. 4-5, pp. 698--721, 2021.

\bibitem[Wu et~al.(2024{\natexlab{a}})Wu, Gu, Wu, Wu, and Zhao]{wu2023infer}
F.~Wu, Z.~Gu, H.~Wu, A.~Wu, and Y.~Zhao, ``Infer and adapt: Bipedal locomotion reward learning from demonstrations via inverse reinforcement learning,'' in \emph{IEEE International Conference on Robotics and Automation}, 2024, pp. 16\,243--16\,250.

\bibitem[Cheng et~al.(2024{\natexlab{b}})Cheng, Li, Yang, Yang, and Wang]{cheng2024tv}
X.~Cheng, J.~Li, S.~Yang, G.~Yang, and X.~Wang, ``Open-television: Teleoperation with immersive active visual feedback,'' in \emph{Annual Conference on Robot Learning}, 2024.

\bibitem[Wang et~al.(2023{\natexlab{b}})Wang, Lin, Zeng, Luo, Zhang, and Zhang]{physhoi}
Y.~Wang, J.~Lin, A.~Zeng, Z.~Luo, J.~Zhang, and L.~Zhang, ``Physhoi: Physics-based imitation of dynamic human-object interaction,'' \emph{arXiv preprint arXiv:2312.04393}, 2023.

\bibitem[Merel et~al.(2017)Merel, Tassa, TB, Srinivasan, Lemmon, Wang, Wayne, and Heess]{gail_stand}
J.~Merel, Y.~Tassa, D.~TB, S.~Srinivasan, J.~Lemmon, Z.~Wang, G.~Wayne, and N.~Heess, ``Learning human behaviors from motion capture by adversarial imitation,'' \emph{arXiv preprint arXiv:1707.02201}, 2017.

\bibitem[Tang et~al.(2024)Tang, Hiraoka, Hiraoka, Shi, Kawaharazuka, Kojima, Okada, and Inaba]{tang2024humanmimic}
A.~Tang, T.~Hiraoka, N.~Hiraoka, F.~Shi, K.~Kawaharazuka, K.~Kojima, K.~Okada, and M.~Inaba, ``Humanmimic: Learning natural locomotion and transitions for humanoid robot via wasserstein adversarial imitation,'' in \emph{IEEE International Conference on Robotics and Automation}, 2024, pp. 13\,107--13\,114.

\bibitem[Zhang et~al.(2024{\natexlab{c}})Zhang, Cui, Yan, Sun, Duan, Han, Zhao, Zhang, Guo, Zhang, and Xu]{zhang2024wholebody}
Q.~Zhang \emph{et~al.}, ``Whole-body humanoid robot locomotion with human reference,'' in \emph{IEEE/RSJ International Conference on Intelligent Robots and Systems}, 2024, pp. 11\,225--11\,231.

\bibitem[Peng et~al.(2018{\natexlab{a}})Peng, Kanazawa, Malik, Abbeel, and Levine]{peng2018sfv}
X.~B. Peng, A.~Kanazawa, J.~Malik, P.~Abbeel, and S.~Levine, ``Sfv: Reinforcement learning of physical skills from videos,'' \emph{ACM Transactions On Graphics}, vol.~37, no.~6, pp. 1--14, 2018.

\bibitem[Zhang et~al.(2019)Zhang, Liu, and Zhou]{gail_locomotion}
H.~Zhang, Y.~Liu, and W.~Zhou, ``Deep adversarial imitation learning of locomotion skills from one-shot video demonstration,'' in \emph{IEEE Annual International Conference on CYBER Technology in Automation, Control, and Intelligent Systems}, 2019, pp. 1257--1261.

\bibitem[Li et~al.(2024{\natexlab{e}})Li, Zhu, Xie, Jiang, Seo, Pavlakos, and Zhu]{okami2024}
J.~Li, Y.~Zhu, Y.~Xie, Z.~Jiang, M.~Seo, G.~Pavlakos, and Y.~Zhu, ``Okami: Teaching humanoid robots manipulation skills through single video imitation,'' in \emph{Annual Conference on Robot Learning}, 2024.

\bibitem[Cisneros-Lim{\'o}n et~al.(2024)Cisneros-Lim{\'o}n, Dallard, Benallegue, Kaneko, Kaminaga, Gergondet, Tanguy, Singh, Sun, Chen, Fournier, Lorthioir, Tsuru, Chefchaouni-Moussaoui, Osawa, Caron, Chappellet, Morisawa, Escande, Ayusawa, Houhou, Kumagai, Ono, Shirasaka, Wada, Wada, Kanehiro, and Kheddar]{cisnero2024soro}
R.~Cisneros-Lim{\'o}n \emph{et~al.}, ``A cybernetic avatar system to embody human telepresence for connectivity, exploration, and skill transfer,'' \emph{International Journal of Social Robotics}, 2024.

\bibitem[Lu et~al.(2024)Lu, Cheng, Li, Yang, Ji, Yuan, Yang, Yi, and Wang]{lu2024mobiletelevision}
C.~Lu, X.~Cheng, J.~Li, S.~Yang, M.~Ji, C.~Yuan, G.~Yang, S.~Yi, and X.~Wang, ``Mobile-television: Predictive motion priors for humanoid whole-body control,'' \emph{arXiv preprint arXiv:2412.07773}, 2024.

\bibitem[Dallard et~al.(2023)Dallard, Benallegue, Kanehiro, and Kheddar]{dallard2023ral}
A.~Dallard, M.~Benallegue, F.~Kanehiro, and A.~Kheddar, ``Synchronized human-humanoid motion imitation,'' \emph{IEEE Robotics and Automation Letters}, vol.~8, no.~7, pp. 4155--4162, 2023.

\bibitem[Xu et~al.(2024)Xu, Zhang, Li, Han, and Lu]{xu2024humanvla}
X.~Xu, Y.~Zhang, Y.-L. Li, L.~Han, and C.~Lu, ``Human{VLA}: Towards vision-language directed object rearrangement by physical humanoid,'' in \emph{Annual Conference on Neural Information Processing Systems}, 2024.

\bibitem[Juravsky et~al.(2024)Juravsky, Guo, Fidler, and Peng]{SuperPADL}
J.~Juravsky, Y.~Guo, S.~Fidler, and X.~B. Peng, ``Superpadl: Scaling language-directed physics-based control with progressive supervised distillation,'' in \emph{Special Interest Group on Computer Graphics and Interactive Techniques Conference Conference Papers ’24}.\hskip 1em plus 0.5em minus 0.4em\relax ACM, Jul. 2024, p. 1–11.

\bibitem[Goodrich et~al.(2013)Goodrich, Crandall, and Barakova]{teleop_humanoid_review}
M.~A. Goodrich, J.~W. Crandall, and E.~Barakova, ``Teleoperation and beyond for assistive humanoid robots,'' \emph{Reviews of Human Factors and Ergonomics}, vol.~9, no.~1, pp. 175--226, 2013.

\bibitem[Bhatnagar et~al.(2022)Bhatnagar, Xie, Petrov, Sminchisescu, Theobalt, and Pons-Moll]{behave}
B.~L. Bhatnagar, X.~Xie, I.~A. Petrov, C.~Sminchisescu, C.~Theobalt, and G.~Pons-Moll, ``Behave: Dataset and method for tracking human object interactions,'' in \emph{Proceedings of the IEEE/CVF Conference on Computer Vision and Pattern Recognition}, 2022, pp. 15\,935--15\,946.

\bibitem[{Carnegie Mellon University}(2002)]{CMU_dataset}
{Carnegie Mellon University}, ``Cmu graphics lab motion capture database,'' 2002.

\bibitem[{Simon Fraser University} and {National University of Singapore}()]{SFU_dataset}
{Simon Fraser University} and {National University of Singapore}, ``Sfu motion capture database.''

\bibitem[Harvey et~al.(2020)Harvey, Yurick, Nowrouzezahrai, and Pal]{LAFAAN_dataset}
F.~G. Harvey, M.~Yurick, D.~Nowrouzezahrai, and C.~Pal, ``Robust motion in-betweening,'' vol.~39, no.~4, 2020.

\bibitem[Mahmood et~al.(2019)Mahmood, Ghorbani, Troje, Pons-Moll, and Black]{AMASS}
N.~Mahmood, N.~Ghorbani, N.~F. Troje, G.~Pons-Moll, and M.~J. Black, ``{AMASS}: Archive of motion capture as surface shapes,'' in \emph{International Conference on Computer Vision}, 2019, pp. 5442--5451.

\bibitem[Varin(2021)]{VarinThesis}
P.~Varin, ``\BIBforeignlanguage{English}{Estimation and planning for dynamic robot behaviors},'' Ph.D. dissertation, Harvard University, 2021.

\bibitem[OpenAI(2024)]{sora}
OpenAI, ``Video generation models as world simulators,'' 2024.

\bibitem[Blattmann et~al.(2023)Blattmann, Dockhorn, Kulal, Mendelevitch, Kilian, Lorenz, Levi, English, Voleti, Letts, et~al.]{blattmann2023stable}
A.~Blattmann \emph{et~al.}, ``Stable video diffusion: Scaling latent video diffusion models to large datasets,'' \emph{arXiv preprint:2311.15127}, 2023.

\bibitem[Zhu et~al.(2024)Zhu, Ma, Ro, Ci, Zhang, Shi, Gao, Tian, and Wang]{Motion_Generation_Survey}
W.~Zhu, X.~Ma, D.~Ro, H.~Ci, J.~Zhang, J.~Shi, F.~Gao, Q.~Tian, and Y.~Wang, ``Human motion generation: A survey,'' \emph{IEEE Transactions on Pattern Analysis and Machine Intelligence}, vol.~46, no.~4, pp. 2430--2449, 2024.

\bibitem[Cheng et~al.(2024{\natexlab{c}})Cheng, Ji, Chen, Yang, Yang, and Wang]{cheng2024expressive}
X.~Cheng, Y.~Ji, J.~Chen, R.~Yang, G.~Yang, and X.~Wang, ``Expressive whole-body control for humanoid robots,'' 2024.

\bibitem[Fabisch et~al.(2022)Fabisch, Uliano, Marschner, Laux, Brust, and Controzzi]{fabisch_modular_2022}
A.~Fabisch, M.~Uliano, D.~Marschner, M.~Laux, J.~Brust, and M.~Controzzi, ``A {Modular} {Approach} to the {Embodiment} of {Hand} {Motions} from {Human} {Demonstrations},'' in \emph{{IEEE}-{RAS} {International} {Conference} on {Humanoid} {Robots}}, 2022, pp. 801--808, iSSN: 2164-0580.

\bibitem[Peng et~al.(2021)Peng, Ma, Abbeel, Levine, and Kanazawa]{AMP}
X.~B. Peng, Z.~Ma, P.~Abbeel, S.~Levine, and A.~Kanazawa, ``Amp: adversarial motion priors for stylized physics-based character control,'' \emph{ACM Transactions On Graphics}, vol.~40, no.~4, 2021.

\bibitem[Dugar et~al.(2024)Dugar, Shrestha, Yu, van Marum, and Fern]{dugar2024DigitMHC}
P.~Dugar, A.~Shrestha, F.~Yu, B.~van Marum, and A.~Fern, ``Learning multi-modal whole-body control for real-world humanoid robots,'' \emph{arXiv preprint arXiv:2408.07295}, 2024.

\bibitem[Peng et~al.(2018{\natexlab{b}})Peng, Abbeel, Levine, and van~de Panne]{DeepMimic2018}
X.~B. Peng, P.~Abbeel, S.~Levine, and M.~van~de Panne, ``Deepmimic: example-guided deep reinforcement learning of physics-based character skills,'' \emph{ACM Transactions On Graphics}, vol.~37, no.~4, 2018.

\bibitem[Yu et~al.(2021)Yu, Park, and Lee]{yu2021human}
R.~Yu, H.~Park, and J.~Lee, ``Human dynamics from monocular video with dynamic camera movements,'' \emph{ACM Transactions on Graphics}, vol.~40, no.~6, pp. 1--14, 2021.

\bibitem[Xie et~al.(2023)Xie, Tseng, Starke, van~de Panne, and Liu]{xie_box_locomanip}
Z.~Xie, J.~Tseng, S.~Starke, M.~van~de Panne, and C.~K. Liu, ``Hierarchical planning and control for box loco-manipulation,'' \emph{Proc. ACM Comput. Graph. Interact. Tech.}, vol.~6, no.~3, 2023.

\bibitem[Hassan et~al.(2023)Hassan, Guo, Wang, Black, Fidler, and Peng]{hassan-box_carry}
M.~Hassan, Y.~Guo, T.~Wang, M.~Black, S.~Fidler, and X.~B. Peng, ``Synthesizing physical character-scene interactions,'' in \emph{ACM SIGGRAPH 2023 Conference Proceedings}, 2023, pp. 1--9.

\bibitem[Wu et~al.(2024{\natexlab{b}})Wu, Gu, Zhao, and Wu]{wu2024learn}
F.~Wu, Z.~Gu, Y.~Zhao, and A.~Wu, ``Learn to teach: Improve sample efficiency in teacher-student learning for sim-to-real transfer,'' \emph{arXiv preprint arXiv:2402.06783}, 2024.

\bibitem[Youm et~al.(2023)Youm, Jung, Kim, Hwangbo, Park, and Ha]{IFM_quad}
D.~Youm, H.~Jung, H.~Kim, J.~Hwangbo, H.-W. Park, and S.~Ha, ``Imitating and finetuning model predictive control for robust and symmetric quadrupedal locomotion,'' \emph{IEEE Robotics and Automation Letters}, vol.~8, no.~11, pp. 7799--7806, 2023.

\bibitem[Smith et~al.(2023)Smith, Kew, Li, Luu, Peng, Ha, Tan, and Levine]{Smith2023LearningAA}
L.~M. Smith, J.~C. Kew, T.~Li, L.~Luu, X.~B. Peng, S.~Ha, J.~Tan, and S.~Levine, ``{Learning and Adapting Agile Locomotion Skills by Transferring Experience},'' in \emph{Robotics: Science and Systems}, 2023.

\bibitem[Jenelten et~al.(2024)Jenelten, He, Farshidian, and Hutter]{jenelten2024dtc}
F.~Jenelten, J.~He, F.~Farshidian, and M.~Hutter, ``Dtc: Deep tracking control,'' \emph{Science Robotics}, vol.~9, no.~86, 2024.

\bibitem[Kang et~al.(2023)Kang, Cheng, Zamora, Zargarbashi, and Coros]{RL_MPC_RAL2023}
D.~Kang, J.~Cheng, M.~Zamora, F.~Zargarbashi, and S.~Coros, ``Rl + model-based control: Using on-demand optimal control to learn versatile legged locomotion,'' \emph{IEEE Robotics and Automation Letters}, vol.~8, no.~10, pp. 6619--6626, 2023.

\bibitem[Levine and Koltun(2013)]{GPS}
S.~Levine and V.~Koltun, ``Guided policy search,'' in \emph{Proceedings of the International Conference on Machine Learning}, vol.~28, no.~3, 2013, pp. 1--9.

\bibitem[Marew et~al.(2024)Marew, Perera, Yu, Roelker, and Kim]{marew2024soccerkicking}
D.~Marew, N.~Perera, S.~Yu, S.~Roelker, and D.~Kim, ``A biomechanics-inspired approach to soccer kicking for humanoid robots,'' in \emph{IEEE-RAS International Conference on Humanoid Robots}, 2024, pp. 722--729.

\bibitem[Kwon et~al.(2020)Kwon, Lee, and Van De~Panne]{LearnCDM2020}
T.~Kwon, Y.~Lee, and M.~Van De~Panne, ``Fast and flexible multilegged locomotion using learned centroidal dynamics,'' \emph{ACM Transactions On Graphics}, vol.~39, no.~4, 2020.

\bibitem[Ijspeert et~al.(2002)Ijspeert, Nakanishi, and Schaal]{Ijspeert_2002}
A.~J. Ijspeert, J.~Nakanishi, and S.~Schaal, ``Learning attractor landscapes for learning motor primitives,'' in \emph{Proceedings of the International Conference on Neural Information Processing Systems}, 2002, p. 1547–1554.

\bibitem[Xie et~al.(2018)Xie, Berseth, Clary, Hurst, and van~de Panne]{Xie_2018_Cassie}
Z.~Xie, G.~Berseth, P.~Clary, J.~Hurst, and M.~van~de Panne, ``Feedback control for cassie with deep reinforcement learning,'' in \emph{IEEE/RSJ International Conference on Intelligent Robots and Systems}, 2018, pp. 1241--1246.

\bibitem[singh et~al.(2023)singh, Xie, Gergondet, and Kanehiro]{singh2023Access}
R.~P. singh, Z.~Xie, P.~Gergondet, and F.~Kanehiro, ``Learning bipedal walking for humanoids with current feedback,'' \emph{IEEE Access}, vol.~11, p. 82013–82023, 2023.

\bibitem[Fuchioka et~al.(2023)Fuchioka, Xie, and Van~de Panne]{OPT-Mimic}
Y.~Fuchioka, Z.~Xie, and M.~Van~de Panne, ``Opt-mimic: Imitation of optimized trajectories for dynamic quadruped behaviors,'' in \emph{International Conference on Robotics and Automation}, 2023, pp. 5092--5098.

\bibitem[Liu et~al.(2024{\natexlab{b}})Liu, Gu, Cai, Zhou, Zhao, Jung, Ha, Chen, Xu, and Zhao]{liu2024opt2skill}
F.~Liu, Z.~Gu, Y.~Cai, Z.~Zhou, S.~Zhao, H.~Jung, S.~Ha, Y.~Chen, D.~Xu, and Y.~Zhao, ``Opt2skill: Imitating dynamically-feasible whole-body trajectories for versatile humanoid loco-manipulation,'' \emph{arXiv preprint arXiv:2409.20514}, 2024.

\bibitem[Paredes and Hereid(2022)]{VictorIROS2022NeuralAdaptation}
V.~C. Paredes and A.~Hereid, ``Resolved motion control for 3d underactuated bipedal walking using linear inverted pendulum dynamics and neural adaptation,'' in \emph{IEEE/RSJ International Conference on Intelligent Robots and Systems}, 2022, pp. 6761--6767.

\bibitem[Coros et~al.(2009)Coros, Beaudoin, and van~de Panne]{Coros09}
S.~Coros, P.~Beaudoin, and M.~van~de Panne, ``Robust task-based control policies for physics-based characters,'' \emph{ACM Transactions on Graphics}, vol.~28, no.~5, p. Article 170, 2009.

\bibitem[Li et~al.(2024{\natexlab{f}})Li, Stanger-Jones, Heim, and bae Kim]{li2024fld}
C.~Li, E.~Stanger-Jones, S.~Heim, and S.~bae Kim, ``{FLD}: Fourier latent dynamics for structured motion representation and learning,'' in \emph{The International Conference on Learning Representations}, 2024.

\bibitem[Kingma and Welling(2014)]{VAE}
D.~P. Kingma and M.~Welling, ``Auto-encoding variational bayes.'' in \emph{International Conference on Learning Representations}, 2014.

\bibitem[Goodfellow et~al.(2020)Goodfellow, Pouget-Abadie, Mirza, Xu, Warde-Farley, Ozair, Courville, and Bengio]{GAN_Goodfellow}
I.~Goodfellow, J.~Pouget-Abadie, M.~Mirza, B.~Xu, D.~Warde-Farley, S.~Ozair, A.~Courville, and Y.~Bengio, ``Generative adversarial networks,'' \emph{Communications of the ACM}, vol.~63, no.~11, p. 139–144, 2020.

\bibitem[Bohez et~al.(2022)Bohez, Tunyasuvunakool, Brakel, Sadeghi, Hasenclever, Tassa, Parisotto, Humplik, Haarnoja, Hafner, et~al.]{bohez2022imitaterepurposelearningreusable}
S.~Bohez \emph{et~al.}, ``Imitate and repurpose: Learning reusable robot movement skills from human and animal behaviors,'' \emph{arXiv preprint arXiv:2203.17138}, 2022.

\bibitem[Tessler et~al.(2023)Tessler, Kasten, Guo, Mannor, Chechik, and Peng]{Tessler2023CALM}
C.~Tessler, Y.~Kasten, Y.~Guo, S.~Mannor, G.~Chechik, and X.~B. Peng, ``Calm: Conditional adversarial latent models for directable virtual characters,'' in \emph{ACM SIGGRAPH 2023 Conference Proceedings}, 2023, pp. 1--9.

\bibitem[Luo et~al.(2024{\natexlab{a}})Luo, Cao, Merel, Winkler, Huang, Kitani, and Xu]{luo2024universal}
Z.~Luo, J.~Cao, J.~Merel, A.~Winkler, J.~Huang, K.~M. Kitani, and W.~Xu, ``Universal humanoid motion representations for physics-based control,'' in \emph{The International Conference on Learning Representations}, 2024.

\bibitem[Liu et~al.(2022)Liu, Zhu, and Zhang]{ijcai2022GCRL_survey}
M.~Liu, M.~Zhu, and W.~Zhang, ``Goal-conditioned reinforcement learning: Problems and solutions,'' in \emph{Proceedings of the International Joint Conference on Artificial Intelligence}, 2022, pp. 5502--5511.

\bibitem[Serifi et~al.(2024)Serifi, Grandia, Knoop, Gross, and B{\"a}cher]{VMP}
A.~Serifi, R.~Grandia, E.~Knoop, M.~Gross, and M.~B{\"a}cher, ``Vmp: Versatile motion priors for robustly tracking motion on physical characters,'' in \emph{Computer Graphics Forum}.\hskip 1em plus 0.5em minus 0.4em\relax Wiley Online Library, 2024, p. e15175.

\bibitem[Ha and Schmidhuber(2018)]{world_model_Ha}
D.~Ha and J.~Schmidhuber, ``World models,'' \emph{Zenodo}, 2018.

\bibitem[Hester et~al.(2010)Hester, Quinlan, and Stone]{Hester2010}
T.~Hester, M.~Quinlan, and P.~Stone, ``Generalized model learning for reinforcement learning on a humanoid robot,'' in \emph{IEEE International Conference on Robotics and Automation}, 2010, pp. 2369--2374.

\bibitem[Gu et~al.(2024)Gu, Wang, Zhu, Shi, Guo, Liu, and Chen]{Gu-RSS-24}
X.~Gu, Y.-J. Wang, X.~Zhu, C.~Shi, Y.~Guo, Y.~Liu, and J.~Chen, ``{Advancing Humanoid Locomotion: Mastering Challenging Terrains with Denoising World Model Learning},'' in \emph{Robotics: Science and Systems}, 2024.

\bibitem[Hansen et~al.(2025)Hansen, V, Sobal, LeCun, Wang, and Su]{hansen2024worldmodelsvisual}
N.~Hansen, J.~S. V, V.~Sobal, Y.~LeCun, X.~Wang, and H.~Su, ``Hierarchical world models as visual whole-body humanoid controllers,'' in \emph{International Conference on Learning Representations}, 2025.

\bibitem[Shi et~al.(2024)Shi, Li, Zhu, Sheng, Han, and Meng]{shi2024efficientmodelbasedapproachlearning}
H.~Shi, T.~Li, Q.~Zhu, J.~Sheng, L.~Han, and M.~Q.-H. Meng, ``An efficient model-based approach on learning agile motor skills without reinforcement,'' in \emph{IEEE International Conference on Robotics and Automation}, 2024, pp. 5724--5730.

\bibitem[Gao et~al.(2024{\natexlab{b}})Gao, Wang, Xiao, Wang, Wang, Cao, Hu, Liu, Dai, and Pang]{gao2024coohoi}
J.~Gao, Z.~Wang, Z.~Xiao, J.~Wang, T.~Wang, J.~Cao, X.~Hu, S.~Liu, J.~Dai, and J.~Pang, ``Coo{HOI}: Learning cooperative human-object interaction with manipulated object dynamics,'' in \emph{The Annual Conference on Neural Information Processing Systems}, 2024.

\bibitem[Dao et~al.(2024)Dao, Duan, and Fern]{dao2023simtoreal}
J.~Dao, H.~Duan, and A.~Fern, ``Sim-to-real learning for humanoid box loco-manipulation,'' in \emph{IEEE International Conference on Robotics and Automation}, 2024, pp. 16\,930--16\,936.

\bibitem[Munn et~al.(2024)Munn, Tidd, Howard, and Gallagher]{munn2024dynamic_throwing}
H.~Munn, B.~Tidd, D.~Howard, and M.~Gallagher, ``Whole-body dynamic throwing with legged manipulators,'' \emph{arXiv preprint arXiv:2410.05681}, 2024.

\bibitem[Yang et~al.(2023)Yang, Pu, Xin, Zhang, and Li]{fall_recovery}
C.~Yang, C.~Pu, G.~Xin, J.~Zhang, and Z.~Li, ``Learning complex motor skills for legged robot fall recovery,'' \emph{IEEE Robotics and Automation Letters}, vol.~8, no.~7, pp. 4307--4314, 2023.

\bibitem[Al-Hafez et~al.(2023)Al-Hafez, Zhao, Peters, and Tateo]{LocoMuJoCo}
F.~Al-Hafez, G.~Zhao, J.~Peters, and D.~Tateo, ``Locomujoco: A comprehensive imitation learning benchmark for locomotion,'' in \emph{Robot Learning Workshop, NeurIPS}, 2023.

\bibitem[Liu et~al.(2023)Liu, Zhu, Gao, Feng, Liu, Zhu, and Stone]{liu2023libero}
B.~Liu, Y.~Zhu, C.~Gao, Y.~Feng, Q.~Liu, Y.~Zhu, and P.~Stone, ``Libero: Benchmarking knowledge transfer for lifelong robot learning,'' \emph{NeurIPS Datasets and Benchmarks Track}, vol.~36, 2023.

\bibitem[Liu et~al.(2024{\natexlab{c}})Liu, Yang, Zhong, Wang, and Yi]{liu2024mimickingbench}
Y.~Liu, B.~Yang, L.~Zhong, H.~Wang, and L.~Yi, ``Mimicking-bench: A benchmark for generalizable humanoid-scene interaction learning via human mimicking,'' \emph{arXiv preprint arXiv:2412.17730}, 2024.

\bibitem[Luo et~al.(2024{\natexlab{b}})Luo, Wang, Liu, Zhang, Tessler, Wang, Yuan, Cao, Lin, Wang, et~al.]{luo2024SMPLOlympics}
Z.~Luo \emph{et~al.}, ``Smplolympics: Sports environments for physically simulated humanoids,'' \emph{arXiv preprint arXiv:2407.00187}, 2024.

\bibitem[Liao et~al.(2024)Liao, Zhang, Huang, Huang, Li, and Sreenath]{liao2024berkeleyhumanoid}
Q.~Liao, B.~Zhang, X.~Huang, X.~Huang, Z.~Li, and K.~Sreenath, ``Berkeley humanoid: A research platform for learning-based control,'' \emph{arXiv preprint arXiv:2407.21781}, 2024.

\bibitem[Qiao(2024)]{Q_Humanoid}
H.~Qiao, ``The q-series humanoid robot system developed by the institute of automation, chinese academy of sciences,'' 2024.

\bibitem[Wilson et~al.(2020)Wilson, Glotfelter, Wang, Mayya, Notomista, Mote, and Egerstedt]{robotarium}
S.~Wilson, P.~Glotfelter, L.~Wang, S.~Mayya, G.~Notomista, M.~Mote, and M.~Egerstedt, ``The robotarium: Globally impactful opportunities, challenges, and lessons learned in remote-access, distributed control of multirobot systems,'' \emph{IEEE Control Systems Magazine}, vol.~40, no.~1, pp. 26--44, 2020.

\bibitem[Wuthrich et~al.(2021)Wuthrich, Widmaier, Grimminger, Joshi, Agrawal, Hammoud, Khadiv, Bogdanovic, Berenz, Viereck, Naveau, Righetti, Sch\"{o}lkopf, and Bauer]{TriFinger}
M.~Wuthrich \emph{et~al.}, ``Trifinger: An open-source robot for learning dexterity,'' in \emph{Conference on Robot Learning}, vol. 155, 16--18 Nov 2021, pp. 1871--1882.

\bibitem[Cheng et~al.(2019{\natexlab{b}})Cheng, Ramirez-Amaro, Beetz, and Kuniyoshi]{Purposive_learning_Cheng2019}
G.~Cheng, K.~Ramirez-Amaro, M.~Beetz, and Y.~Kuniyoshi, ``Purposive learning: Robot reasoning about the meanings of human activities,'' \emph{Science Robotics}, vol.~4, no.~26, p. eaav1530, 2019.

\bibitem[Radford et~al.(2021)Radford, Kim, Hallacy, Ramesh, Goh, Agarwal, Sastry, Askell, Mishkin, Clark, et~al.]{radford2021learning}
A.~Radford \emph{et~al.}, ``Learning transferable visual models from natural language supervision,'' \emph{Proceedings of the 38th International Conference on Machine Learning}, vol. 139, pp. 8748--8763, 2021.

\bibitem[Alayrac et~al.(2022)Alayrac, Donahue, Luc, Miech, Barr, Hasson, Lenc, Mensch, Millican, Reynolds, et~al.]{alayrac2022flamingo}
J.-B. Alayrac \emph{et~al.}, ``Flamingo: a visual language model for few-shot learning,'' \emph{Advances in Neural Information Processing Systems}, vol.~35, pp. 23\,716--23\,736, 2022.

\bibitem[Mao et~al.(2024)Mao, Zhao, Song, Shi, Ye, Zhang, Geng, Malik, Guizilini, and Wang]{mao2024learning}
J.~Mao, S.~Zhao, S.~Song, T.~Shi, J.~Ye, M.~Zhang, H.~Geng, J.~Malik, V.~Guizilini, and Y.~Wang, ``Learning from massive human videos for universal humanoid pose control,'' \emph{arXiv preprint:2412.14172}, 2024.

\bibitem[Ho et~al.(2022)Ho, Chan, Saharia, Whang, Gao, Gritsenko, Kingma, Poole, Norouzi, Fleet, et~al.]{ho2022imagen}
J.~Ho \emph{et~al.}, ``Imagen video: High definition video generation with diffusion models,'' \emph{arXiv preprint arXiv:2210.02303}, 2022.

\bibitem[NVIDIA et~al.(2025)NVIDIA, Bjorck, Castañeda, Cherniadev, Da, Ding, Fan, Fang, Fox, Hu, Huang, Jang, Jiang, Kautz, Kundalia, Lao, Li, Lin, Lin, Liu, Llontop, Magne, Mandlekar, Narayan, Nasiriany, Reed, Tan, Wang, Wang, Wang, Wang, Xiang, Xie, Xu, Xu, Ye, Yu, Zhang, Zhang, Zhao, Zheng, and Zhu]{gr00tn1}
NVIDIA \emph{et~al.}, ``Gr00t n1: An open foundation model for generalist humanoid robots,'' \emph{arXiv preprint arXiv:2503.14734}, 2025.

\bibitem[Qiu et~al.(2025)Qiu, Yang, Cheng, Chawla, Li, He, Yan, Yoon, Hoque, Paulsen, Yang, Zhang, Yi, Shi, and Wang]{humanoidpolicyhuman}
R.-Z. Qiu \emph{et~al.}, ``Humanoid policy ~ human policy,'' \emph{arXiv preprint arXiv:2503.13441}, 2025.

\bibitem[DelPreto et~al.(2022)DelPreto, Liu, Luo, Foshey, Li, Torralba, Matusik, and Rus]{NEURIPS2022_ActionSense}
J.~DelPreto, C.~Liu, Y.~Luo, M.~Foshey, Y.~Li, A.~Torralba, W.~Matusik, and D.~Rus, ``Actionsense: A multimodal dataset and recording framework for human activities using wearable sensors in a kitchen environment,'' in \emph{Advances in Neural Information Processing Systems}, vol.~35, 2022, pp. 13\,800--13\,813.

\bibitem[Grauman et~al.(2023)Grauman, Westbury, Torresani, Kitani, Malik, Afouras, Ashutosh, Baiyya, Bansal, Boote, Byrne, Chavis, Chen, Cheng, Chu, Crane, Dasgupta, Dong, Escobar, Forigua, Gebreselasie, Haresh, Huang, Islam, Jain, Khirodkar, Kukreja, Liang, Liu, Majumder, Mao, Martin, Mavroudi, Nagarajan, Ragusa, Ramakrishnan, Seminara, Somayazulu, Song, Su, Xue, Zhang, Zhang, Castillo, Chen, Fu, Furuta, Gonzalez, Gupta, Hu, Huang, Huang, Khoo, Kumar, Kuo, Lakhavani, Liu, Luo, Luo, Meredith, Miller, Oguntola, Pan, Peng, Pramanick, Ramazanova, Ryan, Shan, Somasundaram, Song, Southerland, Tateno, Wang, Wang, Yagi, Yan, Yang, Yu, Zha, Zhao, Zhao, Zhu, Zhuo, Arbel{\'a}ez, Bertasius, Crandall, Damen, Engel, Farinella, Furnari, Ghanem, Hoffman, Jawahar, Newcombe, Park, Rehg, Sato, Savva, Shi, Shou, and Wray]{grauman2024egoexo4dunderstandingskilledhuman}
K.~Grauman \emph{et~al.}, ``Ego-exo4d: Understanding skilled human activity from first- and third-person perspectives,'' in \emph{IEEE/CVF Conference on Computer Vision and Pattern Recognition}, 2023, pp. 19\,383--19\,400.

\bibitem[Werling et~al.(2023)Werling, Bianco, Raitor, Stingel, Hicks, Collins, Delp, and Liu]{AddBiomechanics}
K.~Werling, N.~A. Bianco, M.~Raitor, J.~Stingel, J.~L. Hicks, S.~H. Collins, S.~L. Delp, and C.~K. Liu, ``Addbiomechanics: Automating model scaling, inverse kinematics, and inverse dynamics from human motion data through sequential optimization,'' \emph{PLOS ONE}, vol.~18, no.~11, pp. 1--25, 2023.

\bibitem[Bommasani et~al.(2022)Bommasani, Hudson, Adeli, Altman, Arora, von Arx, Bernstein, Bohg, Bosselut, Brunskill, Brynjolfsson, Buch, Card, Castellon, Chatterji, Chen, Creel, Davis, Demszky, Donahue, Doumbouya, Durmus, Ermon, Etchemendy, Ethayarajh, Fei-Fei, Finn, Gale, Gillespie, Goel, Goodman, Grossman, Guha, Hashimoto, Henderson, Hewitt, Ho, Hong, Hsu, Huang, Icard, Jain, Jurafsky, Kalluri, Karamcheti, Keeling, Khani, Khattab, Koh, Krass, Krishna, Kuditipudi, Kumar, Ladhak, Lee, Lee, Leskovec, Levent, Li, Li, Ma, Malik, Manning, Mirchandani, Mitchell, Munyikwa, Nair, Narayan, Narayanan, Newman, Nie, Niebles, Nilforoshan, Nyarko, Ogut, Orr, Papadimitriou, Park, Piech, Portelance, Potts, Raghunathan, Reich, Ren, Rong, Roohani, Ruiz, Ryan, Ré, Sadigh, Sagawa, Santhanam, Shih, Srinivasan, Tamkin, Taori, Thomas, Tramèr, Wang, Wang, Wu, Wu, Wu, Xie, Yasunaga, You, Zaharia, Zhang, Zhang, Zhang, Zhang, Zheng, Zhou, and Liang]{bommasani2021opportunities}
R.~Bommasani \emph{et~al.}, ``On the opportunities and risks of foundation models,'' \emph{arXiv preprint arXiv:2108.07258}, 2022.

\bibitem[OpenAI(2023)]{openai2023gpt4}
OpenAI, ``Gpt-4 technical report,'' 2023.

\bibitem[Brohan et~al.(2023)Brohan, Brown, Carbajal, Chebotar, Dabis, Finn, Gopalakrishnan, Hausman, Herzog, Hsu, Ibarz, Ichter, Irpan, Jackson, Jesmonth, Joshi, Julian, Kalashnikov, Kuang, Leal, Lee, Levine, Lu, Malla, Manjunath, Mordatch, Nachum, Parada, Peralta, Perez, Pertsch, Quiambao, Rao, Ryoo, Salazar, Sanketi, Sayed, Singh, Sontakke, Stone, Tan, Tran, Vanhoucke, Vega, Vuong, Xia, Xiao, Xu, Xu, Yu, and Zitkovich]{brohan2022rt}
A.~Brohan \emph{et~al.}, ``{RT-1: Robotics Transformer for Real-World Control at Scale},'' in \emph{Proceedings of Robotics: Science and Systems}, 2023.

\bibitem[Zitkovich et~al.(2023)Zitkovich, Yu, Xu, Xu, Xiao, Xia, Wu, Wohlhart, Welker, Wahid, Vuong, Vanhoucke, Tran, Soricut, Singh, Singh, Sermanet, Sanketi, Salazar, Ryoo, Reymann, Rao, Pertsch, Mordatch, Michalewski, Lu, Levine, Lee, Lee, Leal, Kuang, Kalashnikov, Julian, Joshi, Irpan, Ichter, Hsu, Herzog, Hausman, Gopalakrishnan, Fu, Florence, Finn, Dubey, Driess, Ding, Choromanski, Chen, Chebotar, Carbajal, Brown, Brohan, Arenas, and Han]{brohan2023rt}
B.~Zitkovich \emph{et~al.}, ``Rt-2: Vision-language-action models transfer web knowledge to robotic control,'' in \emph{Conference on Robot Learning}, vol. 229, 2023, pp. 2165--2183.

\bibitem[Jiang et~al.(2023{\natexlab{b}})Jiang, Gupta, Zhang, Wang, Dou, Chen, Fei-Fei, Anandkumar, Zhu, and Fan]{jiang2023vima}
Y.~Jiang, A.~Gupta, Z.~Zhang, G.~Wang, Y.~Dou, Y.~Chen, L.~Fei-Fei, A.~Anandkumar, Y.~Zhu, and L.~Fan, ``Vima: General robot manipulation with multimodal prompts,'' in \emph{International Conference on Machine Learning}, 2023.

\bibitem[Kim et~al.(2025)Kim, Pertsch, Karamcheti, Xiao, Balakrishna, Nair, Rafailov, Foster, Sanketi, Vuong, Kollar, Burchfiel, Tedrake, Sadigh, Levine, Liang, and Finn]{kim2024openvla}
M.~J. Kim \emph{et~al.}, ``Openvla: An open-source vision-language-action model,'' in \emph{Conference on Robot Learning}, vol. 270, 2025, pp. 2679--2713.

\bibitem[Team et~al.(2025)Team, Abeyruwan, Ainslie, Alayrac, Arenas, Armstrong, Balakrishna, Baruch, Bauza, Blokzijl, Bohez, Bousmalis, Brohan, Buschmann, Byravan, Cabi, Caluwaerts, Casarini, Chang, Chen, Chen, Chiang, Choromanski, D'Ambrosio, Dasari, Davchev, Devin, Palo, Ding, Dostmohamed, Driess, Du, Dwibedi, Elabd, Fantacci, Fong, Frey, Fu, Giustina, Gopalakrishnan, Graesser, Hasenclever, Heess, Hernaez, Herzog, Hofer, Humplik, Iscen, Jacob, Jain, Julian, Kalashnikov, Karagozler, Karp, Kew, Kirkland, Kirmani, Kuang, Lampe, Laurens, Leal, Lee, Lee, Liang, Lin, Maddineni, Majumdar, Michaely, Moreno, Neunert, Nori, Parada, Parisotto, Pastor, Pooley, Rao, Reymann, Sadigh, Saliceti, Sanketi, Sermanet, Shah, Sharma, Shea, Shu, Sindhwani, Singh, Soricut, Springenberg, Sterneck, Surdulescu, Tan, Tompson, Vanhoucke, Varley, Vesom, Vezzani, Vinyals, Wahid, Welker, Wohlhart, Xia, Xiao, Xie, Xie, Xu, Xu, Xu, Xu, Yang, Yao, Yaroshenko, Yu, Yuan, Zhang, Zhang, Zhou, and Zhou]{gemini_robotics_2025}
G.~R. Team \emph{et~al.}, ``Gemini robotics: Bringing ai into the physical world,'' 2025.

\bibitem[Bu et~al.(2025)Bu, Cai, Chen, Cui, Ding, Feng, Gao, He, Huang, Jiang, et~al.]{bu2025agibot}
Q.~Bu \emph{et~al.}, ``Agibot world colosseo: A large-scale manipulation platform for scalable and intelligent embodied systems,'' \emph{arXiv preprint arXiv:2503.06669}, 2025.

\bibitem[Ma et~al.(2024{\natexlab{a}})Ma, Liang, Wang, Huang, Bastani, Jayaraman, Zhu, Fan, and Anandkumar]{ma2023eureka}
Y.~J. Ma, W.~Liang, G.~Wang, D.-A. Huang, O.~Bastani, D.~Jayaraman, Y.~Zhu, L.~Fan, and A.~Anandkumar, ``Eureka: Human-level reward design via coding large language models,'' in \emph{International Conference on Learning Representations}, 2024.

\bibitem[Liang et~al.(2023)Liang, Huang, Xia, Xu, Hausman, Ichter, Florence, and Zeng]{liang2023code}
J.~Liang, W.~Huang, F.~Xia, P.~Xu, K.~Hausman, B.~Ichter, P.~Florence, and A.~Zeng, ``Code as policies: Language model programs for embodied control,'' in \emph{IEEE International Conference on Robotics and Automation}, 2023, pp. 9493--9500.

\bibitem[brian ichter et~al.(2022)brian ichter, Brohan, Chebotar, Finn, Hausman, Herzog, Ho, Ibarz, Irpan, Jang, Julian, Kalashnikov, Levine, Lu, Parada, Rao, Sermanet, Toshev, Vanhoucke, Xia, Xiao, Xu, Yan, Brown, Ahn, Cortes, Sievers, Tan, Xu, Reyes, Rettinghouse, Quiambao, Pastor, Luu, Lee, Kuang, Jesmonth, Jeffrey, Ruano, Hsu, Gopalakrishnan, David, Zeng, and Fu]{saycan2022arxiv}
brian ichter \emph{et~al.}, ``Do as i can, not as i say: Grounding language in robotic affordances,'' in \emph{Annual Conference on Robot Learning}, 2022.

\bibitem[Ma et~al.(2024{\natexlab{b}})Ma, Liang, Wang, Zhu, Fan, Bastani, and Jayaraman]{ma2024dreureka}
Y.~J. Ma, W.~Liang, H.-J. Wang, Y.~Zhu, L.~Fan, O.~Bastani, and D.~Jayaraman, ``{DrEureka: Language Model Guided Sim-To-Real Transfer},'' in \emph{Robotics: Science and Systems}, 2024.

\bibitem[Yao et~al.(2024)Yao, He, Gu, Du, Tan, Zhu, and Lu]{yao2024anybipeendtoendframeworktraining}
Y.~Yao, W.~He, C.~Gu, J.~Du, F.~Tan, Z.~Zhu, and J.~Lu, ``Anybipe: An end-to-end framework for training and deploying bipedal robots guided by large language models,'' \emph{arXiv preprint arXiv:2409.08904}, 2024.

\bibitem[Chen et~al.(2024{\natexlab{b}})Chen, Lessing, Tang, Chada, Smith, Levine, and Finn]{chen2024VLM-PC}
A.~S. Chen, A.~M. Lessing, A.~Tang, G.~Chada, L.~Smith, S.~Levine, and C.~Finn, ``Commonsense reasoning for legged robot adaptation with vision-language models,'' \emph{arXiv preprint arXiv:2407.02666}, 2024.

\bibitem[Juravsky et~al.(2022)Juravsky, Guo, Fidler, and Peng]{PADL}
J.~Juravsky, Y.~Guo, S.~Fidler, and X.~B. Peng, ``Padl: Language-directed physics-based character control,'' in \emph{SIGGRAPH Asia 2022 Conference Papers}, 2022, pp. 1--9.

\bibitem[Sun et~al.(2024)Sun, Zhang, Duan, Jiang, Cheng, and Xu]{Prompt_Plan_Perform}
J.~Sun, Q.~Zhang, Y.~Duan, X.~Jiang, C.~Cheng, and R.~Xu, ``Prompt, plan, perform: Llm-based humanoid control via quantized imitation learning,'' in \emph{IEEE International Conference on Robotics and Automation}, 2024, pp. 16\,236--16\,242.

\bibitem[Wang et~al.(2024)Wang, Dai, Wang, Rossini, Ruscelli, and Tsagarakis]{wang2024hypermotion}
J.~Wang, R.~Dai, W.~Wang, L.~Rossini, F.~Ruscelli, and N.~Tsagarakis, ``{HYPER}motion: Learning hybrid behavior planning for autonomous loco-manipulation,'' in \emph{Annual Conference on Robot Learning}, 2024.

\bibitem[Singh et~al.(2023)Singh, Blukis, Mousavian, Goyal, Xu, Tremblay, Fox, Thomason, and Garg]{ProgPrompt2023}
I.~Singh, V.~Blukis, A.~Mousavian, A.~Goyal, D.~Xu, J.~Tremblay, D.~Fox, J.~Thomason, and A.~Garg, ``Progprompt: Generating situated robot task plans using large language models,'' in \emph{IEEE International Conference on Robotics and Automation}, 2023, pp. 11\,523--11\,530.

\bibitem[Yu et~al.(2023)Yu, Gileadi, Fu, Kirmani, Lee, Arenas, Chiang, Erez, Hasenclever, Humplik, brian ichter, Xiao, Xu, Zeng, Zhang, Heess, Sadigh, Tan, Tassa, and Xia]{yu2023language}
W.~Yu \emph{et~al.}, ``Language to rewards for robotic skill synthesis,'' in \emph{Annual Conference on Robot Learning}, 2023.

\bibitem[Jiang et~al.(2024{\natexlab{b}})Jiang, Xie, Li, Yuan, Zhu, and Zhu]{jiang2024harmon}
Z.~Jiang, Y.~Xie, J.~Li, Y.~Yuan, Y.~Zhu, and Y.~Zhu, ``Harmon: Whole-body motion generation of humanoid robots from language descriptions,'' in \emph{Annual Conference on Robot Learning}, 2024.

\bibitem[Kumar et~al.(2023)Kumar, Essa, and Ha]{kumar2023Words_into_Action}
K.~N. Kumar, I.~Essa, and S.~Ha, ``Words into action: Learning diverse humanoid robot behaviors using language guided iterative motion refinement,'' \emph{CoRR}, vol. abs/2310.06226, 2023.

\bibitem[Xiao et~al.(2024)Xiao, Wang, Wang, Cao, Zhang, Dai, Lin, and Pang]{xiao2024unified}
Z.~Xiao, T.~Wang, J.~Wang, J.~Cao, W.~Zhang, B.~Dai, D.~Lin, and J.~Pang, ``Unified human-scene interaction via prompted chain-of-contacts,'' in \emph{The International Conference on Learning Representations}, 2024.

\bibitem[Totsila et~al.(2024)Totsila, Rouxel, Mouret, and Ivaldi]{totsila2024words2contact}
D.~Totsila, Q.~Rouxel, J.-B. Mouret, and S.~Ivaldi, ``Words2contact: Identifying support contacts from verbal instructions using foundation models,'' in \emph{IEEE-RAS International Conference on Humanoid Robots}, 2024, pp. 9--16.

\bibitem[Vaswani et~al.(2017)Vaswani, Shazeer, Parmar, Uszkoreit, Jones, Gomez, Kaiser, and Polosukhin]{Attention_Need}
A.~Vaswani, N.~Shazeer, N.~Parmar, J.~Uszkoreit, L.~Jones, A.~N. Gomez, L.~u. Kaiser, and I.~Polosukhin, ``Attention is all you need,'' in \emph{Advances in Neural Information Processing Systems}, vol.~30, 2017.

\bibitem[Black et~al.(2024)Black, Brown, Driess, Esmail, Equi, Finn, Fusai, Groom, Hausman, Ichter, Jakubczak, Jones, Ke, Levine, Li-Bell, Mothukuri, Nair, Pertsch, Shi, Tanner, Vuong, Walling, Wang, and Zhilinsky]{black2024pi0}
K.~Black \emph{et~al.}, ``$\pi_0$: A vision-language-action flow model for general robot control,'' \emph{arXiv preprint arXiv:2410.24164}, 2024.

\bibitem[O’Neill et~al.(2024)O’Neill, Rehman, Maddukuri, Gupta, Padalkar, Lee, Pooley, Gupta, Mandlekar, Jain, Tung, Bewley, Herzog, Irpan, Khazatsky, Rai, Gupta, Wang, Singh, Garg, Kembhavi, Xie, Brohan, Raffin, Sharma, Yavary, Jain, Balakrishna, Wahid, Burgess-Limerick, Kim, Schölkopf, Wulfe, Ichter, Lu, Xu, Le, Finn, Wang, Xu, Chi, Huang, Chan, Agia, Pan, Fu, Devin, Xu, Morton, Driess, Chen, Pathak, Shah, Büchler, Jayaraman, Kalashnikov, Sadigh, Johns, Foster, Liu, Ceola, Xia, Zhao, Stulp, Zhou, Sukhatme, Salhotra, Yan, Feng, Schiavi, Berseth, Kahn, Wang, Su, Fang, Shi, Bao, Ben~Amor, Christensen, Furuta, Walke, Fang, Ha, Mordatch, Radosavovic, Leal, Liang, Abou-Chakra, Kim, Drake, Peters, Schneider, Hsu, Bohg, Bingham, Wu, Gao, Hu, Wu, Wu, Sun, Luo, Gu, Tan, Oh, Wu, Lu, Yang, Malik, Silvério, Hejna, Booher, Tompson, Yang, Salvador, Lim, Han, Wang, Rao, Pertsch, Hausman, Go, Gopalakrishnan, Goldberg, Byrne, Oslund, Kawaharazuka, Black, Lin, Zhang, Ehsani, Lekkala, Ellis, Rana, Srinivasan, Fang,
  Singh, Zeng, Hatch, Hsu, Itti, Chen, Pinto, Fei-Fei, Tan, Fan, Ott, Lee, Weihs, Chen, Lepert, Memmel, Tomizuka, Itkina, Castro, Spero, Du, Ahn, Yip, Zhang, Ding, Heo, Srirama, Sharma, Kim, Kanazawa, Hansen, Heess, Joshi, Suenderhauf, Liu, Di~Palo, Shafiullah, Mees, Kroemer, Bastani, Sanketi, Miller, Yin, Wohlhart, Xu, Fagan, Mitrano, Sermanet, Abbeel, Sundaresan, Chen, Vuong, Rafailov, Tian, Doshi, Martín-Martín, Baijal, Scalise, Hendrix, Lin, Qian, Zhang, Mendonca, Shah, Hoque, Julian, Bustamante, Kirmani, Levine, Lin, Moore, Bahl, Dass, Sonawani, Song, Xu, Haldar, Karamcheti, Adebola, Guist, Nasiriany, Schaal, Welker, Tian, Ramamoorthy, Dasari, Belkhale, Park, Nair, Mirchandani, Osa, Gupta, Harada, Matsushima, Xiao, Kollar, Yu, Ding, Davchev, Zhao, Armstrong, Darrell, Chung, Jain, Vanhoucke, Zhan, Zhou, Burgard, Chen, Wang, Zhu, Geng, Liu, Liangwei, Li, Lu, Ma, Kim, Chebotar, Zhou, Zhu, Wu, Xu, Wang, Bisk, Cho, Lee, Cui, Cao, Wu, Tang, Zhu, Zhang, Jiang, Li, Li, Iwasawa, Matsuo, Ma, Xu, Cui, Zhang, and
  Lin]{open_x_embodiment_rt_x_2023}
A.~O’Neill \emph{et~al.}, ``Open x-embodiment: Robotic learning datasets and rt-x models : Open x-embodiment collaboration0,'' in \emph{IEEE International Conference on Robotics and Automation}, 2024, pp. 6892--6903.

\bibitem[Reed et~al.(2022)Reed, Zolna, Parisotto, Colmenarejo, Novikov, Barth-maron, Gim{\'e}nez, Sulsky, Kay, Springenberg, Eccles, Bruce, Razavi, Edwards, Heess, Chen, Hadsell, Vinyals, Bordbar, and de~Freitas]{reed2022generalist}
S.~Reed \emph{et~al.}, ``A generalist agent,'' \emph{Transactions on Machine Learning Research, issn: 2835-8856}, 2022.

\bibitem[Gu and Dao(2023)]{gu2023mamba}
A.~Gu and T.~Dao, ``Mamba: Linear-time sequence modeling with selective state spaces,'' \emph{arXiv preprint arXiv:2312.00752}, 2023.

\bibitem[Ghosh et~al.(2024)Ghosh, Walke, Pertsch, Black, Mees, Dasari, Hejna, Kreiman, Xu, Luo, Tan, Chen, Vuong, Xiao, Sanketi, Sadigh, Finn, and Levine]{octo}
D.~Ghosh \emph{et~al.}, ``{Octo: An Open-Source Generalist Robot Policy},'' in \emph{Robotics: Science and Systems}, 2024.

\bibitem[Liu et~al.(2025)Liu, Wu, Li, Tan, Chen, Wang, Xu, Su, and Zhu]{liu2024rdt1b}
S.~Liu, L.~Wu, B.~Li, H.~Tan, H.~Chen, Z.~Wang, K.~Xu, H.~Su, and J.~Zhu, ``{RDT}-1b: a diffusion foundation model for bimanual manipulation,'' in \emph{International Conference on Learning Representations}, 2025.

\bibitem[Cheang et~al.(2024)Cheang, Chen, Jing, Kong, Li, Li, Liu, Wu, Xu, Yang, et~al.]{cheang2024gr2}
C.-L. Cheang \emph{et~al.}, ``Gr-2: A generative video-language-action model with web-scale knowledge for robot manipulation,'' \emph{arXiv preprint arXiv:2410.06158}, 2024.

\bibitem[Leal et~al.(2024)Leal, Choromanski, Jain, Dubey, Varley, Ryoo, Lu, Liu, Sindhwani, Vuong, Sarlos, Oslund, Hausman, and Rao]{leal2023sarart}
I.~Leal \emph{et~al.}, ``Sara-rt: Scaling up robotics transformers with self-adaptive robust attention,'' in \emph{International Conference on Robotics and Automation}, 2024, pp. 6920--6927.

\bibitem[Touvron et~al.(2023)Touvron, Lavril, Izacard, Martinet, Lachaux, Lacroix, Rozi{\`e}re, Goyal, Hambro, Azhar, et~al.]{touvron2023llama}
H.~Touvron \emph{et~al.}, ``Llama: Open and efficient foundation language models,'' \emph{arXiv preprint arXiv:2302.13971}, 2023.

\bibitem[Kaplan et~al.(2020)Kaplan, McCandlish, Henighan, Brown, Chess, Child, Gray, Radford, Wu, and Amodei]{kaplan2020scaling}
J.~Kaplan, S.~McCandlish, T.~Henighan, T.~B. Brown, B.~Chess, R.~Child, S.~Gray, A.~Radford, J.~Wu, and D.~Amodei, ``Scaling laws for neural language models,'' \emph{arXiv preprint arXiv:2001.08361}, 2020.

\bibitem[Lin et~al.(2024{\natexlab{b}})Lin, Hu, Sheng, Wen, You, and Gao]{lin2024data}
F.~Lin, Y.~Hu, P.~Sheng, C.~Wen, J.~You, and Y.~Gao, ``Data scaling laws in imitation learning for robotic manipulation,'' \emph{arXiv preprint arXiv:2410.18647}, 2024.

\bibitem[Nasiriany et~al.(2024)Nasiriany, Xia, Yu, Xiao, Liang, Dasgupta, Xie, Driess, Wahid, Xu, et~al.]{nasiriany2024pivot}
S.~Nasiriany \emph{et~al.}, ``Pivot: Iterative visual prompting elicits actionable knowledge for vlms,'' \emph{arXiv preprint arXiv:2402.07872}, 2024.

\bibitem[Kim et~al.(2024{\natexlab{b}})Kim, Woo, and Lee]{kim2024addressing}
D.~Kim, H.~Woo, and Y.~Lee, ``Addressing bias and fairness using fair federated learning: A synthetic review,'' \emph{Electronics}, vol.~13, no.~23, p. 4664, 2024.

\bibitem[Zhuang and Zhao(2025)]{zhuang2025embracecollisionshumanoidshadowing}
Z.~Zhuang and H.~Zhao, ``Embrace collisions: Humanoid shadowing for deployable contact-agnostics motions,'' 2025.

\bibitem[Ficht and Behnke(2021)]{ficht2021bipedal}
G.~Ficht and S.~Behnke, ``Bipedal humanoid hardware design: A technology review,'' \emph{Current Robotics Reports}, vol.~2, pp. 201--210, 2021.

\bibitem[Park and Cho(2023)]{park2023design}
M.-H. Park and B.-K. Cho, ``Design of new drive mechanism for dynamic movement of humanoid robot legs,'' in \emph{2023 IEEE-RAS 22nd International Conference on Humanoid Robots (Humanoids)}, 2023, pp. 1--8.

\bibitem[Mach{\'\i}n et~al.(2024)Mach{\'\i}n, Morant, and M{\'a}rquez]{machin2024advancements}
A.~Mach{\'\i}n, C.~Morant, and F.~M{\'a}rquez, ``Advancements and challenges in solid-state battery technology: An in-depth review of solid electrolytes and anode innovations,'' \emph{Batteries}, vol.~10, no.~1, p.~29, 2024.

\bibitem[Apptronik(2024)]{Apptronik}
Apptronik, ``Apptronik apollo,'' 2024.

\bibitem[robotics(2024)]{H1}
U.~robotics, ``Unitree h1 robot,'' 2024.

\end{thebibliography}


 




\vfill

\end{document}